%% file: main.tex
\documentclass{article}
\usepackage{microtype}
\usepackage{graphicx}
\usepackage{wrapfig}
\usepackage{booktabs} 
\usepackage{makecell}
\usepackage{graphicx}
\usepackage{caption}
\usepackage{courier}
\usepackage{multirow}
\usepackage{color}
\usepackage{adjustbox}
\usepackage{subcaption}
\usepackage{amssymb}
\usepackage{amsmath}
\usepackage{gensymb}
\usepackage{amsfonts}       
\usepackage{enumitem}
\usepackage{multirow}
\usepackage{algorithm}
\usepackage{algorithmic}
\usepackage{amsmath,nccmath}

\usepackage{hyperref}

\usepackage{bbm}

\usepackage[accepted]{icml2021}

\icmltitlerunning{Improving Generalization in Meta-learning via Task Augmentation}

\newtheorem{Theorem}{Theorem}
\newtheorem{cor}{Corollary}
\newtheorem{lemma}{Lemma}
\newtheorem{Proof}{Proof}

\newcommand{\yao}[1]{{\color{black}#1}}
\newcommand{\yaocr}[1]{{\color{black}#1}}
\newcommand{\kai}[1]{{\color{black}#1}}
\newcommand{\ying}[1]{{\color{black}#1}}
\newcommand{\yingicml}[1]{{\color{black}#1}}

\newcommand{\ours}{{MetaMix}}
\newcommand{\cf}{{Channel Shuffle}}

\begin{document}

\twocolumn[
\icmltitle{Improving Generalization in Meta-learning via Task Augmentation}

\begin{icmlauthorlist}
\icmlauthor{Huaxiu Yao$^\dag$}{stanford}
\icmlauthor{Long-Kai Huang}{tencent}
\icmlauthor{Linjun Zhang}{rutgers}
\icmlauthor{Ying Wei}{cityu}
\icmlauthor{Li Tian}{tencent}
\\
\icmlauthor{James Zou}{stanford}
\icmlauthor{Junzhou Huang}{tencent}
\icmlauthor{Zhenhui Li}{psu}
\end{icmlauthorlist}

\icmlaffiliation{psu}{Pennsylvania State University, PA, USA}
\icmlaffiliation{tencent}{Tencent AI Lab, Shenzhen, China}
\icmlaffiliation{cityu}{City University of Hong Kong, Hong Kong}
\icmlaffiliation{rutgers}{Rutgers University, NJ, USA}
\icmlaffiliation{stanford}{Stanford University, CA, USA}

\icmlcorrespondingauthor{Ying Wei}{yingwei@cityu.edu.hk}

\icmlkeywords{Machine Learning, ICML}

\vskip 0.3in

]

\printAffiliationsAndNotice{$^\dag$Part of the work was done when H.Y. was a student at Penn State University. H.Y. and LK.H. contribute equally.}

\input{abstract}
\input{introduction}
\input{preliminary}
\input{method}
\input{theoretic_study}
\input{relatedwork}
\input{experiment}
\input{conclusion}
\section*{Acknowledgments}
H.Y. and Z.L. are supported in part by NSF awards IIS-\#1652525 and IIS-\#1618448. L.Z. is supported by NSF awards DMS-\#2015378. The views and conclusions contained in this paper are those of the authors and should not be interpreted as representing any funding agencies.
\bibliography{ref}
\bibliographystyle{icml2021}
\clearpage
\onecolumn
\input{app-validity}
\input{app-method}
\input{app-generalization}

\input{app-setting}
\input{app-drug}
\input{app-pose}

\input{app-image}
\end{document}

%% file: abstract.tex
\begin{abstract}
Meta-learning has proven to be a powerful paradigm for transferring the knowledge 
from previous tasks to facilitate the learning of a novel task. Current dominant
algorithms train a well-generalized model initialization which is adapted to each task via the support set. The crux lies in optimizing the generalization capability of the initialization, which is measured by the performance of the adapted model on the query set of each task.
Unfortunately, this generalization measure, evidenced by empirical results, pushes the initialization to overfit the meta-training tasks, which significantly impairs the generalization and adaptation to novel tasks. To address this issue, we actively \yingicml{augment a meta-training task with ``more data''}
when evaluating the generalization.
Concretely, 
\yingicml{we propose two task augmentation methods, including MetaMix and Channel Shuffle.}
MetaMix linearly combines \yingicml{features and labels of}
samples from both the support and query sets. For each class of samples, \yingicml{C}hannel Shuffle randomly replaces a subset of their channels with the corresponding \yingicml{ones} from a different class. Theoretical studies show how task augmentation 
improve\yingicml{s} the generalization of meta-learning. Moreover, both MetaMix and Channel Shuffle 
\yingicml{outperform}
state-of-the-art results by a large margin across many datasets and are compatible with existing meta-learning algorithms. 
\end{abstract}

%% file: introduction.tex
\section{Introduction}
Meta-learning\ying{, or learning to learn~\cite{thrun2012learning}, empowers agents with the core aspect of intelligence--quickly learning a new task with as little as a few 
examples by drawing upon the knowledge learned from prior tasks.
The resurgence of meta-learning recently pushes ahead with more effective 
algorithms that have been deployed in areas}
such as 
computer vision~\cite{kang2019few,liu2019few,sung2018learning}, natural language processing~\cite{dou2019investigating,gu2018meta,madotto2019personalizing}, \ying{and robotics~\cite{xie2018few,yu2018one}}. 
\ying{Some of the dominant algorithms learn a transferable metric space from previous tasks~\cite{snell2017prototypical,vinyals2016matching}, unfortunately being only applicable to classification problems. 
Instead,}
gradient-based 
\ying{algorithms}~\cite{finn2017model,finn2018probabilistic} \ying{framing meta-learning as a bi-level optimization problem are flexible and general enough to be independent of problem types, which we focus on in this work.}

\begin{figure*}[t]
\centering
\begin{subfigure}[c]{0.3\textwidth}
	\centering
	\includegraphics[width=\textwidth]{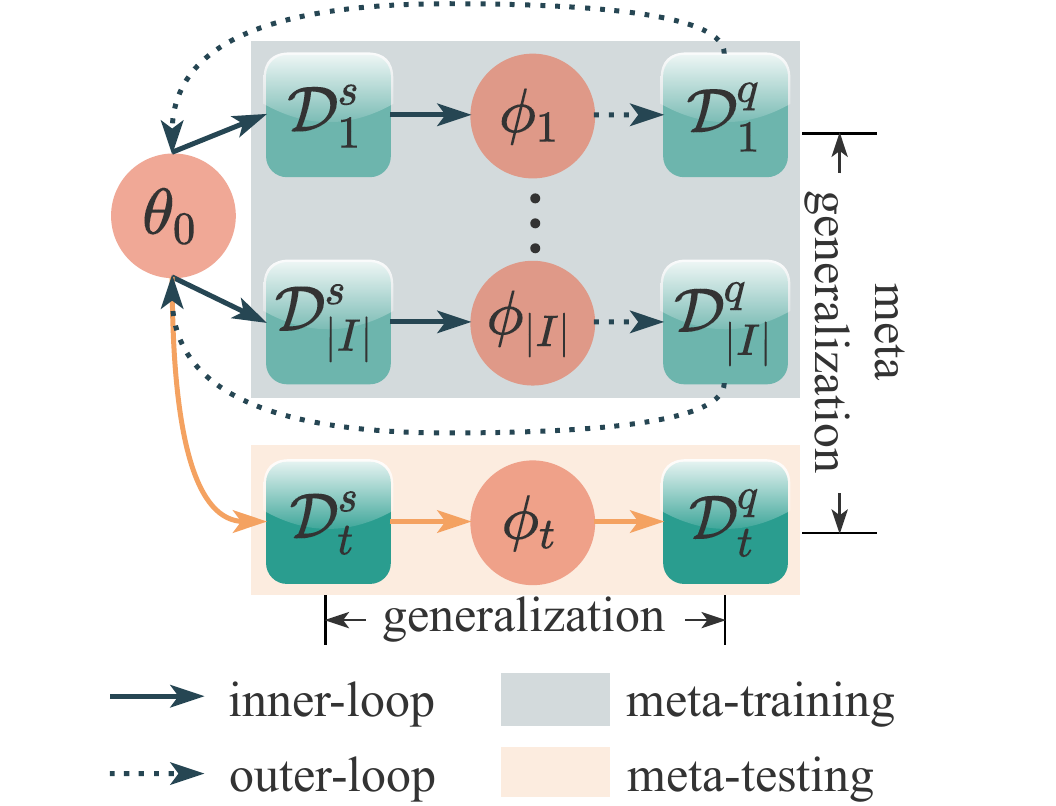}
	\caption{\label{fig:optimization_illustration}: Gradient-based Meta-learning}
\end{subfigure}
\begin{subfigure}[c]{0.3\textwidth}
	\centering
	\includegraphics[width=\textwidth]{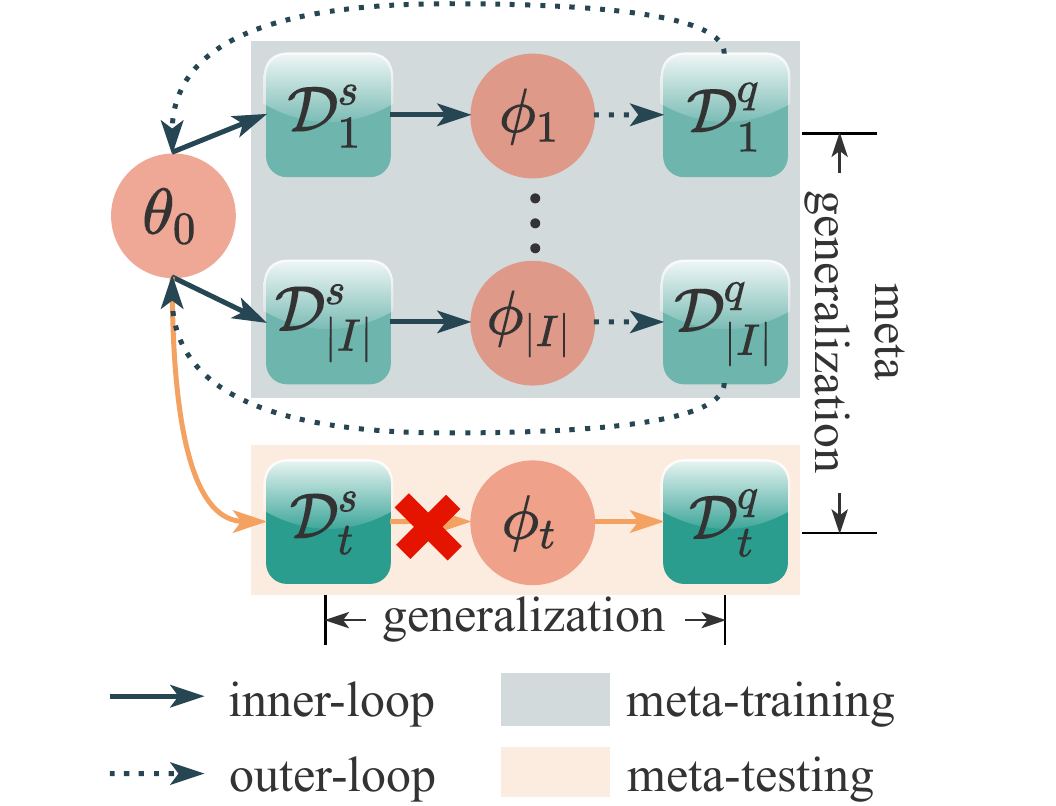}
	\caption{\label{fig:memorization_overfitting}: Memorization Overfitting}
\end{subfigure}
\begin{subfigure}[c]{0.3\textwidth}
	\centering
	\includegraphics[width=\textwidth]{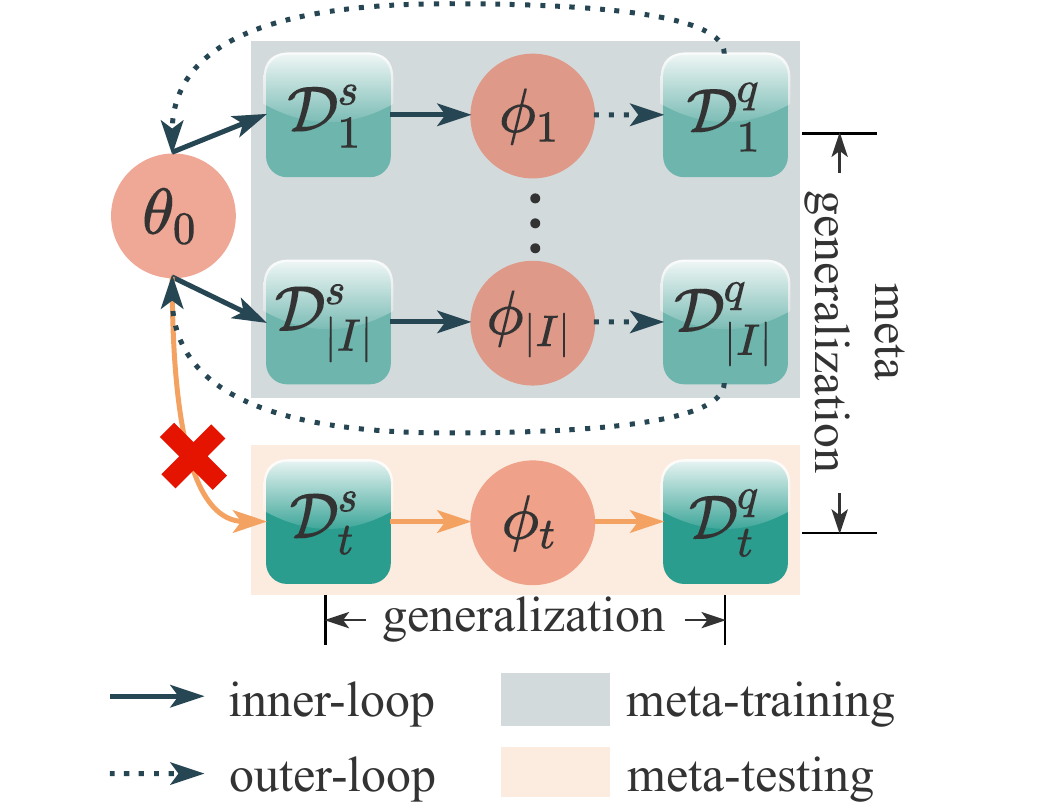}
	\caption{\label{fig:learner_overfitting}: Learner Overfitting}
\end{subfigure}
\caption{(a) Illustration of the gradient-based meta-learning process and two types of generalization; (b)\&(c) Two forms of overfitting in gradient-based meta-learning. The red cross represents where the learned knowledge can not be well-generalized.}
\vspace{-1em}
\end{figure*}

\ying{The bi-level optimization procedure of gradient-based 
algorithms is illustrated in Figure~\ref{fig:optimization_illustration}.
In the inner-loop, the initialization of a base model (a.k.a., base learner) globally shared across 
tasks (i.e., \begin{small}$\theta_0$\end{small}) is adapted to each task (e.g., \begin{small}$\phi_1$\end{small} for the first task) via gradient descent over the support set of the task.
To reach the desired goal that optimizing from this initialization leads to fast adaptation and generalization, a meta-training objective evaluating the generalization capability of the initialization on all meta-training tasks is optimized in the outer-loop.
Specifically, the generalization capability on each task is measured by the performance of the adapted model on a set distinguished from the support, namely the query set.
}

The learned initialization, however, is at high risk of two forms of overfitting: (1) \emph{memorization overfitting}\yingicml{~\cite{yin2020meta}} (Figure~\ref{fig:memorization_overfitting}) where
\yingicml{it solves meta-training tasks via rote memorization and does not rely on}
\yingicml{support sets for}
inner-loop adaptation 
and (2) \emph{learner overfitting}~\cite{rajendran2020meta} (Figure~\ref{fig:learner_overfitting}) where it 
\yingicml{overfits to the meta-training tasks and fails to}
generalize to the meta-testing tasks 
\yingicml{though support sets come into play during inner-loop adaptation. Both types of overfitting hurt the}
generalization from meta-training to meta-testing tasks, which we call meta-generalization \yingicml{in Figure~\ref{fig:optimization_illustration}}.
\yingicml{Improving the meta-generalization}
is especially challenging -- standard regularizers like weight decay lose their power as they 
\yingicml{limit}
the flexibility of fast adaptation in the inner-loop.

To this end, the few existing solutions attempt to regularize the search space of the initialization~\cite{yin2020meta} or enforce a fair performance of the initialization across all meta-training tasks~\cite{jamal2019task} while preserving the expressive power for adaptation. Rather than passively imposing regularization 
on the initialization, recently,~\citet{rajendran2020meta} turned towards an active data augmentation way, aiming to anticipate more data to meta-train the initialization by injecting the same noise to the labels 
\yingicml{of both}
support and query sets (i.e., label shift). Though the label shift with a random constant increases the dependenc\yingicml{e} 
\yingicml{of the base learner on the}
support set, 
\yingicml{learning the constant is as easy as modifying a bias.} Therefore, \yingicml{little extra knowledge is introduced to meta-train the initialization.}

\yingicml{This paper sets out to investigate}
\yao{
more flexible and powerful ways} to produce ``more'' data via task augmentation. The goal for task augmentation is to 
increase the dependenc\yingicml{e} 
of \yingicml{target predictions on the} support set 
and 
\yingicml{provide}
additional knowledge to optimize the model initialization. 
\yingicml{To meet the goal,}
we propose two \yingicml{task augmentation} strategies -- \textbf{\ours} and \textbf{Channel Shuffle}. \ours\ linearly combines
\ying{either original features or hidden representations of}
the support and query sets
, and performs the same linear interpolation between their corresponding labels. For classification problem\yingicml{s}, 
\ours\ is further enhanced by 
\yingicml{the strategy of}
Channel Shuffle, 
which is named as MMCF. For samples of each class, Channel Shuffle randomly
\yingicml{selects}
a subset of channels \yingicml{to replace} with 
corresponding ones \yingicml{of samples} from a different class.
These additional signals for the meta-training objective improve the meta-generalization \yingicml{of the learned initialization} as expected.

We would highlight the primary contributions of this work.
(1) We identify and formalize 
\yingicml{effective}
task augmentation \yingicml{that is sufficient for alleviating} 
both memorization overfitting and learner overfitting \yingicml{and thereby improving meta-generalization}, resulting in two \yingicml{task} augmentation methods.
(2) Both task augmentation strategies 
\yingicml{have been}
theoretically proved to \yingicml{indeed} improve the \yingicml{meta-}generalization.
(3) Throughout comprehensive experiments, we demonstrate two significant benefits of \yingicml{the two} 
augmentation 
strategies.
First, in various real-world datasets, the performances are substantially improved over state-of-the-art meta-learning algorithms and other strategies for overcoming overfitting~\cite{jamal2019task,yin2020meta}.
Second, both \ours\ and MMCF are compatible with existing and advanced meta-learning algorithms and ready to boost their performances.

%% file: preliminary.tex
\section{Preliminaries}
\label{sec:preliminaries}
Gradient-based meta-learning algorithms \ying{assume}
a set of tasks 
to be
sampled from a 
distribution \begin{small}$p(\mathcal{T})$\end{small}. Each task \begin{small}$\mathcal{T}_i$\end{small} consists of a support sample 
set \begin{small}$\mathcal{D}^{s}_i=\{(\mathbf{x}^{s}_{i,j},\mathbf{y}^{s}_{i,j})\}_{j=1}^{K^s}$\end{small}
and a query sample set \begin{small}$\mathcal{D}^{q}_i=\{(\mathbf{x}^{q}_{i,j},\mathbf{y}^{q}_{i,j})\}_{j=1}^{K^q}$\end{small}, where \begin{small}$K^s$\end{small} and \begin{small}$K^q$\end{small} denote the number of source and query samples, respectively.
The objective of meta-learning is to 
\ying{master}
new tasks quickly by adapting a well-generalized model learned over the task distribution \begin{small}$p(\mathcal{T})$\end{small}. Specifically, 
\ying{the} model $f$ 
\ying{parameterized by}
\begin{small}$\theta$\end{small} is trained on 
\ying{massive}
tasks \ying{sampled from $p(\mathcal{T})$}
during meta-training. 
\ying{When it comes to meta-testing, $f$ is adapted to}
a new task \begin{small}$\mathcal{T}_{t}$\end{small} 
\ying{with the help of}
the support set \begin{small}$\mathcal{D}_{t}^{s}$\end{small}
and 
\ying{evaluated}
on the query set \begin{small}$\mathcal{D}_{t}^{q}$\end{small}.

Take model-agnostic meta-learning (MAML)~\citep{finn2017model} as an example.
\ying{The well-generalized model is grounded to}
an 
initialization for $f$, \ying{i.e.,} \begin{small}$\theta_0$\end{small}, 
which \ying{is} 
adapted to 
\ying{each $i$-th task}
\ying{in}
a few gradient steps 
\ying{by}
\ying{its}
\ying{support}
set \begin{small}$\mathcal{D}_i^s$\end{small}. The generalization performance of 
\ying{the adapted model, i.e.,}
\begin{small}$\phi_i$\end{small},
\ying{is measured}
on \ying{the} query set \begin{small}$\mathcal{D}_i^q$\end{small}, \ying{and in turn} 
used to optimize the 
initialization \begin{small}$\theta_0$\end{small} during 
meta-training. 
Let \begin{small}$\mathcal{L}$\end{small} and \begin{small}$\mu$\end{small} 
\ying{denote}
the loss function and the inner-loop learning rate, respectively. \ying{The above interleaved process is formulated as a}
bi-level optimization \ying{problem,} 
\begin{equation}
\small
\begin{split}
\label{eq:meta_update_expectation}
    &\theta_0^{*}:=\min_{\theta_0}\mathbb{E}_{\mathcal{T}_i\sim p(\mathcal{T})}\left [\mathcal{L}(f_{\phi_i}(\mathbf{X}_i^q),\mathbf{Y}_i^q)\right],\;\\ &\ying{\mathrm{s.t.}}\;\phi_i=\theta_0-\mu\nabla_{\theta_0}\mathcal{L}(f_{\theta_0}(\mathbf{X}_i^s), \mathbf{Y}_i^s),
\end{split}
\end{equation}
where \begin{small}$\mathbf{X}_i^{s(q)}$\end{small} and \begin{small}$\mathbf{Y}_i^{s(q)}$\end{small} represent the 
\yingicml{collection}
of 
samples and their corresponding labels for \ying{the} support (query) set, respectively. The predicted value \begin{small}$f_{\phi_i}(\mathbf{X}_i^{s(q)})$\end{small} is denoted as \begin{small}$\hat{\mathbf{Y}}_i^{s(q)}$\end{small}.
\ying{In the}
meta-testing phase, 
\ying{to solve}
the new task \begin{small}$\mathcal{T}_t$\end{small}, 
the 
\ying{optimal}
initialization  \begin{small}$\theta_0^{*}$\end{small} is fine\ying{-}tuned
\ying{on its support set \begin{small}$\mathcal{D}_t^s$\end{small}}
\ying{to the resulting}
task-specific parameters \begin{small}$\phi_t$\end{small}.

%% file: method.tex
\newtheorem{definition}{Definition}
\section{Task Augmentation}
\label{sec:ours}
\ying{
In practical situations, the distribution \begin{small}$p(\mathcal{T})$\end{small} is unknown for estimation of the expected performance in Eqn.~\eqref{eq:meta_update_expectation}. Instead, the common practice is to approximate it with the empirical performance, i.e.,
\begin{equation}
\small
\begin{split}
\label{eq:meta_update}
    &\theta_0^{*}:=\min_{\theta_0}\frac{1}{n_T}\sum_{i=1}^{n_T}\left [\mathcal{L}(f_{\phi_i}(\mathbf{X}_i^q),\mathbf{Y}_i^q)\right],\;\\ &\ying{\mathrm{s.t.}}\;\phi_i=\theta_0-\mu\nabla_{\theta_0}\mathcal{L}(f_{\theta_0}(\mathbf{X}_i^s), \mathbf{Y}_i^s).
\end{split}
\end{equation}

Unfortunately, this empirical risk observes the generalization ability of the initialization \begin{small}$\theta_0$\end{small} only at a finite set of \begin{small}$n_T$\end{small} tasks.
When the function \begin{small}$f$\end{small} is sufficiently powerful, a trivial solution of \begin{small}$\theta_0$\end{small} is to overfit all tasks.
\yingicml{Compared to standard supervised learning, the overfitting is more complicated with two cases: }\emph{memorization overfitting} and \emph{learner overfitting} \yingicml{which differ primarily in whether the support set contributes to inner-loop adaptation}. In memorization overfitting, 
\yingicml{$\theta_0^*$ memorizes all tasks, so that the adaptation to each task via its support set is even futile~\cite{yin2020meta}.} In learner overfitting, 
\yingicml{$\theta_0^*$ fails to generalize to new tasks, though it adapts to solve each   meta-training task sufficiently} with the corresponding support set~\cite{rajendran2020meta}. \yingicml{Both} 
overfitting lead 
to poor meta-generalization (see Figure~\ref{fig:optimization_illustration}).

Inspired by data augmentation~\cite{cubuk2019autoaugment,zhang2018mixup,zhong2020random,zhang2020does} which is used to mitigate the overfitting of training samples in conventional supervised learning, we propose to alleviate the problem of task overfitting via task augmentation. Before proceeding to our solutions, we first formally define two criteria for an effective task augmentation as:

\begin{definition}[Criteria of Effective Task Augmentation]
\label{def:task_aug}
An 
\yingicml{effective}
task augmentation for meta-learning is an augmentation function \begin{small}$g(\cdot)$\end{small} that transforms a task \begin{small}$\mathcal{T}_i=\{\mathcal{D}_i^s,\mathcal{D}_i^q\}$\end{small} to an augmentated task \begin{small}$\mathcal{T}_i^{'}=\{g(\mathcal{D}_i^s),g(\mathcal{D}_i^q)\}$\end{small}, so that the following two criteria are met:

(1) \begin{small}$I(g(\hat{\mathbf{Y}}_i^q);g(\mathcal{D}_i^s)|\theta_0,g(\mathbf{X}_i^q)) - I(\hat{\mathbf{Y}}_i^q;\mathcal{D}_i^s|\theta_0,\mathbf{X}_i^q) > 0$\end{small},
           
(2) \begin{small}$I(\theta_0;g(\mathcal{D}_i^q) | \mathcal{D}_i^q) > 0$\end{small}.
\label{def:task_augmentation}
\end{definition}

The augmented task satisfying the first criterion is expected to alleviate the memorization overfitting, as the model more heavily relies on the 
\yingicml{support set}
\begin{small}$\mathcal{D}_i^s$\end{small} to make predictions, i.e., increasing the mutual information between \begin{small}$g(\hat{\mathbf{Y}}_i^q)$\end{small} and \begin{small}$g(\mathcal{D}_i^s)$\end{small}.
The second criterion guarantees that the augmented task contributes additional knowledge to update the initialization in the outer-loop. With more augmented meta-training tasks satisfying this criterion, the meta-generalization ability of the initialization to meta-testing tasks improves. 
\yingicml{Building on this,}
we will introduce the proposed 
\yingicml{task}
augmentation strategies.

\textbf{MetaMix.}
One of the most immediate choices for task augmentation is 
\yingicml{directly incorporating}
support sets in the outer-loop, while it is far from enough. The support sets contribute little to the value and gradients of the meta-training objective, as the meta-training objective is formulated as the performance of the adapted model which is exactly optimized via support sets. Thus, we are motivated to produce “more” data out of the accessible support and query sets, resulting in \ours, which meta-trains \begin{small}$\theta_0$\end{small} by mixing samples from both \yingicml{the} query set 
and \yingicml{the} support set. 

\ying{In detail, t}he strategy of mixing follows Manifold Mixup~\cite{verma2019manifold} where not only inputs but also hidden representations are mixed up.} Assume \ying{that} the 
model \begin{small}$f$\end{small} 
\ying{consists of}
\begin{small}$L$\end{small} layers. 
\ying{The hidden representation of a sample set \begin{small}$\mathbf{X}$\end{small} at the \begin{small}$l$\end{small}-th}
layer \ying{is denoted}
as
\begin{small}$f_{\theta^l}(\mathbf{X})$\end{small}~\ying{(\begin{small}$0 \leq\! l\! \leq L-1 $\end{small})}, where \begin{small}$f_{\theta^0}(\mathbf{X})=\mathbf{X}$\end{small}.
\ying{
For a pair of support and query sets with their corresponding labels in the $i$-th task \begin{small}$\mathcal{T}_i$\end{small}, i.e.,  \begin{small}$(\mathbf{X}^s_i, \mathbf{Y}^s_i)$\end{small} and \begin{small}$(\mathbf{X}^q_i,\mathbf{Y}^q_i)$\end{small}, we
randomly sample a value of \begin{small}$l\in\mathcal{C}=\{0,1,\cdots,L-1\}$\end{small} and 
compute the mixed batch of data for meta-training as,
}
\begin{equation}
\small
\begin{aligned}
\label{eq:metamix}
&\mathbf{X}^{mix}_{i,l}=\pmb{\lambda} f_{\phi_i^l}(\mathbf{X}^s_i)+(\mathbf{I}-\pmb{\lambda})f_{\phi_i^l}(\mathbf{X}^q_i),\\\;\; &\mathbf{Y}^{mix}_i=\pmb{\lambda}\mathbf{Y}_i^s+(\mathbf{I}-\pmb{\lambda})\mathbf{Y}_i^q,
\end{aligned}
\end{equation}
where \begin{small}$\pmb{\lambda}\!=\!\mathrm{diag}(\{\lambda_j\}_{j=1}^{K^q})$\end{small} and each coefficient \begin{small}$\lambda_j\!\sim\! \mathrm{Beta}(\alpha, \beta)$\end{small}. Here, we assume \ying{that} the \ying{size of the} support set 
and \ying{that of} the query 
are equal, i.e., \begin{small}$K^s\!=\!K^q$\end{small}. If \begin{small}$K^s\!<\!K^q$\end{small}, for each 
sample in \ying{the} query set, we randomly select one sample from \yingicml{the} support set for 
\ying{mixup}. 
\yingicml{The similar sampling applies to \begin{small}$K^s\!>\!K^q$\end{small}.}
In Appendix B.1, we illustrate the Beta distribution \ying{in} 
both symmetric (i.e., \begin{small}$\alpha=\beta$\end{small}) and skewed shapes (i.e., \begin{small}$\alpha\neq \beta$\end{small}). 
\ying{Using the mixed batch by \ours},
\ying{we reformulate}
the 
outer\ying{-}loop optimization 
problem as,
\begin{equation}
\label{eq:metamix}
\small
\theta_0^{*}:=\min_{\theta_0}
\frac{1}{n_T}\sum_{i=1}^{n_T}
\mathbb{E}_{\pmb{\lambda}\sim \mathrm{Beta}(\alpha, \beta)}\mathbb{E}_{l\sim \mathcal{C}}[\mathcal{L}(f_{{\phi}_i^{L-l}}(\mathbf{X}_{i,l}^{mix}), \mathbf{Y}_i^{mix})]\ying{,}
\end{equation}
where $f_{\phi_i^{L-l}}$ represents the rest of layers after the mixed layer $l$.
\ying{\ours\ is flexible enough to be 
compatible with off-the-shelf gradient-based meta-learning algorithms, by replacing the query set 
with the mixed batch for meta-training.} Further, to verify the effectiveness of \ours, we exam\yingicml{ine} whether the criteria in Definition~\ref{def:task_augmentation} are met in the follows. 
\begin{cor}\label{cor:metamix}
Assume that \yingicml{the support set is sampled independently from the query set.} Then the following \yingicml{two} equations 
hold:
\begin{equation}
    \small
    \begin{aligned}
        &I(\hat{\mathbf{Y}}_i^{mix};(\mathbf{X}_i^{s}, \mathbf{Y}_i^{s})|\theta_0,\mathbf{X}_i^{mix}) - I(\hat{\mathbf{Y}}_i^q;(\mathbf{X}_i^s, \mathbf{Y}_i^s)|\theta_0,\mathbf{X}_i^q)\\=&  H(\hat{\mathbf{Y}}_i^s|\theta_0,\mathbf{X}_i^{s}) \geq 0;\\
        & I(\theta_0;\mathbf{X}_i^{mix}, \mathbf{Y}_i^{mix}|\mathbf{X}_i^q, \mathbf{Y}_i^q)= H(\theta_0) - H(\theta_0|\mathbf{X}_i^{s},\mathbf{Y}_i^s).
    \end{aligned}
\end{equation}
\end{cor}
\yingicml{The first criterion is easily satisfied -- \begin{small}$H(\hat{\mathbf{Y}}_i^s|\theta_0,\mathbf{X}_i^{s})$\end{small} hardly equals zero as $\theta_0$ unlikely fits the support set in meta-learning.}
The second criteri\yingicml{on} 
indicates that MetaMix contributes a novel task as long as the support set of the task being augmented 
is capable of reducing the uncertainty of the initialization $\theta_0$, which is often the case. We provide the detailed proof of Corollary~\ref{cor:metamix} in Appendix A.1.

\textbf{MetaMix enhanced with Channel Shuffle.}
In classification, the proposed \ours\ can be further enhanced by another task augmentation strategy \yingicml{named} 
Channel Shuffle (CF). 
\ying{Channel Shuffle aims}
to randomly 
\yingicml{replace}
a subset of channels 
\yingicml{through}
samples of each class by the corresponding ones in a different class. Assume \yingicml{that} the hidden representation \begin{small}$f_{\phi_i^l}(\mathbf{x}^{s(q)}_{i,j})$\end{small} of each sample consists of \begin{small}$p$\end{small} channels, i.e., \begin{small}$f_{\phi_i^l}(\mathbf{x}^{s(q)}_{i,j})=[f_{\phi_i^l}^{(1)}(\mathbf{x}^{s(q)}_{i,j});\ldots;f_{\phi_i^l}^{(p)}(\mathbf{x}^{s(q)}_{i,j})]$\end{small}. 
\yingicml{Provided with 1)} a pair of classes \begin{small}$c$\end{small} and \begin{small}$c'$\end{small} with corresponding sample sets \yingicml{(\begin{small}$\mathbf{X}^{s(q)}_{i;c},\mathbf{Y}^{s(q)}_{i;c}$\end{small}), \begin{small}$(\mathbf{X}^{s(q)}_{i;c'},\mathbf{Y}^{s(q)}_{i;c'}$\end{small})} 
, and 
\yingicml{2)}
a random variable \begin{small}$\mathbf{R}\yingicml{_{c,c'}}=\text{diag}(r_1,...,r_p)$\end{small} with \begin{small}$r_t\sim \text{Bernoulli}(\delta)$\end{small} and \begin{small}$\delta>0.5$\end{small} for \begin{small}$t\in[p]$\end{small}, the channel shuffle process is formulated as:
\begin{equation}
\small
\begin{aligned}
\label{eq:channel_shuffle}
\mathbf{X}^{s(q),cf}_{i;c}&=\pmb{\mathrm{R}}\yingicml{_{c,c'}} f_{\phi_i^l}(\mathbf{X}^{s(q)}_{i;c})+(\mathbf{I}-\pmb{\mathrm{R}}\yingicml{_{c,c'}})f_{\phi_i^l}(\mathbf{X}^{s(q)}_{i;c'}),\\\;\; \mathbf{Y}^{s(q),cf}_{i;c}&=\mathbf{Y}_{i;c}^{s(q)}.
\end{aligned}
\end{equation}
The channel shuffle strategy is then 
applied in both support and query sets, with 
\begin{small}$\mathbf{R}_\yingicml{_{c,c'}}$\end{small} 
shared \yingicml{between the two sets}. We denote the shuffled support set and query set as \begin{small}$(\mathbf{X}_i^{s,cf}, \mathbf{Y}_i^{s,cf})$\end{small} and \begin{small}$(\mathbf{X}_i^{q,cf}, \mathbf{Y}_i^{q,cf})$\end{small}, respectively.  Then, in the outer-loop, the channel shuffled samples will be integrated into the \ours\ and the Eqn.~\eqref{eq:metamix} is reformulated as:
\begin{equation}
\small
\begin{aligned}
\label{eq:metamix+cs}
&\mathbf{X}^{mmcf}_{i,l}=\pmb{\lambda} \mathbf{X}^{s,cf}_i+(\mathbf{I}-\pmb{\lambda})\mathbf{X}^{q,cf}_i,\\\;\; &\mathbf{Y}^{mmcf}_i=\pmb{\lambda}\mathbf{Y}_i^{s,cf}+(\mathbf{I}-\pmb{\lambda})\mathbf{Y}_i^{q,cf},
\end{aligned}
\end{equation}
We name \yingicml{the MetaMix enhanced with}
channel shuffle 
as MMCF.
\yingicml{In Appendix A.2, we prove that MMCF not only meets the first criterion in Definition~\ref{def:task_aug}, but also outperforms MetaMix regarding the second criterion.}
Taking MAML as an example, we show \ours\ and MMCF in Alg.~\ref{alg:metamix_training} and Appendix B.2, \yingicml{respectively}.
\begin{algorithm}[h]
    \caption{Meta-training Process of MAML-\ours}
    \label{alg:metamix_training}
    \begin{algorithmic}[1]
    \REQUIRE Task distribution $p(\mathcal{T})$; Learning rate $\mu$, $\eta$; Beta distribution parameters $\alpha$, $\beta$; MetaMix candidate layer set $\mathcal{C}$ 
    \STATE Randomly initialize parameter $\theta_0$
    \WHILE{not converge}
    \STATE Sample a batch of tasks \begin{small}$\{\mathcal{T}_i\}_{i=1}^{n}$\end{small}
    \FORALL{\begin{small}$\mathcal{T}_i$\end{small}}
    \STATE Sample support set \begin{small}$\mathcal{D}^{s}_i=\{(\mathbf{x}^{s}_{i,j},\mathbf{y}^{s}_{i,j})\}_{j=1}^{K^s}$\end{small} and query set  \begin{small}$\mathcal{D}^{q}_i=\{(\mathbf{x}^{q}_{i,j},\mathbf{y}^{q}_{i,j})\}_{j=1}^{K^q}$\end{small} from \begin{small}$\mathcal{T}_i$\end{small}
    \STATE Compute the task-specific parameter $\phi_i$ via the inner-loop gradient descent, i.e., \begin{small}$\phi_i=\theta_0-\mu\nabla_{\theta_0}\mathcal{L}(f_{\theta_0}(\mathbf{X}_i^s), \mathbf{Y}_i^s)$\end{small}
    \STATE Sample MetaMix parameter \begin{small}$\pmb{\lambda}\sim \mathrm{Beta}(\alpha, \beta)$\end{small} and mixed layer $l$ from $\mathcal{C}$
    \STATE Forward both support and query sets and mixed them at layer \begin{small}$l$\end{small} as: \begin{small}$\mathbf{X}^{mix}_{i,l}=\pmb{\lambda} f_{\phi_i^l}(\mathbf{X}^s_i)+(\mathbf{I}-\pmb{\lambda})f_{\phi_i^l}(\mathbf{X}^q_i)$\end{small}, \begin{small}$\mathbf{Y}^{mix}_i=\pmb{\lambda}\mathbf{Y}_i^s+(\mathbf{I}-\pmb{\lambda})\mathbf{Y}_i^q$\end{small}
    \STATE Continual forward \begin{small}$\mathbf{X}^{mix}_{i,l}$\end{small} to the rest of layers and compute the loss as \begin{small}$\mathcal{L}(f_{\phi_i^{L-l}}(\mathbf{X}_{i,l}^{mix}), \mathbf{Y}_{i}^{mix})$\end{small}
    \ENDFOR
    \STATE Update \begin{small}$\theta_0\leftarrow \theta_0 - \eta\frac{1}{n}\sum_{i=1}^{n}
    \mathbb{E}_{\pmb{\lambda}\sim \mathrm{Beta}(\alpha, \beta)}\mathbb{E}_{l\sim \mathcal{C}}[\mathcal{L}(f_{\phi_i^{L-l}}(\mathbf{X}_{i,l}^{mix}), \mathbf{Y}_i^{mix})]$\end{small}
    \ENDWHILE
    \end{algorithmic}
\end{algorithm}

%% file: theoretic_study.tex
\section{Theoretic Analysis}
In this section, we theoretically investigate how the proposed task augmentation methods improve generalization, by analyzing the following two-layer neural network model. For each task $\mathcal{T}_i$, we consider minimizing the squared loss \begin{small}$\mathcal{L}(f_{\phi_i}(\mathbf{X}_i),\mathbf{Y}_i)=(f_{\phi_i}(\mathbf{X}_i)-\mathbf{Y}_i)^2$\end{small} with $f_{\phi_i}$ modeled by
\begin{equation}
\small
f_{\phi_i}(\mathbf{X}_i)=\phi_i^{\top}\sigma(\mathbf{W}\mathbf{X}_i),
\end{equation}
where \begin{small}$\phi_i$\end{small} is the task adapted parameters and \begin{small}$\mathbf{W}$\end{small} is the global shared parameter. Note that, the formulation of function \begin{small}$f$\end{small} is the equivalent to ANIL~\cite{raghu2020rapid} under the two-layer neural network scenario, where only the head layer is adapted during the inner-loop. In the following, we will detail the analysis of \ours\ and \cf.

\textbf{Analysis of MetaMix.}
In the analysis of \ours, we consider a symmetric version of MetaMix algorithm for technical reasons. Empirically we find that this symmetric version and the proposed MetaMix algorithm generate mostly identical results (see Appendix C for details). Specifically, for each task $\mathcal T_i$, we denote \begin{small}$\mathbf Z_i=\{\mathbf{x}_{i,j}, \mathbf{y}_{i,j}\}_{j=1}^{K^m}=\{\mathbf{x}^s_{i,j}, \mathbf{y}^s_{i,j}\}_{j=1}^{K^s}\cup \{\mathbf{x}^q_{i,j}, \mathbf{y}^q_{i,j}\}_{j=1}^{K^q}$\end{small}, and \begin{small}$K^m=K^s+K^q$\end{small}. Further, we consider the following MetaMix algorithm trains the second layer parameter \begin{small}$\phi_i$\end{small} by minimizing the squared loss on the mixed batch of data \begin{small}$\mathbf{Z}_i^{mix}=\{\mathbf{x}^{mix}_{i,j}, \mathbf{y}_{i,j}^{mix}\}_{j=1}$\end{small}, where \begin{small}$\mathbf{Z}_i^{mix}$\end{small} is constructed as
\begin{equation}
\small
\begin{split}
\label{eq:metamix2}
&\mathbf{x}^{mix}_{i,j}={\lambda} \sigma(\mathbf{W}\mathbf{x}_{i,j})+(1-{\lambda})\sigma(\mathbf{W}\mathbf{x}_{i,j'}),\\ &\mathbf{y}^{mix}_{i,j}={\lambda}{\mathbf{y}}_{i,j}+({1}-{\lambda})\mathbf{y}_{i,j'},
 \end{split}
\end{equation}
where \begin{small}$j'$\end{small} is a uniform sample from \begin{small}$[K^m]$\end{small} and \begin{small}$\lambda\sim \mathrm{Beta}(\alpha,\beta)$\end{small}. 

Extending the analysis in \cite{zhang2020does}, we have the following theorem on the second-order approximation of \begin{small}$\mathcal{L}(\mathbf{Z}_i^{mix})$\end{small}.  
\begin{lemma}~\label{thm:mixup-reg}
Consider the model set-up described above with mixup distribution \begin{small}$\lambda\sim \mathrm{Beta}(\alpha,\beta)$\end{small}. Then the second-order approximation of \begin{small}$\mathcal{L}(\mathbf{Z}_i^{mix})$\end{small} is given by  
\begin{equation}
\small
\label{eq:loss}
\mathcal{L}(\mathbf{Z}_i)+c\cdot \phi_i^\top(\frac{1}{K^m}\sum_{j=1}^{K^m}\sigma(\mathbf{W}\mathbf{x}_{i,j})\sigma(\mathbf{W} \mathbf{x}_{i,j} )^\top)\phi_i,
\end{equation}
where \begin{small}$c=\mathbb E_{\mathcal{D}_\lambda}[\frac{(1-\lambda)^2}{2\lambda^2}]$\end{small} with \begin{small}$\mathcal{D}_\lambda\sim\frac{\alpha}{\alpha+\beta}\mathrm{Beta}(\alpha+1,\beta)+\frac{\beta}{\alpha+\beta}\mathrm{Beta}(\beta+1,\alpha)$\end{small}.
\end{lemma}
This result suggests that the MetaMix algorithm is imposing a quadratic regularization on $\phi_i$ for the $i$-th task, and therefore reduces the complexity of the solution space and leads to a better generalization. 

To quantify the improvement of the generalization, let us denote the population meta-risk by 
\begin{equation}
\small
\mathcal{R}=\mathbb{E}_{\mathcal{T}_i\sim p(\mathcal{T})}\mathbb{E}_{(\mathbf{X}_i,\mathbf{Y}_i)\sim T_i}[\mathcal{L}(f_{\phi_i}(\mathbf{X}_i), \mathbf{Y}_i)],
\end{equation}
and the empirical version by

\begin{equation}
\small
\begin{split}
    \mathcal{R}(\{\mathbf{Z}_i\}_{i=1}^{n_T})=&\frac{1}{n_T}\sum_{i=1}^{n_T}\frac{1}{K^m}\sum_{j=1}^{K^m} \mathcal{L}(f_{\phi_i}(\mathbf{x}_{i,j}), \mathbf{y}_{i,j})\\=&\mathbb{E}_{\mathcal{T}_i\sim\hat p(\mathcal{T})}\mathbb{E}_{(\mathbf{X}_i,\mathbf{Y}_i)\sim\hat p(\mathcal{T}_i)}\mathcal{L}(f_{\phi_i}(\mathbf{X}_i), \mathbf{Y}_i).
\end{split}
\end{equation}
According to Theorem~\ref{thm:mixup-reg}, we study the generalization problem by considering the following function class that is closely related to the dual problem of Eqn.~\eqref{eq:loss}
\begin{equation}
\small
    \mathcal{F}_{\mathcal{T}}=\{\phi^\top \sigma(\mathbf{W}\mathbf{X}): \phi^\top \Sigma_{\sigma,\mathcal{T}}\phi\le\gamma \},
\end{equation}
where \begin{small}$\Sigma_{\sigma, \mathcal{T}}=\mathbb{E}_\mathcal{T}[\sigma(\mathbf W\mathbf X )\sigma(\mathbf W\mathbf X )^\top]$\end{small}. Notation-wise, let us also define  \begin{small}$\mu_{\sigma,\mathcal{T}}=\mathbb{E}_\mathcal{T}[\sigma(\mathbf{W}\mathbf{X})]$\end{small}. Further, we also assume the condition of the task distribution \begin{small}$\mathcal{T}$\end{small}: for all \begin{small}$\mathcal T\sim p(\mathcal T)$\end{small}, \begin{small}$\mathcal{T}$\end{small} satisfies 
\begin{equation}
\small
\label{assump}
rank(\Sigma_{\sigma,\mathcal{T}})\le r, \|\Sigma_{\sigma,\mathcal{T}}^{\mathbf{W}\dagger/2}\mu_{\sigma,\mathcal{T}}\|\le B,
\end{equation}
where \begin{small}$p(\mathcal T)$\end{small} is the distribution of the task distribution.

We then have the following theorem showing the improvement on the meta-generalization gap induced by the MetaMix algorithm. 
\begin{Theorem}
Suppose \begin{small}$\mathbf X$\end{small}, \begin{small}$\mathbf Y$\end{small} and \begin{small}$\phi$\end{small} are all bounded, and also assume assumption Eqn.~\eqref{assump} holds. Then there exists constants \begin{small}$C_1, C_2, C_3, C_4 > 0$\end{small}, such that for all \begin{small}$f_\mathcal T\in\mathcal{F}_{\mathcal T}$\end{small}, we have, with probability at least \begin{small}$1 -\delta$\end{small}, 
\begin{equation}
\small
\begin{split}
 |\mathcal{R}(\{\mathbf{Z}_i\}_{i=1}^{n_T})-\mathcal{R}|\le &C_1\sqrt{\frac{\gamma\cdot(r+B)}{K^m}}+C_2\sqrt\frac{\log(n_T/\delta)}{K^m}\\+&C_3\sqrt{\frac{\gamma\cdot B+1}{n_T}}+C_4\sqrt\frac{\log(1/\delta)}{n_T}.
\end{split}
\end{equation}
\end{Theorem}

According to Lemma 1, Mixup is regularizing $\phi^\top\Sigma_{\sigma,T}\phi$ and making $\gamma$ small. With this interpretation, Theorem 2 then suggests that a smaller value of $\gamma$ induced by Mixup will help reduce the generalization error, and therefore mitigate the overfitting. 

\textbf{Analysis of Channel Shuffle.}
We then analyze the channel shuffle strategy under the same two-layer neural network model considered above, with binary class \begin{small}$\mathbf{y}_{i,j}\in\{0,1\}$\end{small}. Instead of applying the mixup on 
\begin{small}$\mathbf{Z}_i=\{\mathbf{x}_{i,j}, \mathbf{y}_{i,j}\}_{j=1}^{K^m}:=\{ \mathbf{x}_{i,j;0}, 0\}_{j=1}^{K^{m_0}}\cup\{\mathbf{x}_{i,j;1}, 1\}_{j=1}^{K^{m_1}}$\end{small}, we now apply the channel 
shuffle strategy. Specifically, we consider the shuffled data \begin{small}$\mathbf{Z}_i^{cf}=\{\mathbf{x}^{cf}_{i,j}, \mathbf{y}_{i,j}^{}\}_{j=1}^{K^m}=\{\mathbf{x}^{cf}_{i,j;0}, 0\}_{j=1}^{K^{m_0}}\cup\{\mathbf{x}^{cf}_{i,j;1}, 1\}_{j=1}^{K^{m_1}}$\end{small}. According to Eqn.~\eqref{eq:channel_shuffle}, \begin{small}$\{\mathbf{x}^{cf}_{i,j;k}\}$\end{small} (\begin{small}$k\in\{0,1\}$\end{small}) is constructed as 
\begin{equation}
\small
 \begin{aligned}
\label{eq:cf2}
\mathbf{x}^{cf}_{i,j;k}&=\frac{1}{\delta}\cdot\left(\mathbf{R} \sigma(\mathbf{W}\mathbf{x}_{i,j;k})+(\mathbf{I}-\mathbf{R})\sigma(\mathbf{W}\mathbf{x}_{i,j';1-k})\right)\\&\text{ for $j\in[K^{m_k}]$, $k\in\{0,1\}$}.
\end{aligned}
\end{equation}
Let us denote such randomness by \begin{small}$\xi$\end{small}. Recall that \begin{small}$\mathbf{R}=\text{diag}(r_1,...,r_p)$\end{small} with \begin{small}$r_t\sim \text{Bernoulli}(\delta)$\end{small}, the scaling \begin{small}$\frac{1}{\delta}$\end{small} is added for technical convenience. Since the last layer is linear, the scaling \begin{small}$\frac{1}{\delta}$\end{small} will not affect the training and prediction results. 

We now define \begin{small}$\mathcal{L}(\mathbf{Z}_i^{cf})=\frac{1}{K^m}\sum_{j=1}^{K^m}\mathcal{L}({\phi_i}^\top(\mathbf{x}^{cf}_{i,j}), \mathbf{y}^{}_{i,j})$\end{small}.  For a generic vector \begin{small}$\mathbf{v}\in\mathbb{R}^p$\end{small}, we denote \begin{small}$\mathbf{v}^{\circ 2}=(v_1^2,...,v_p^2)$\end{small} and \begin{small}$diag(\mathbf{v}^{\circ 2})=diag(v_1^2,...,v_p^2)$\end{small} as the diagonal matrix with diagonal elements \begin{small}$(v_1^2,...,v_p^2)$\end{small}. We then have the following theorem on the second-order approximation of \begin{small}$\mathcal{L}(\mathbf{Z}_i^{cf})$\end{small}.  
\begin{Theorem}~\label{thm:cs-reg}
Consider the model set-up described above and recall that $\xi$ is the randomness involved in the data argumentation. Assume the training data is preprocessed as \begin{small}$\frac{1}{K^{m_0}}\sum_{j=1}^{K^{m_0}}\sigma(\mathbf{W}\mathbf{x}_{i,j;0})=\frac{1}{K^{m_1}}\sum_{j=1}^{K^{m_1}}\sigma(\mathbf{W}\mathbf{x}_{i,j;1})=0$\end{small}. There exists a constant \begin{small}$c>0$\end{small}, such that  the second-order approximation of \begin{small}$\mathbb{E}_{\xi} \mathcal{L}(\mathbf{Z}_i^{cf})$\end{small} is given by  \begin{equation}\label{eq:loss_cf}
\small
\begin{aligned}
\mathcal{L}(\mathbf{Z}_i)&+\frac{1-\delta}{\delta}\phi_i^\top(\frac{1}{K^m}\sum_{j=1}^{K^m} diag( \sigma(\mathbf{W}\mathbf{x}_{i,j})^{\circ 2})\phi_i+\\
&+\frac{1-\delta}{\delta}\phi_i^\top(\frac{1}{K^{m_0}}\sum_{j=1}^{K^{m_0}}\sigma(\mathbf{W}\mathbf{x}_{i,j;0} )\sigma(\mathbf{W}\mathbf{x}_{i,j;0} )^\top\\ 
&+\frac{1}{K^{m_1}}\sum_{j=1}^{K^{m_1}}\sigma(\mathbf{W} \mathbf{x}_{i,j;1} )\sigma(\mathbf{W}\mathbf{x}_{i,j;1} )^\top) \phi_i.
\end{aligned}
\end{equation}
\end{Theorem}
\vspace{-0.3em}
Theorem~\ref{thm:cs-reg} suggests that the Channel Shuffle algorithm will also impose a quadratic data-adaptive regularization on $\phi_i$, and the second quadratic term resembles the one induced by MetaMix in Lemma~\ref{thm:mixup-reg}. As a result, it will make the $\gamma$ in Theorem~\ref{thm:cs-reg} smaller and further reduce the overfitting. We provide more details and the full proof about theoretical analysis in Appendix C.

%% file: relatedwork.tex
\section{Discussion with Related Work\ying{s}}
One 
\ying{influential}
line of meta-learning algorithms is learning a 
transferable metric
\ying{space between samples from previous} 
tasks~\cite{mishra2018simple,oreshkin2018tadam,snell2017prototypical,vinyals2016matching}, which classify
\ying{samples via lazy learning with the learned distance metric}
(e.g., Euclidean distance~\cite{snell2017prototypical}, cosine distance~\cite{vinyals2016matching}). 
However, 
\ying{their applications are limited to}
classification problems, 
\ying{being}
infeasible in other problems (e.g., regression). 
In this work, we focus on gradient-based meta-learning algorithms 
that
learn a well-generalized model initialization from meta-training
tasks~\cite{finn2017meta,finn2017model,finn2018probabilistic,flennerhag2020meta,grant2018recasting,lee2018gradient,li2017meta,park2019meta},
\ying{being agnostic to problems}. 
\ying{This initialization is adapted}
to each task via \ying{the} support set,
\ying{and in turn the initialization is updated by maximizing the generalization performance on the query set.} \ying{These}
approaches 
\ying{are at}
high risk of overfitting the meta-training tasks 
\ying{and generalizing poorly to meta-testing tasks.}

\ying{Common techniques}
\ying{increase}
the generalization capability
\ying{via}
regularizations
such as 
weight decay~\cite{krogh1992simple},
\ying{dropout}~\cite{gal2016dropout,srivastava2014dropout},
\ying{and}
incorporating 
noise~\cite{achille2018information,alemi2017deep,tishby2015deep}. 
However, the adapted model by only a few steps on the support set in the inner-loop likely performs poorly on the query set. 
To improve such generalization for better adaptation,  either the number of parameters to adapt is reduced~\cite{raghu2020rapid,zintgraf2019fast,oh2021boil} or adpative noise is added~\cite{lee2020meta}.
The contribution of addressing this inner-loop overfitting towards meta-regularization, though positive, is limited.

Until very recently, two regularizers were proposed to specifically improve meta-generalization, including MR-MAML~\cite{yin2020meta} which regularizes the search space of the initialization while meanwhile allows it to be sufficiently adapted in the inner-loop, and TAML~\cite{jamal2019task} enforcing the initialization to behave similarly across tasks. \ying{Instead of imposing regularizers on the initialization},~\citet{rajendran2020meta} 
\yingicml{proposed to}
inject 
a random constant noise to 
label\yingicml{s} of both support and query sets. The shared noise,
however, 
is easy to be learned in the inner-loop.
\yingicml{Besides, as we prove in Appendix A.3, this augmentation}
fails to meet the second criterion in Definition~\ref{def:task_aug} 
and therefore little additional information
\yingicml{is} 
provided to meta-train the initialization.
Our work takes sufficiently powerful ways actively soliciting more data to meta-train the initialization.
Note that our 
\yingicml{task}
augmentation strategies are more than just a simple application of conventional data augmentation strategies~\cite{cubuk2019autoaugment,verma2019manifold,zhang2018mixup}, which 
\yingicml{have} been proved in both~\cite{lee2020meta} and our  experiments to have a very limited role.
We initiate to 
\yingicml{include}
more query data that satisfy the proposed Criterion 1 in the meta-training phase, \yingicml{so that} 
the dependence 
\yingicml{on}
support sets \yingicml{during inner-loop adaptation is increased}
and \yingicml{the meta-generalization is improved.}

%% file: experiment.tex
\section{Experiments}
\label{sec:exp}
To show the effectiveness of \ours, we conduct experiments on three meta-learning problems, namely: (1) drug activity prediction\ying{,} 
(2) pose prediction\ying{,} 
\ying{and}
(3) image classification. We apply \ours\ on four gradient-based meta-learning algorithms, including MAML~\citep{finn2017model}, MetaSGD~\citep{li2017meta}, T-Net~\citep{lee2018gradient}, and ANIL~\citep{raghu2020rapid}. For comparison, we consider 
\ying{the following regularizers: Weight Decay as the traditional regularizer, CAVIA~\citep{zintgraf2019fast} and Meta-dropout~\citep{lee2020meta} which regularize the inner-loop, and 
MR-MAML~\citep{yin2020meta}, TAML~\citep{jamal2019task}, and Meta-Aug~\cite{rajendran2020meta}, all of which handle meta-generalization.}
\subsection{Drug Activity Prediction}
\textbf{Experimental \ying{Setup}.}
\ying{
We solve a real-world application of drug activity prediction~\citep{martin2019all} where there are 4,276 target assays (i.e., tasks) each of which consists of a few  drug compounds with tested activities against the target protein.} We randomly selected 100 assays for meta-testing, 76 for meta-validation and the rest for meta-training. We repeat the random process four times and 
construct four groups of meta-testing assays for evaluation. 
\ying{Following~\citep{martin2019all}, we evaluate}
the 
\ying{square of Pearson coefficient}
\begin{small}$R^2$\end{small} between \ying{the} predicted 
\begin{small}$\hat{\mathbf{y}}_i^q$\end{small} and the 
\ying{groundtruth}
\begin{small}$\mathbf{y}_i^q$\end{small} of all query samples
\ying{for each $i$-th 
task, and report
the mean and median \begin{small}$R^2$\end{small} values over all meta-testing assays as well as the number of assays with \begin{small}$R^2>0.3$\end{small} which is deemed as an indicator of reliability in pharmacology.}
We use \ying{a base model of} two fully connected layers with 500 hidden \ying{units}. 
In \begin{small}$\mathrm{Beta}(\alpha, \beta)$\end{small}, we set \begin{small}$\alpha=\beta\ying{=0.5}$\end{small}.
More details 
\ying{on the dataset and settings}
 are discussed in Appendix D.1.
\begin{table*}[t]
\small
\caption{Performance of drug activity prediction.}
\label{tab:drug_results}
\begin{center}
\setlength{\tabcolsep}{1mm}{
\begin{tabular}{l|ccc|ccc|ccc|ccc}
\toprule
\multirow{2}{*}{Model}  & \multicolumn{3}{c|}{Group 1} & \multicolumn{3}{c|}{Group 2} & \multicolumn{3}{c|}{Group 3} & \multicolumn{3}{c}{Group 4}\\
 & Mean & Med. & \begin{small}$>$\end{small}0.3 & Mean & Med. & \begin{small}$>$\end{small}0.3 & Mean & Med. & \begin{small}$>$\end{small}0.3 & Mean & Med. & \begin{small}$>$\end{small}0.3 \\\midrule
\ying{pQSAR-max~\citep{martin2019all}}  & $0.390$ & $0.335$ & $51$ & $0.335$ & $0.280$ & $44$ & $0.373$ & $0.315$ & $50$ & $0.362$ & $0.260$ & $46$\\\midrule
Weight Decay   & $0.307$ & $0.228$ & $40$ & $0.243$ & $0.157$ & $34$ & $0.259$ & $0.171$ & $38$ & $0.290$ & $0.241$ & $47$\\
CAVIA   & $0.300$ & $0.232$ & $42$ & $0.234$ & $0.132$ & $35$ & $0.260$ & $0.184$ & $39$ & $0.317$ & $0.292$ & $46$\\
Meta-dropout   & $0.319$ & $0.203$ & $41$ & $0.250$ & $0.172$ & $35$  & $0.281$ & $0.214$ & $39$ & $0.316$ & $0.275$ & $47$\\
Meta-Aug & $0.317$ & $0.201$ & $43$ & $0.253$ & $0.193$ & $38$ & $0.286$ & $0.220$ & $41$ & $0.303$ & $0.224$ & $42$ \\
MR-ANIL  & $0.297$ & $0.202$ & $41$ & $0.232$ & $0.152$ & $32$ & $0.289$ & $0.217$ & $40$ & $0.293$ & $0.249$ & $43$\\
TAML   & $0.296$ & $0.200$ & $41$ & $0.260$ & $0.203$ & $36$ & $0.260$ & $0.227$ & $40$ & $0.308$ & $0.281$ & $46$\\\midrule
MetaSGD   & $0.331$ & $0.224$ & $45$ & $0.249$ & $0.187$ & $33$ & $0.282$ & $0.226$ & $40$ & $0.312$ & $0.287$ & $48$\\
T-Net & $0.323$ & $0.264$ & $46$ & $0.236$ & $0.170$ & $29$ & $0.285$ & $0.220$ & $43$ & $0.285$ & $0.239$ & $42$\\
ANIL   & $0.299$ & $0.184$ & $41$ & $0.226$ & $0.143$ & $30$ & $0.268$ & $0.199$ & $37$ & $0.304$ & $0.282$ & $48$ \\
ANIL++   & $0.367$ & $0.299$ & $50$ & $0.315$ & $0.252$ & $43$ & $0.335$ & $0.289$ & $48$ & $0.362$ & $0.324$ & $51$\\\midrule
\textbf{MetaSGD-\ours}   & $0.364$ & $0.296$ & $49$ & $0.271$ & $0.230$ & $45$ & $0.312$ & $0.267$ & $48$ & $0.338$ & $0.319$ & $51$\\
\textbf{T-Net-\ours}  & $0.352$ & $0.291$ & $50$ & $0.276$ & $0.229$ & $42$ & $0.310$ & $0.285$ & $47$ & $0.336$ & $0.298$ & $50$\\
\textbf{ANIL-\ours}   & $0.347$ & $0.292$ & $49$ &  $0.301$ & $0.282$  & $47$ & $0.302$ & $0.258$ & $45$ & $0.348$ & $0.303$ & $51$\\
\textbf{ANIL++-\ours}    & $\mathbf{0.413}$ & $\mathbf{0.393}$ & $\mathbf{59}$ & $\mathbf{0.337}$ & $\mathbf{0.301}$ & $\mathbf{51}$ & $\mathbf{0.381}$ & $\mathbf{0.362}$ & $\mathbf{55}$ & $\mathbf{0.380}$ & $\mathbf{0.348}$ & $\mathbf{55}$\\\bottomrule
\end{tabular}}
\end{center}
\vspace{-1.5em}
\end{table*}

\textbf{Performance.}
In practice, we notice that 
\ying{only updating the final layer in the inner-loop}
achieves the best performance, which is equivalent to 
ANIL. Thus, we apply \ying{this} 
inner-loop update strategy to all baselines. 
\ying{For stability, here we also use}
ANIL++~\citep{antoniou2019train} 
\ying{which stabilizes ANIL} for comparison.
In Table~\ref{tab:drug_results}, we
\ying{compare \ours\ with the}
baselines \ying{on the} 
four drug evaluation groups. We observe that \ours\ consistently improve\ying{s} the performance despite of the backbone meta-learning algorithms (i.e., ANIL, ANIL++, MetaSGD, T-Net) in all scenarios. In addition, 
\ying{ANIL-MetaMix}
outperforms other anti-overfitting strategies. Particularly, compared to Meta-Aug, the better performance of ANIL-\ours\ indicates that additional information 
provided by MetaMix benefits the meta-generalization. In summary, the consistent superior performance\ying{, even significantly better than the state-of-the-art pQSAR-max for this dataset}, demonstrates that (1) \ours\ is compatible with 
\ying{existing}
meta-learning algorithms; (2) \ours\ is capable of improving \ying{the} meta-generalization ability. 
Besides, in Appendix E.1, we investigate the influence of different hyperparameter settings (e.g., \begin{small}$\alpha$\end{small} in \begin{small}$\mathrm{Beta}(\alpha, \alpha)$\end{small}), 
\ying{and demonstrate the}
robustness of \ours\ under
different settings.

\begin{figure*}[h]
	\centering
	\begin{subfigure}[c]{0.23\textwidth}
		\centering
		\includegraphics[width=\textwidth]{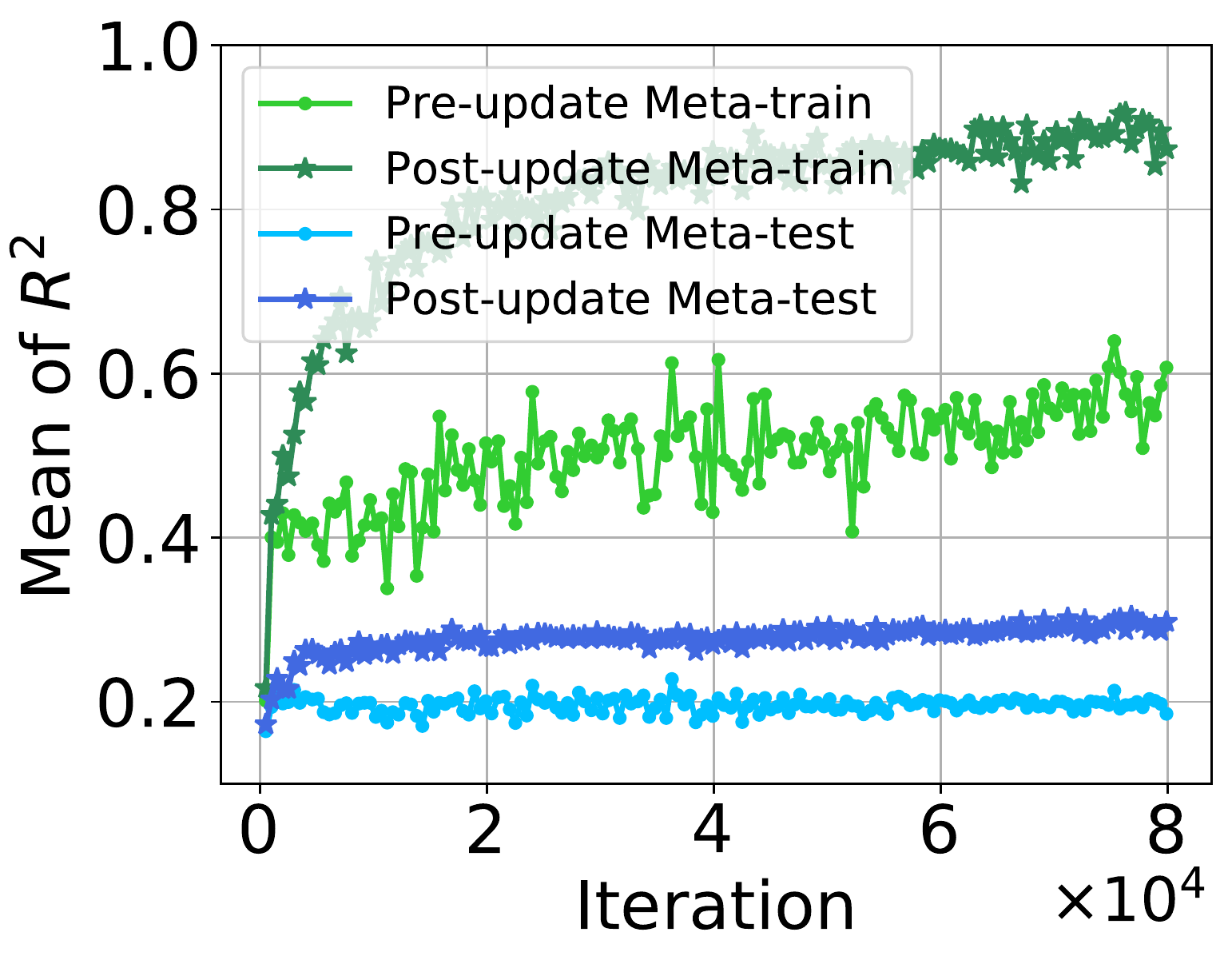}
		\caption{\label{fig:curve_anil}: ANIL}
	\end{subfigure}
    \begin{subfigure}[c]{0.23\textwidth}
		\centering
		\includegraphics[width=\textwidth]{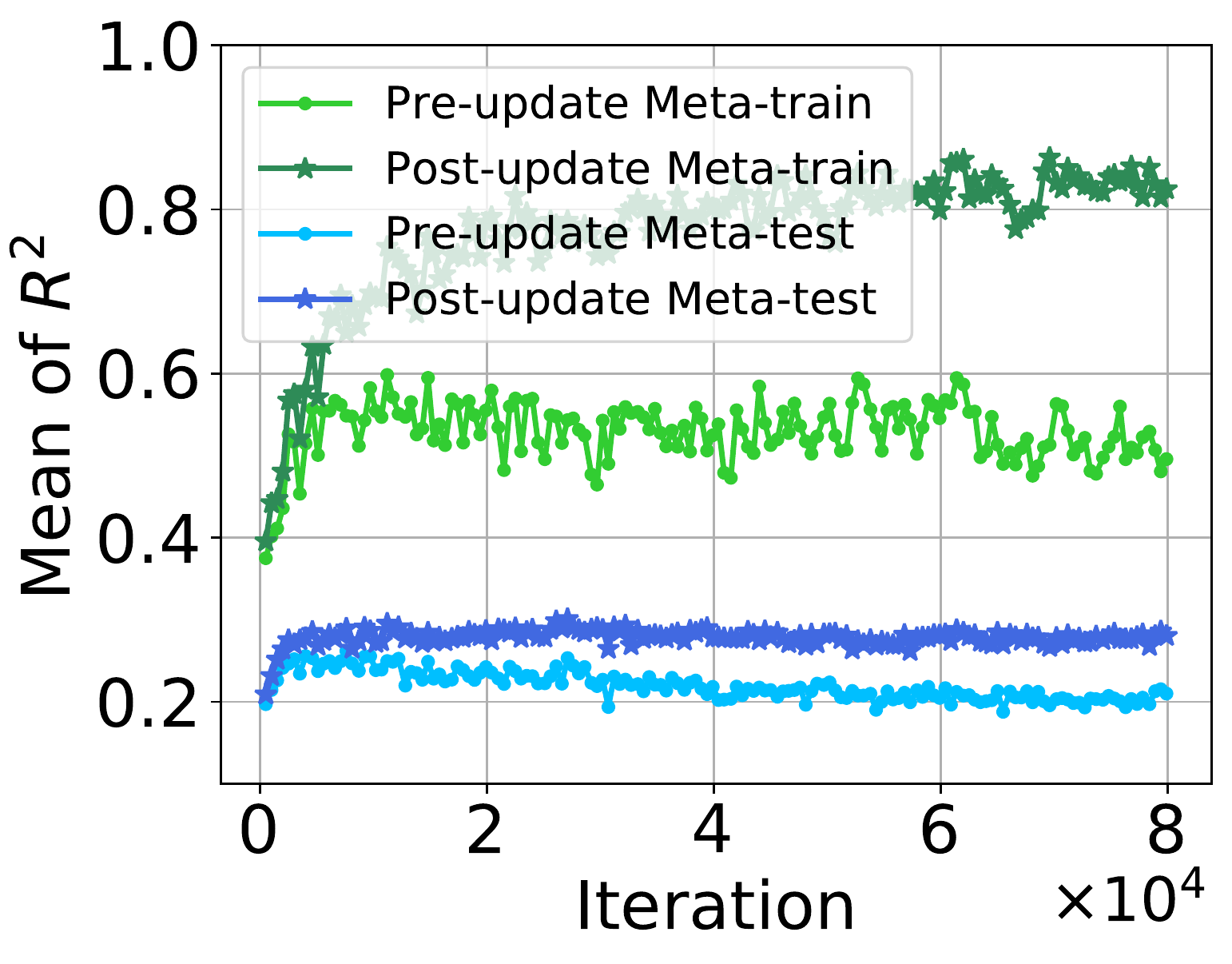}
		\caption{\label{fig:curve_mrmaml}: MR-ANIL}
	\end{subfigure}
	\begin{subfigure}[c]{0.23\textwidth}
		\centering
		\includegraphics[width=\textwidth]{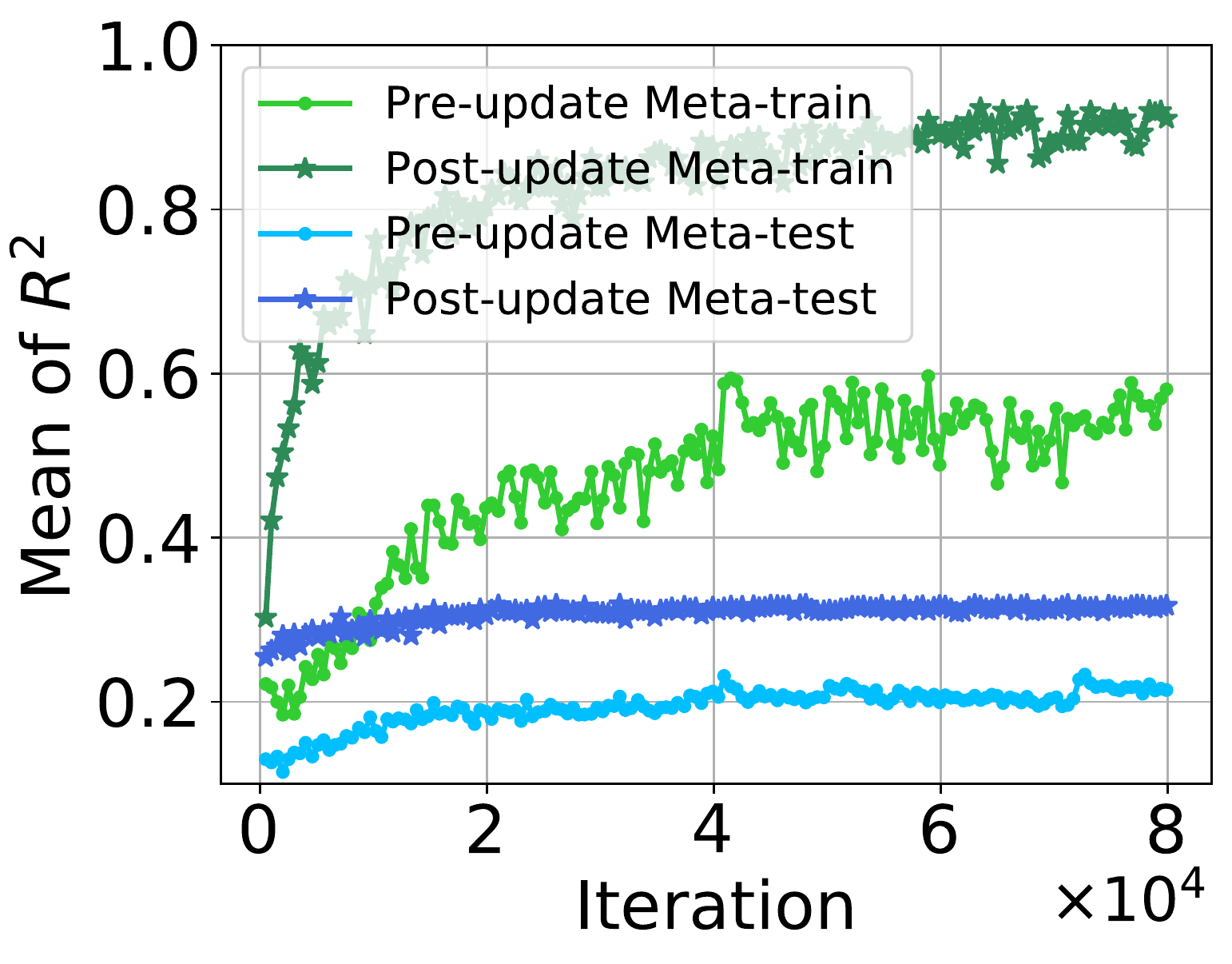}
		\caption{\label{fig:curve_metaaug}: Meta-Aug}
	\end{subfigure}
	\begin{subfigure}[c]{0.23\textwidth}
		\centering
		\includegraphics[width=\textwidth]{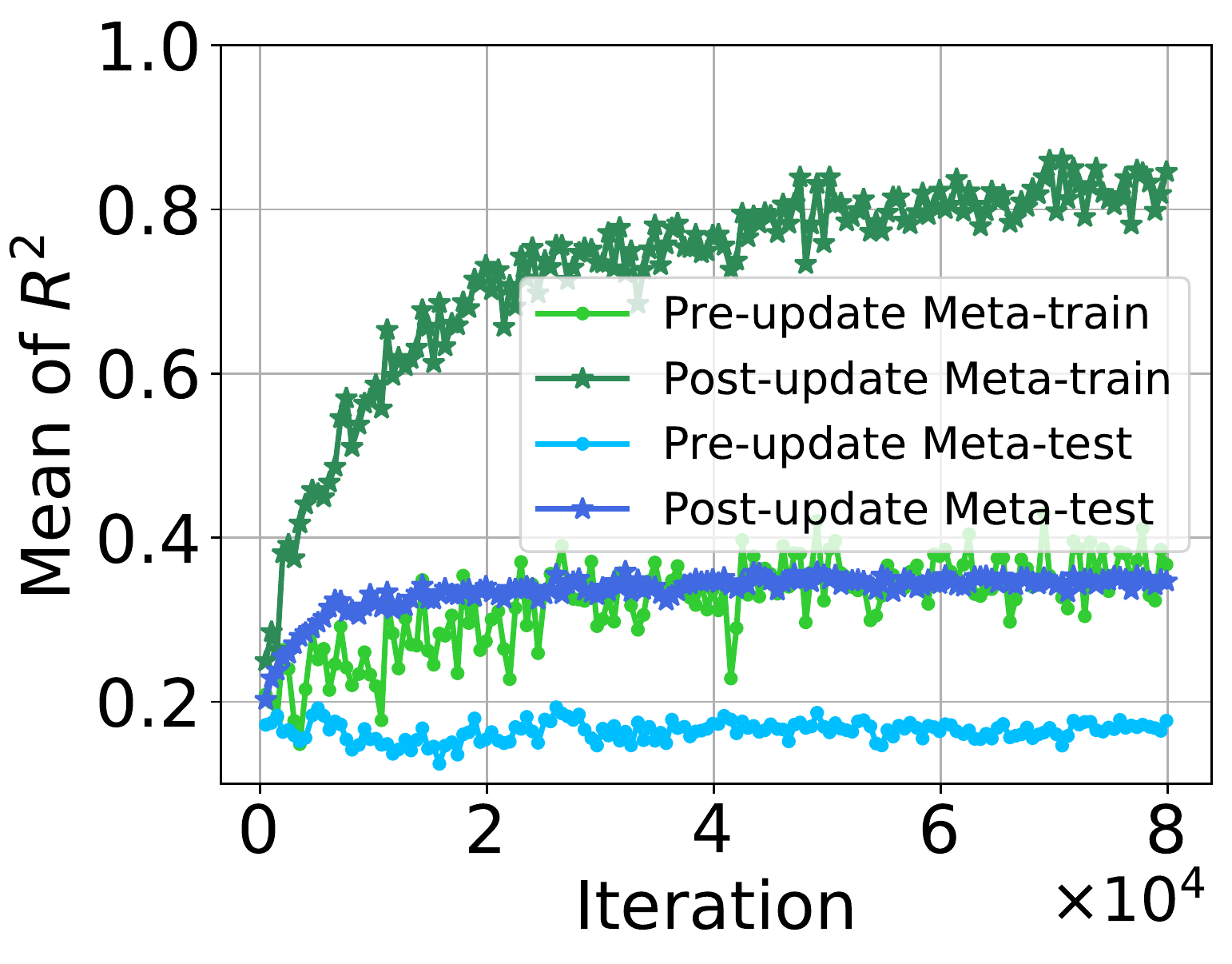}
		\caption{\label{fig:curve_anilmm}: ANIL-\ours}
	\end{subfigure}
	\vspace{-0.5em}
	\caption{Overfitting analysis on Group 1 of drug activity prediction. All models use the same inner-loop update strategy as ANIL.}
	\vspace{-1em}
	\label{fig:overfit_analysis}
\end{figure*}

\textbf{Analysis of Overfitting.} In Figure~\ref{fig:overfit_analysis}, we visualize the meta-training and meta-testing performance of ANIL, ANIL-\ours\ and other two representative anti-overfitting strategies (i.e., MR-ANIL, Meta-Aug) with respect to the training iteration. Interestingly, we find (1) in the meta-testing phase, applying \ours\ significantly increases the performance gap between pre-update ($\theta_0$) and post-update ($\phi_i$), indicating that \ours\ improves the dependence of target prediction on support sets, and therefore alleviates memorization overfitting; (2) compared to Meta-Aug and MR-ANIL, the worse pre-update meta-training performance but better post-update meta-testing performance of \ours\ demonstrates its superiority to mitigate the learner overfitting.

\textbf{Effect of Data Mixture Strategy in \ours.} To further investigate where \ying{the} improvement 
stem\ying{s} from, we 
\ying{adopt}
five different 
\ying{mixup} strategies for meta-training. 
The 
results are reported in Table~\ref{tab:drug_mixture_strategy}. We use Mixup(\begin{small}$\mathcal{D}^m$\end{small}, \begin{small}$\mathcal{D}^n$\end{small}) to denote the mixup 
of data \begin{small}$\mathcal{D}^m$\end{small} and \begin{small}$\mathcal{D}^n$\end{small} (e.g., Mixup(\begin{small}$\mathcal{D}^s$\end{small}, \begin{small}$\mathcal{D}^q$\end{small}) \ying{in our case}). \begin{small}$\mathcal{D}^{cob}\!=\!\mathcal{D}^{s}\!\oplus\!\mathcal{D}^{q}$\end{small} represents the  concatenat\ying{ion}
of \begin{small}$\mathcal{D}^{s}$\end{small} and \begin{small}$\mathcal{D}^q$\end{small}. \yaocr{In drug activity prediction, since the support and query sets are pre-split based on the biological domain knowledge, we also introduce set shuffle as another ablation model by randomly shuffling the pre-split sets.} In Table~\ref{tab:drug_mixture_strategy}, we find that (1) \ours\ achieves the best performance compared with other ablation models; (2) \ying{the fact that \ours\ enjoys better performance than}
Mixup(\begin{small}$\mathcal{D}^
q, \mathcal{D}^q$\end{small}) suggests that \ours\ 
is much more than simple data augmentation -- it increases the dependency of the learner on 
support sets and thereby minimizes the memorization; (3) involving \ying{the} support set \ying{only} is insufficient \ying{for }
meta-generalization 
\ying{due to}
its relative small gradient norm, which is \ying{further} verified by the \ying{unsatisfactory} 
performance of \begin{small}$\mathcal{D}^s\oplus \mathcal{D}^q$\end{small} compared with \ours\ .
\begin{table*}[h]
\caption{Effect of mixture strategies on drug activity prediction. All strategies are applied on ANIL++.}
\label{tab:drug_mixture_strategy}
\begin{center}
\small
\setlength{\tabcolsep}{0.8mm}{
\begin{tabular}{l|ccc|ccc|ccc|ccc}
\toprule
\multirow{2}{*}{Strategies}  & \multicolumn{3}{c|}{Group 1} & \multicolumn{3}{c|}{Group 2} & \multicolumn{3}{c|}{Group 3} & \multicolumn{3}{c}{Group 4}\\
  & Mean & Med. & \begin{small}$>$\end{small}0.3 & Mean & Med. & \begin{small}$>$\end{small}0.3 & Mean & Med. & \begin{small}$>$\end{small}0.3 & Mean & Med. & \begin{small}$>$\end{small}0.3 \\\midrule
\begin{small}$\mathcal{D}^q$\end{small} & $0.367$ & $0.299$ & $50$ & $0.315$ & $0.252$ & $43$  & $0.335$ & $0.289$ & $48$ & $0.362$ & $0.324$ & $51$\\
Set Shuffle & $0.371$ & $0.352$ & $55$ & $0.293$ & $0.224$ & $42$ & $0.339$ & $0.297$ & $50$ & $0.360$ & $0.300$ & $50$\\
Mixup(\begin{small}$\mathcal{D}^s$\end{small}, \begin{small}$\mathcal{D}^s$\end{small})   & $0.224$  & $0.164$ & $33$ & $0.210$ & $0.164$ & $31$ & $0.214$ & $0.154$ & $29$ & $0.191$ & $0.141$ & $22$\\
Mixup(\begin{small}$\mathcal{D}^q$\end{small}, \begin{small}$\mathcal{D}^q$\end{small})   & $0.388$ & $0.354$  & $55$ & $0.322$ & $0.264$ & $46$ & $0.341$ & $0.306$ & $50$ & $0.358$ & $0.325$ & $53$ \\
\begin{small}$\mathcal{D}^{cob}=\mathcal{D}^s\oplus \mathcal{D}^q$\end{small}   & $0.376$ & $0.324$ & $52$ & $0.301$ & $0.242$ & $44$ & $0.333$ & $0.329$ &  $51$& $0.336$ & $0.281$ & $48$\\\midrule
\textbf{\ours}   & $\mathbf{0.413}$ & $\mathbf{0.393}$ & $\mathbf{59}$ & $\mathbf{0.337}$ & $\mathbf{0.301}$ & $\mathbf{51}$ & $\mathbf{0.381}$ & $\mathbf{0.362}$ & $\mathbf{55}$ & $\mathbf{0.380}$ & $\mathbf{0.348}$ & $\mathbf{55}$\\\bottomrule
\end{tabular}}
\end{center}
\vspace{-0.8em}
\end{table*}

\yaocr{\textbf{Analysis of Criteria.}
We further analyze augmentation methods on drug data (Group 1) with respect to the two criteria (C1, C2) we propose and the CE-increasing criterion \begin{small}H(Y$|$X)$\uparrow$\end{small} proposed by Meta-Aug. We report the results in Table~\ref{tab:criteria_analysis}, where Mix-all applies Mixup to the whole dataset without 
differentiating different tasks.
We observe that C1 and C2 are qualified to guide the design of task augmentation methods, as evidenced in Table~\ref{tab:criteria_analysis} where methods satisfying more 
of C1 and C2 tend to perform better.
}
\begin{table}[h]
	\centering
	\caption{Criteria analysis on Group 1 of drug activity prediction. All models use ANIL as the backbone meta-learning algorithm.}
	\small
        \begin{tabular}{l|c|c|c|c}
        \toprule
        Aug. Method & C1 & C2 & H(Y$|$X)$\uparrow$ & Mean $R^2$ \\\midrule
        Mix-All &  &  & & 0.292 \\
        Mixup(\begin{small}$\mathcal{D}^{q}$\end{small}, \begin{small}$\mathcal{D}^{q}$\end{small}) &  & $\surd$ & $\surd$ & 0.322 \\
        Meta-Aug & $\surd$ &  & $\surd$ & 0.317 \\\midrule
        ANIL-MetaMix & $\surd$ & $\surd$ & $\surd$ & \textbf{0.347} \\\bottomrule
        \end{tabular}
	\label{tab:criteria_analysis}
	\vspace{-1.5em}
\end{table}

\subsection{Pose Prediction}
\textbf{Experimental \ying{Setup}.}
Following~\citep{yin2020meta}, we use the regression dataset created from Pascal 3D data~\citep{xiang2014beyond}, where 
a
$128\!\times\!128$ grey-scale image is used as input \ying{and the orientation relative to a fixed pose} \ying{labels each image.}
50 and 15 objects are randomly selected for meta-training and meta-testing, respectively. Following~\cite{yin2020meta}, the base model consists of 
an
encoder with three convolutional blocks and a decoder with four convolutional blocks. For \ours, we set \begin{small}$\alpha\!=\!\beta\!=\!0.5$\end{small} in \begin{small}$\mathrm{Beta}(\alpha, \beta)$\end{small} and only perform Mainfold Mixup on the decoder (see Appendix D.2 for detailed 
settings). 
\begin{table}[h]
\begin{center}
\caption{Performance (MSE $\pm$ 95\% confidence interval) of pose prediction.}
\small
\label{tab:pose_results}
\begin{tabular}{l|cc}
\toprule
Model & 10-shot & 15-shot \\\midrule
Weight Decay & \yao{$2.772\pm0.259$} & \yao{$2.307\pm0.226$}\\
CAVIA & \yao{$3.021\pm0.248$} & \yao{$2.397\pm0.191$} \\
Meta-dropout & \yao{$3.236\pm0.257$} & \yao{$2.425\pm0.209$}\\
Meta-Aug & \yao{$2.553 \pm 0.265$} & \yao{$2.152 \pm 0.227$}\\
MR-MAML & \yao{$2.907\pm0.255$} & \yao{$2.276\pm0.169$} \\
TAML & \yao{$2.785\pm0.261$} & \yao{$2.196\pm0.163$}\\\midrule
ANIL & \yao{$6.746 \pm 0.416$} & \yao{$ 6.513 \pm 0.384$} \\
MAML & \yao{$3.098\pm0.242$} & \yao{$2.413\pm0.177$} \\
MetaSGD & \yao{$2.803\pm0.239$} & \yao{$2.331\pm0.182$} \\
T-Net & \yao{$2.835\pm0.189$} & \yao{$2.609\pm0.213$} \\\midrule
\textbf{ANIL-\ours} & $\yao{6.354\pm0.393}$ & $\yao{6.112\pm0.381}$    \\
\textbf{MAML-\ours} & \yao{$2.438\pm 0.196$} &  \yao{$2.003\pm0.147$}\\
\textbf{MetaSGD-\ours} & \yao{$\mathbf{2.390\pm0.191}$} & \yao{$\mathbf{1.952\pm0.154}$} \\
\textbf{T-Net-\ours} & \yao{$2.563\pm0.201$} & \yao{$2.418\pm0.182$} \\\bottomrule
\end{tabular}
\end{center}
\vspace{-1.5em}
\end{table}

\textbf{Results.} Table~\ref{tab:pose_results} shows the performance (averaged MSE with 95\% confidence interval) of baselines and \ours\ under 10/15-shot scenarios. 
\ying{The inner-loop regularizers are not as effective as MR-MAML, TAML and Meta-Aug in improving meta-generalization; MAML-MetaMix and Meta-Aug significantly improve MR-MAML, suggesting the effectiveness of bringing more data in than imposing meta-regularizers only.} The better performance of MAML-\ours\ than Meta-Aug further verifies the effectiveness of introducing additional 
knowledge
to learn the initialization.
We also investigate the influence of 
mixup strategies and hyperparameters on pose prediction 
in Appendix F.1 and F.2, respectively. The results again advocate 
the effectiveness and robustness of the proposed mixup strategy in improving the meta-generalization ability. 
\begin{table*}[t]

\small
\caption{Performance (accuracy $\pm$ 95\% confidence interval) of image classification on Omniglot and MiniImagenet.}
\label{tab:homo_image_classification}
\begin{center}
\begin{tabular}{l|cc|cc}
\toprule
\multirow{2}{*}{Model}  & \multicolumn{2}{c|}{Omniglot} & \multicolumn{2}{c}{MiniImagenet}\\
 & 20-way 1-shot & 20-way 5-shot & 5-way 1-shot & 5-way 5-shot \\\midrule
Weight Decay & $86.81\pm0.64\%$ & $96.20\pm0.17\%$ & $33.19\pm1.76\%$ & $52.27\pm0.96\%$\\
CAVIA & $87.63\pm0.58\%$ & $94.16\pm0.20\%$ & $34.27\pm1.79\%$ & $50.23\pm0.98\%$ \\
MR-MAML & $89.28\pm0.59\%$ & $96.66\pm0.18\%$ & $35.00\pm1.60\%$ & $54.39\pm0.97\%$ \\
Meta-dropout & $85.60\pm0.63\%$ & $95.56\pm0.17\%$ & $34.32\pm1.78\%$ & $52.40\pm0.96\%$ \\
TAML & $87.50\pm0.63\%$ & $95.78\pm0.19\%$ & $33.16\pm1.68\%$ & $52.78\pm0.97\%$ \\\midrule
MAML & $87.40\pm0.59\%$ & $93.51\pm0.25\%$ & $32.93\pm1.70\%$ & $51.95\pm0.97\%$ \\
MetaSGD & $87.72\pm0.61\%$ & $95.52\pm0.18\%$ & $33.70\pm1.63\%$ & $52.14\pm0.92\%$  \\
T-Net & $87.71\pm0.62\%$ & $95.67\pm0.20\%$ & $33.73\pm1.72\%$ &  $54.04\pm0.99\%$ \\
ANIL & $88.35\pm0.56\%$ & $95.85\pm0.19\%$ & $34.13\pm1.67 \%$ & $52.59\pm0.96\%$ \\\midrule
\textbf{MAML-MMCF} & $92.06 \pm 0.51\%$ & $97.95 \pm 0.17\%$ & $39.26 \pm 1.79\%$ & $58.96 \pm 0.95\%$ \\
\textbf{MetaSGD-MMCF} & $\mathbf{93.59 \pm 0.45\%}$ & $98.24 \pm 0.16\%$ & $\mathbf{40.06 \pm 1.76\%}$ & $\mathbf{60.19 \pm 0.96\%}$ \\
\textbf{T-Net-MMCF} & $93.27 \pm 0.46\%$ & $98.09\pm 0.15\%$ & $38.33 \pm 1.73\%$ & $59.13 \pm 0.99\%$ \\
\textbf{ANIL-MMCF} & $92.24 \pm 0.48\%$ & $\mathbf{98.36 \pm 0.13\%}$ & $37.94 \pm 1.75\%$ & $59.03 \pm 0.93\%$  \\
\bottomrule
\end{tabular}
\end{center}
\vspace{-0.5em}
\end{table*}
\begin{table*}[h]
\small
\centering
\caption{Performance (accuracy $\pm$ 95\% confidence interval) of MiniImagenet and Omniglot w.r.t. different data augmentation strategies applied on MAML.}
\label{tab:miniimagenet_mixture_strategy}
\begin{tabular}{l|cc|cc}
\toprule
\multirow{2}{*}{Strategy}   &  \multicolumn{2}{c}{Omniglot} & \multicolumn{2}{c}{MiniImagenet}\\
 & 20-way 1-shot & 20-way 5-shot  & 5-way 1-shot & 5-way 5-shot \\\midrule
\begin{small}$\mathcal{D}^q$\end{small} & $87.40\pm0.59\%$ & $93.51\pm0.25\%$ & $32.93\pm1.70\%$ & $51.95\pm0.97\%$\\
Mixup(\begin{small}$\mathcal{D}^s$\end{small}, \begin{small}$\mathcal{D}^s$\end{small}) & $46.98\pm0.92\%$ & $85.56\pm0.28\%$ & $24.39\pm1.48\%$ & $33.18\pm0.82\%$\\
Mixup(\begin{small}$\mathcal{D}^q$\end{small}, \begin{small}$\mathcal{D}^q$\end{small}) & $90.65\pm0.56\%$ & $96.90\pm0.16\%$  & $34.56\pm1.77\%$ & $55.80\pm0.97\%$\\
\begin{small}$\mathcal{D}^{cob}=\mathcal{D}^{s}\oplus\mathcal{D}^{q}$\end{small} & $86.74\pm0.54\%$ & $95.54\pm0.19\%$ & $33.33\pm1.70\%$  & $51.97\pm0.96\%$ \\\midrule
\textbf{\ours}\ & $91.53\pm0.53\%$ & $97.63\pm0.15\%$ & $38.53\pm1.79\%$ &  $57.55\pm1.01\%$\\
\textbf{\cf}\ & $89.81 \pm 0.55\%$ & $97.10 \pm 0.17\%$ & $35.50 \pm 1.73\%$ & $54.52 \pm 0.96\%$ \\\midrule
\textbf{MMCF} & $\mathbf{92.06 \pm 0.51\%}$ & $\mathbf{97.95 \pm 0.17\%}$ & $\mathbf{39.26 \pm 1.79\%}$ & $\mathbf{58.96 \pm 0.95\%}$\\\bottomrule
\end{tabular}
\end{table*}

\subsection{Image Classification}
\textbf{Experimental \ying{Setup}.} For image classification problem\ying{s}, standard benchmarks (e.g., Omniglot~\citep{lake2011one} and MiniImagenet~\citep{vinyals2016matching}) are considered as mutually-exclusive tasks by introducing the shuffling mechanism of labels, which significantly alleviates the meta-overfitting issue~\citep{yin2020meta}. To show the power of proposed augmentation strategies, \ying{following~\citep{yin2020meta},}
we adopt the non-mutually-exclusive setting for each image classification benchmark: each class 
\ying{with}
its 
classification label 
remains unchanged across different meta-training tasks \ying{during meta-training}. Besides, we 
\ying{study image classification for heterogeneous tasks in Appendix G.1.}
\ying{We use the multi-dataset in~\citep{yao2019hierarchically}} which consists of four subdatasets, \ying{i.e.,} Bird, Texture, Aircraft, and Fungi. \ying{The} non-mutually-exclusive setting is also applied 
\ying{to this}
multi-dataset. Three representative
\ying{
heterogeneous meta-learning algorithms}
(i.e., MMAML~\citep{vuorio2019multimodal}, HSML~\citep{yao2019hierarchically}, ARML~\citep{yao2020automated}) are 
\ying{taken as baselines and applied with task augmentation stategies.}
For each task, \ying{the} classical N-way, K-shot setting 
\ying{is}
used to evaluate the performance. We use the standard four-block convolutional neural network as \ying{the} base model. We set \begin{small}$\alpha\!=\!\beta\!=\!2.0$\end{small} for all datasets. Detailed descriptions of experiment settings and hyperparameters are discussed in Appendix D.3.

\textbf{Results.} In Table~\ref{tab:homo_image_classification} and Appendix G.1, we report the performance (accuracy with 95\% confidence interval) on homogeneous datasets (i.e., Omniglot, MiniImagenet) and heterogeneous 
dataset\ying{s}, 
respectively. As described in Section~\ref{sec:ours}, we will use 
Channel Shuffle enhanced \ours\ (MMCF) in image classification problems.
\ying{Aligned with} other problems, in all non-mutually-exclusive datasets, applying the MMCF consistently 
\ying{improves existing meta-learning algorithms.
For example, MAML-MMCF significantly boosts MAML and most importantly outperforms MR-MAML, substantiating the effectiveness of MMCF in improving the meta-generalization ability.
}
\ying{It is worth mentioning that}
we \ying{also} conduct the experiments on \ying{the standard} mutually-exclusive setting of MiniImagenet in Appendix G.2.
Though the
label shuffling 
\ying{has} significantly mitigate\ying{d} meta-overfitting, 
applying MMCF 
still improves the meta-generalization 
to some extent. Besides, under \ying{the} MiniImagenet 5-shot scenario, we 
\ying{investigate}
the 
\ying{influence of}
different hyperparameter\ying{s, including sampling $\lambda$ from the Beta distribution with different values of} \begin{small}$\alpha$\end{small} \ying{and \begin{small}$\beta$\end{small}}, 
\ying{varying different fixed values of $\lambda$}, and \ying{adjusting} the \ying{layer to} mixup 
(i.e., $\mathcal{C}$ in Eqn.~\eqref{eq:metamix}) \ying{in} 
Appendix G.3. 
\ying{All} these 
studies indicate the robustness of \ours\ and \cf\ in improving the meta-generalization.

\textbf{Ablation Study.} To align 
with other problems, for MMCF, we vary the mixup and data augmentation strategies (i.e., \ours, \cf) in image classification in Table~\ref{tab:miniimagenet_mixture_strategy}. First, comparing \ours\ to other data mixup strategies, we again corroborate the effectiveness of \ours\ in improving meta-generalization. 
Second, we compare MMCF with MetaMix and \cf, the better performance of MMCF indicates the additional effects of \cf\ to enhance \ours\ in classification problems, as our theoretic analyses suggest.

%% file: conclusion.tex
\section{Conclusion}
Current gradient-based meta-learning algorithms are at high risk of
overfitting on meta-training tasks but poorly generalizing to 
meta-testing tasks. To address this issue, we propose two novel data augmentation strategies -- \ours\ and \cf, which actively involve more data in the outer-loop optimization process. Specifically, \ours\ linearly 
interpolates the features and labels of support and target sets. In classification problems, \ours\ is further enhanced by \cf, which randomly replaces a subset of channels with the corresponding ones from another class. We theoretically demonstrate that all strategies can improve the meta-generalization capability. The state-of-the-art results on different real-world datasets demonstrate the effectiveness and compatibility of the proposed methods. 

%% file: app-validity.tex
\appendix
\section{Validity of Different Task Augmentation Strategies (Detailed Proof)}

\subsection{Proof of Corollary 1}
\begin{Proof}[Proof of Corollary 1]
We check the validity of MetaMix as a task augmentation algorithm by examining whether the two criteria in Definition 1 in Section 3 are met.
First, we check the increase of mutual information between predictions of the query set and the support set. 
\begin{align}
& I(\hat{\mathbf{Y}}^{mix};(\mathbf{X}^{s}, \mathbf{Y}^{s})|\theta_0,\mathbf{X}^{mix}) - I(\hat{\mathbf{Y}}^q;(\mathbf{X}^s, \mathbf{Y}^s)|\theta_0,\mathbf{X}^q) \nonumber \\ 
= & H(\hat{\mathbf{Y}}^{mix} |\theta_0,\mathbf{X}^{mix})-H(\hat{\mathbf{Y}}^{mix} |\theta_0,\mathbf{X}^{mix},\mathbf{X}^s, \mathbf{Y}^s) -H(\hat{\mathbf{Y}}^q |\theta_0,\mathbf{X}^q) + H(\hat{\mathbf{Y}}^q|\theta_0,\mathbf{X}^q,\mathbf{X}^s, \mathbf{Y}^s) \nonumber \\ 
= & \mathbb{E}[-\log(p(\lambda\hat{\mathbf{Y}}^s + (1-\lambda)\hat{\mathbf{Y}}^q|\theta_0,\mathbf{X}^{mix}))] - \mathbb{E}[-\log(p(\lambda\hat{\mathbf{Y}}^s + (1-\lambda)\hat{\mathbf{Y}}^q|\theta_0,\mathbf{X}^{mix},\mathbf{X}^s,\mathbf{Y}^s))] \nonumber \\
& - \mathbb{E}[-\log(p(\hat{\mathbf{Y}}^q |\theta_0,\mathbf{X}^q))] + \mathbb{E}[-\log(p(\hat{\mathbf{Y}}^q|\theta_0,\mathbf{X}^q,\mathbf{X}^s, \mathbf{Y}^s))] \nonumber \\
= & \mathbb{E}[-\log(p(\lambda\hat{\mathbf{Y}}^s + (1-\lambda)\hat{\mathbf{Y}}^q|\theta_0,\mathbf{X}^{mix}))] 
- \mathbb{E}[-\log(p(\hat{\mathbf{Y}}^s|\theta_0,\mathbf{X}^{mix},\mathbf{X}^s,\mathbf{Y}^s))] \nonumber \\
& - \mathbb{E}[-\log(p(\hat{\mathbf{Y}}^q|\theta_0,\mathbf{X}^{mix},\mathbf{X}^s,\mathbf{Y}^s))] - \mathbb{E}[-\log(p(\hat{\mathbf{Y}}^q |\theta_0,\mathbf{X}^q))] + \mathbb{E}[-\log(p(\hat{\mathbf{Y}}^q|\theta_0,\mathbf{X}^q,\mathbf{X}^s, \mathbf{Y}^s))] \nonumber \\
= & \mathbb{E}[-\log(p(\lambda\hat{\mathbf{Y}}^s + (1-\lambda)\hat{\mathbf{Y}}^q|\theta_0,\mathbf{X}^{mix}))] 
 - \mathbb{E}[-\log(p(\hat{\mathbf{Y}}^q|\theta_0,\mathbf{X}^{mix},\mathbf{X}^s,\mathbf{Y}^s))] \nonumber \\
 & - \mathbb{E}[-\log(p(\hat{\mathbf{Y}}^q |\theta_0,\mathbf{X}^q))] + \mathbb{E}[-\log(p(\hat{\mathbf{Y}}^q|\theta_0,\mathbf{X}^q,\mathbf{X}^s, \mathbf{Y}^s))] \nonumber \\
 \geq & \mathbb{E}[-\log(p(\lambda\hat{\mathbf{Y}}^s + (1-\lambda)\hat{\mathbf{Y}}^q|\theta_0,\mathbf{X}^{q},\mathbf{X}^s))] 
 - \mathbb{E}[-\log(p(\hat{\mathbf{Y}}^q|\theta_0,\mathbf{X}^{mix},\mathbf{X}^s,\mathbf{Y}^s))] \nonumber \\
 & - \mathbb{E}[-\log(p(\hat{\mathbf{Y}}^q |\theta_0,\mathbf{X}^q))] + \mathbb{E}[-\log(p(\hat{\mathbf{Y}}^q|\theta_0,\mathbf{X}^q,\mathbf{X}^s, \mathbf{Y}^s))] \nonumber \\
 = & \mathbb{E}[-\log(p(\hat{\mathbf{Y}}^s|\theta_0,\mathbf{X}^{q},\mathbf{X}^{s}))] + \mathbb{E}[-\log(p(\hat{\mathbf{Y}}^q|\theta_0,\mathbf{X}^{q},\mathbf{X}^{s}))] - \mathbb{E}[-\log(p(\hat{\mathbf{Y}}^q|\theta_0,\mathbf{X}^{q},\mathbf{X}^s,\mathbf{Y}^s))] \nonumber \\
 & - \mathbb{E}[-\log(p(\hat{\mathbf{Y}}^q |\theta_0,\mathbf{X}^q))] + \mathbb{E}[-\log(p(\hat{\mathbf{Y}}^q|\theta_0,\mathbf{X}^q,\mathbf{X}^s, \mathbf{Y}^s))] \nonumber \\
 = & \mathbb{E}[-\log(p(\hat{\mathbf{Y}}^s|\theta_0,\mathbf{X}^{s}))] = H(\hat{\mathbf{Y}}^s|\theta_0,\mathbf{X}^{s}) \geq 0
\end{align}

Note that the third and the sixth equality holds as the mapping $(\hat{\mathbf{Y}^s},\hat{\mathbf{Y}}^q|\mathbf{X}^{mix},\mathbf{X}^s)\mapsto(\mathbf{Y}^{mix}|\mathbf{X}^{mix},\mathbf{X}^s)$ is one-to-one after $\lambda$ is specified. Besides, labels of the support (query) set are independent of features of the query (support) set, leading to the seventh equation. We investigate the capability of MetaMix producing additionally informative tasks in the following. 
\begin{align}
& I(\theta_0;\mathbf{X}^{mix}, \mathbf{Y}^{mix}|\mathbf{X}^q, \mathbf{Y}^q) \nonumber \\ 
= & H(\theta_0 | \mathbf{X}^q, \mathbf{Y}^q) - H(\theta_0 | \mathbf{X}^q, \mathbf{Y}^q, \mathbf{X}^{mix}, \mathbf{Y}^{mix}) \nonumber \\
= & \mathbb{E}[-\log(p(\theta_0 | \mathbf{X}^q, \mathbf{Y}^q))] - \mathbb{E}[-\log(p(\theta_0 | \mathbf{X}^q, \mathbf{Y}^q, \mathbf{X}^{mix}, \mathbf{Y}^{mix}))] \nonumber \\
= & \mathbb{E}[-\log(p(\theta_0 | \mathbf{X}^q, \mathbf{Y}^q))] - \mathbb{E}[-\log(\frac{p(\mathbf{X}^{mix},\mathbf{Y}^{mix}|\mathbf{X}^{q},\mathbf{Y}^{q},\theta_0)p(\mathbf{X}^{q},\mathbf{Y}^{q}|\theta_0)p(\theta_0)}{p(\mathbf{X}^{mix},\mathbf{Y}^{mix}|\mathbf{X}^{q},\mathbf{Y}^{q})p(\mathbf{X}^{q},\mathbf{Y}^{q})})] \nonumber \\
= & \mathbb{E}[-\log(p(\theta_0 | \mathbf{X}^q, \mathbf{Y}^q))] - \mathbb{E}[-\log(\frac{p(\mathbf{X}^{mix},\mathbf{Y}^{mix}|\mathbf{X}^{q},\mathbf{Y}^{q},\theta_0)}{p(\mathbf{X}^{mix},\mathbf{Y}^{mix}|\mathbf{X}^{q},\mathbf{Y}^{q})})]- \mathbb{E}[-\log(p(\theta_0 | \mathbf{X}^q, \mathbf{Y}^q))] \nonumber \\
= & -\mathbb{E}[-\log(\frac{p(\mathbf{Y}^{mix}|\mathbf{X}^{mix},\mathbf{X}^{q},\mathbf{Y}^{q},\theta_0)p(\mathbf{X}^{mix}|\mathbf{X}^{q},\mathbf{Y}^{q},\theta_0)}{p(\mathbf{Y}^{mix}|\mathbf{X}^{mix},\mathbf{X}^{q},\mathbf{Y}^{q})p(\mathbf{X}^{mix}|\mathbf{X}^{q},\mathbf{Y}^{q})})] \nonumber \\
= & -\mathbb{E}[-\log(\frac{p(\mathbf{Y}^s|\mathbf{X}^{mix},\mathbf{X}^{q},\mathbf{Y}^{q},\theta_0)p(\mathbf{Y}^q|\mathbf{X}^{mix},\mathbf{X}^{q},\mathbf{Y}^{q},\theta_0)p(\mathbf{X}^{s}|\mathbf{X}^{q},\mathbf{Y}^{q},\theta_0)p(\mathbf{X}^{q}|\mathbf{X}^{q},\mathbf{Y}^{q},\theta_0)}{p(\mathbf{Y}^s|\mathbf{X}^{mix},\mathbf{X}^{q},\mathbf{Y}^{q})p(\mathbf{Y}^q|\mathbf{X}^{mix},\mathbf{X}^{q},\mathbf{Y}^{q})p(\mathbf{X}^{s}|\mathbf{X}^{q},\mathbf{Y}^{q})p(\mathbf{X}^{q}|\mathbf{X}^{q},\mathbf{Y}^{q})})]\nonumber \\
= & -\mathbb{E}[-\log(\frac{p(\mathbf{Y}^s|\mathbf{X}^{mix},\mathbf{X}^{q},\mathbf{Y}^{q},\theta_0)p(\mathbf{X}^{s}|\mathbf{X}^{q},\mathbf{Y}^{q},\theta_0)}{p(\mathbf{Y}^s|\mathbf{X}^{mix},\mathbf{X}^{q},\mathbf{Y}^{q})p(\mathbf{X}^{s}|\mathbf{X}^{q},\mathbf{Y}^{q})})] \nonumber \\
= & -\mathbb{E}[-\log(\frac{p(\mathbf{Y}^s|\mathbf{X}^{s},\mathbf{X}^{q},\mathbf{Y}^{q},\theta_0)p(\mathbf{X}^{s}|\mathbf{X}^{q},\mathbf{Y}^{q},\theta_0)}{p(\mathbf{Y}^s|\mathbf{X}^{s},\mathbf{X}^{q},\mathbf{Y}^{q})p(\mathbf{X}^{s}|\mathbf{X}^{q},\mathbf{Y}^{q})})] \nonumber \\
= & -\mathbb{E}[-\log(\frac{p(\mathbf{X}^{s},\mathbf{Y}^s|\mathbf{X}^{q},\mathbf{Y}^{q},\theta_0)}{p(\mathbf{X}^{s},\mathbf{Y}^s|\mathbf{X}^{q},\mathbf{Y}^{q})})] \nonumber \\
= & -\mathbb{E}[-\log(\frac{p(\mathbf{X}^{s},\mathbf{Y}^s|\theta_0)p(\theta_0)}{p(\mathbf{X}^{s},\mathbf{Y}^s)})] + \mathbb{E}[-\log(p(\theta_0))] \nonumber \\
= & H(\theta_0) - H(\theta_0|\mathbf{X}^{s},\mathbf{Y}^s)). 
\end{align}
This indicates that MetaMix contributes a novel task as long as the support set of the task being augmented is capable of reducing the uncertainty of the initialization $\theta_0$, which is often the case.
Again, we would also note that the sixth equation holds due to the one-to-one mapping mentioned above after $\lambda$ is specified.
The tenth equation holds because the support set is assumed to be sampled independently from the query set. 
\end{Proof}

\subsection{Analysis of MetaMix enhanced with channel shuffle} We consider the support and the query set with channel shuffle to be $\mathbf{X}^{s,cf}=\varphi_{cf}(\mathbf{X}^s)$ and $\mathbf{X}^{q,cf}=\varphi_{cf}(\mathbf{X}^q)$, where $\varphi_{cf}$ is the non-linear function that replaces some channels of one class with the corresponding ones of the other class (refer to Eqn. (6) for the detailed discussion). Building on this, we validate the first criterion as follows.
\begin{align}
& I(\hat{\mathbf{Y}}^{mmcf};(\mathbf{X}^{s,cf}, \mathbf{Y}^{s})|\theta_0,\mathbf{X}^{mmcf}) - I(\hat{\mathbf{Y}}^q;(\mathbf{X}^s, \mathbf{Y}^s)|\theta_0,\mathbf{X}^q) \nonumber \\ 
= & H(\hat{\mathbf{Y}}^{mmcf} |\theta_0,\mathbf{X}^{mmcf})-H(\hat{\mathbf{Y}}^{mmcf} |\theta_0,\mathbf{X}^{m+cf},\mathbf{X}^{s,cf}, \mathbf{Y}^{s}) -H(\hat{\mathbf{Y}}^q |\theta_0,\mathbf{X}^q) + H(\hat{\mathbf{Y}}^q|\theta_0,\mathbf{X}^q,\mathbf{X}^s, \mathbf{Y}^s) \nonumber \\ 
= & \mathbb{E}[-\log(p(\hat{\mathbf{Y}}^{mmcf}|\theta_0,\mathbf{X}^{mmcf}))] 
 - \mathbb{E}[-\log(p(\hat{\mathbf{Y}}^q|\theta_0,\mathbf{X}^{mmcf},\mathbf{X}^{s,cf},\mathbf{Y}^s))] \nonumber \\
 & - \mathbb{E}[-\log(p(\hat{\mathbf{Y}}^q |\theta_0,\mathbf{X}^q))] + \mathbb{E}[-\log(p(\hat{\mathbf{Y}}^q|\theta_0,\mathbf{X}^q,\mathbf{X}^s, \mathbf{Y}^s))] \nonumber \\
 \geq & \mathbb{E}[-\log(p(\hat{\mathbf{Y}}^{mmcf}|\theta_0,\mathbf{X}^{q,cf},\mathbf{X}^{s,cf}))] 
 - \mathbb{E}[-\log(p(\hat{\mathbf{Y}}^q|\theta_0,\mathbf{X}^{mmcf},\mathbf{X}^{s,cf},\mathbf{Y}^s))] \nonumber \\
 & - \mathbb{E}[-\log(p(\hat{\mathbf{Y}}^q |\theta_0,\mathbf{X}^q))] + \mathbb{E}[-\log(p(\hat{\mathbf{Y}}^q|\theta_0,\mathbf{X}^q,\mathbf{X}^s, \mathbf{Y}^s))] \nonumber \\
  = & \mathbb{E}[-\log(p(\hat{\mathbf{Y}}^s|\theta_0,\mathbf{X}^{q,cf},\mathbf{X}^{s,cf}))] + \mathbb{E}[-\log(p(\hat{\mathbf{Y}}^q|\theta_0,\mathbf{X}^{q,cf},\mathbf{X}^{s,cf}))] \nonumber \\
& - \mathbb{E}[-\log(p(\hat{\mathbf{Y}}^q|\theta_0,\mathbf{X}^{q,cf},\mathbf{X}^{s,cf},\mathbf{Y}^s))]  - \mathbb{E}[-\log(p(\hat{\mathbf{Y}}^q |\theta_0,\mathbf{X}^q))] \nonumber \\
& + \mathbb{E}[-\log(p(\hat{\mathbf{Y}}^q|\theta_0,\mathbf{X}^q,\mathbf{X}^s, \mathbf{Y}^s))] \nonumber \\
= & \mathbb{E}[-\log(p(\hat{\mathbf{Y}}^s|\theta_0,\mathbf{X}^{q},\mathbf{X}^{s},\varphi_{cf}))] + \mathbb{E}[-\log(p(\hat{\mathbf{Y}}^q|\theta_0,\mathbf{X}^{q},\mathbf{X}^{s},\varphi_{cf}))] \nonumber \\
& - \mathbb{E}[-\log(p(\hat{\mathbf{Y}}^q|\theta_0,\mathbf{X}^{q},\mathbf{X}^{s},\mathbf{Y}^s,\varphi_{cf}))]  - \mathbb{E}[-\log(p(\hat{\mathbf{Y}}^q |\theta_0,\mathbf{X}^q))] \nonumber \\
& + \mathbb{E}[-\log(p(\hat{\mathbf{Y}}^q|\theta_0,\mathbf{X}^q,\mathbf{X}^s, \mathbf{Y}^s))] \nonumber \\
= & \mathbb{E}[-\log(p(\hat{\mathbf{Y}}^s|\theta_0,\mathbf{X}^{s},\varphi_{cf}))] + \mathbb{E}[-\log(p(\hat{\mathbf{Y}}^q|\theta_0,\mathbf{X}^{q},\varphi_{cf}))] \nonumber \\
& - \mathbb{E}[-\log(p(\hat{\mathbf{Y}}^q|\theta_0,\mathbf{X}^{q},\varphi_{cf}))]  - \mathbb{E}[-\log(p(\hat{\mathbf{Y}}^q |\theta_0,\mathbf{X}^q))] \nonumber \\
& + \mathbb{E}[-\log(p(\hat{\mathbf{Y}}^q|\theta_0,\mathbf{X}^q))] \nonumber \\
= & \mathbb{E}[-\log(p(\hat{\mathbf{Y}}^s|\theta_0,\mathbf{X}^{s},\varphi_{cf}))] = H(\hat{\mathbf{Y}}^s|\theta_0,\mathbf{X}^{s},\varphi_{cf}) \geq 0
\end{align}
In the next, we verify that the channel shuffle as expected produces a task that contributes more knolwedge to the initialization $\theta_0$ compared to using MetaMix only, thereby improving meta-generalization.
\begin{align}
& I(\theta_0;\mathbf{X}^{mmcf}, \mathbf{Y}^{mmcf}|\mathbf{X}^q, \mathbf{Y}^q) \nonumber \\ 
= & -\mathbb{E}[-\log(\frac{p(\mathbf{Y}^{mmcf}|\mathbf{X}^{mmcf},\mathbf{X}^{q},\mathbf{Y}^{q},\theta_0)p(\mathbf{X}^{mmcf}|\mathbf{X}^{q},\mathbf{Y}^{q},\theta_0)}{p(\mathbf{Y}^{mmcf}|\mathbf{X}^{mmcf},\mathbf{X}^{q},\mathbf{Y}^{q})p(\mathbf{X}^{mmcf}|\mathbf{X}^{q},\mathbf{Y}^{q})})] \nonumber \\
= & -\mathbb{E}[-\log(\frac{p(\mathbf{Y}^s|\mathbf{X}^{mmcf},\mathbf{X}^{q},\mathbf{Y}^{q},\theta_0)p(\mathbf{Y}^q|\mathbf{X}^{mmcf},\mathbf{X}^{q},\mathbf{Y}^{q},\theta_0)p(\varphi_{cf}(\mathbf{X}^{s})|\mathbf{X}^{q},\mathbf{Y}^{q},\theta_0)p(\varphi_{cf}(\mathbf{X}^{q})|\mathbf{X}^{q},\mathbf{Y}^{q},\theta_0)}{p(\mathbf{Y}^s|\mathbf{X}^{mmcf},\mathbf{X}^{q},\mathbf{Y}^{q})p(\mathbf{Y}^q|\mathbf{X}^{mmcf},\mathbf{X}^{q},\mathbf{Y}^{q})p(\varphi_{cf}(\mathbf{X}^{s})|\mathbf{X}^{q},\mathbf{Y}^{q})p(\varphi_{cf}(\mathbf{X}^{q})|\mathbf{X}^{q},\mathbf{Y}^{q})})]\nonumber \\
= & -\mathbb{E}[-\log(\frac{p(\mathbf{Y}^s|\mathbf{X}^{mmcf},\mathbf{X}^{q},\mathbf{Y}^{q},\theta_0)p(\varphi_{cf}(\mathbf{X}^{s})|\mathbf{X}^{q},\mathbf{Y}^{q},\theta_0)}{p(\mathbf{Y}^s|\mathbf{X}^{mmcf},\mathbf{X}^{q},\mathbf{Y}^{q})p(\varphi_{cf}(\mathbf{X}^{s})|\mathbf{X}^{q},\mathbf{Y}^{q})})] \nonumber \\
= & -\mathbb{E}[-\log(\frac{p(\mathbf{Y}^s|\mathbf{X}^{s},\mathbf{X}^{q},\mathbf{Y}^{q},\theta_0,\varphi_{cf})p(\mathbf{X}^{s},\varphi_{cf}|\mathbf{X}^{q},\mathbf{Y}^{q},\theta_0)}{p(\mathbf{Y}^s|\mathbf{X}^{s},\mathbf{X}^{q},\mathbf{Y}^{q},\varphi_{cf})p(\mathbf{X}^{s},\varphi_{cf}|\mathbf{X}^{q},\mathbf{Y}^{q})})] \nonumber \\
= & -\mathbb{E}[-\log(\frac{p(\mathbf{X}^{s},\mathbf{Y}^s,\varphi_{cf}|\mathbf{X}^{q},\mathbf{Y}^{q},\theta_0)}{p(\mathbf{X}^{s},\mathbf{Y}^s,\varphi_{cf}|\mathbf{X}^{q},\mathbf{Y}^{q})})] \nonumber \\
= & -\mathbb{E}[-\log(\frac{p(\mathbf{X}^{s},\mathbf{Y}^s,\varphi_{cf}|\theta_0)p(\theta_0)}{p(\mathbf{X}^{s},\mathbf{Y}^s,\varphi_{cf})})] + \mathbb{E}[-\log(p(\theta_0))] \nonumber \\
= & H(\theta_0) - H(\theta_0|\mathbf{X}^{s},\mathbf{Y}^s,\varphi_{cf})) \geq H(\theta_0) - H(\theta_0|\mathbf{X}^{s},\mathbf{Y}^s)). 
\end{align}

\subsection{Analysis of meta-augmentation}
Meta-augmentation that augments a task by randomly injecting a noise to the labels of both support and query set is not 
\yingicml{an effective task augmentation method,}
as it fails to meet the second criterion in Definition 1.
First of all, we check the validity of the first criterion.
\begin{align}
& I(\hat{\mathbf{Y}}^q + \epsilon;(\mathbf{X}^s, \mathbf{Y}^s+\epsilon)|\theta_0,\mathbf{X}^q) - I(\hat{\mathbf{Y}}^q;(\mathbf{X}^s, \mathbf{Y}^s)|\theta_0,\mathbf{X}^q) \nonumber \\ 
= & H(\hat{\mathbf{Y}}^q + \epsilon|\theta_0,\mathbf{X}^q)-H(\hat{\mathbf{Y}}^q + \epsilon|\theta_0,\mathbf{X}^q,\mathbf{X}^s, \mathbf{Y}^s+\epsilon) -H(\hat{\mathbf{Y}}^q |\theta_0,\mathbf{X}^q) + H(\hat{\mathbf{Y}}^q|\theta_0,\mathbf{X}^q,\mathbf{X}^s, \mathbf{Y}^s) \nonumber \\ 
 = &\mathbb{E}[-\log(p(\hat{\mathbf{Y}}^q + \epsilon|\theta_0,\mathbf{X}^q))] - \mathbb{E}[-\log(p(\hat{\mathbf{Y}}^q + \epsilon|\theta_0,\mathbf{X}^q,\mathbf{X}^s, \mathbf{Y}^s+\epsilon))] - \mathbb{E}[-\log(p(\hat{\mathbf{Y}}^q |\theta_0,\mathbf{X}^q))] \nonumber \\
& + \mathbb{E}[-\log(p(\hat{\mathbf{Y}}^q|\theta_0,\mathbf{X}^q,\mathbf{X}^s, \mathbf{Y}^s))] \nonumber \\
= & \mathbb{E}[-\log(p(\hat{\mathbf{Y}}^q|\theta_0,\mathbf{X}^q))] +  \mathbb{E}[-\log(p(\epsilon|\theta_0,\mathbf{X}^q))] - \mathbb{E}[-\log(p(\hat{\mathbf{Y}}^q|\theta_0,\mathbf{X}^q,\mathbf{X}^s, \mathbf{Y}^s+\epsilon))] \nonumber \\
& - \mathbb{E}[-\log(p(\epsilon|\theta_0,\mathbf{X}^q,\mathbf{X}^s, \mathbf{Y}^s+\epsilon))] - \mathbb{E}[-\log(p(\hat{\mathbf{Y}}^q |\theta_0,\mathbf{X}^q))]+ \mathbb{E}[-\log(p(\hat{\mathbf{Y}}^q|\theta_0,\mathbf{X}^q,\mathbf{X}^s, \mathbf{Y}^s))] \nonumber \\
= & \mathbb{E}[-\log(p(\epsilon))] - \mathbb{E}[-\log(p(\hat{\mathbf{Y}}^q|\theta_0,\mathbf{X}^q,\mathbf{X}^s, \mathbf{Y}^s+\epsilon))] - \mathbb{E}[-\log(p(\epsilon|\mathbf{Y}^s+\epsilon))] \nonumber \\
& + \mathbb{E}[-\log(p(\hat{\mathbf{Y}}^q|\theta_0,\mathbf{X}^q,\mathbf{X}^s, \mathbf{Y}^s))] \nonumber \\
\geq & \mathbb{E}[-\log(p(\epsilon))] - \mathbb{E}[-\log(p(\hat{\mathbf{Y}}^q|\theta_0,\mathbf{X}^q,\mathbf{X}^s, \mathbf{Y}^s))] - \mathbb{E}[-\log(p(\epsilon|\mathbf{Y}^s+\epsilon))] \nonumber \\
& + \mathbb{E}[-\log(p(\hat{\mathbf{Y}}^q|\theta_0,\mathbf{X}^q,\mathbf{X}^s, \mathbf{Y}^s))] \nonumber \\
= & H(\epsilon) \geq 0
\end{align}
Note that the third equality holds following the one-to-one assumption from $(\epsilon,x,y)\mapsto(x,y')$ in~\cite{rajendran2020meta} and the fact that $\epsilon$ is independent of $\hat{\mathbf{Y}}^q$, and the fourth holds as $\epsilon$ is independent of $\mathbf{X}^s$, $\mathbf{X}^q$, and $\theta_0$.
The first inequality is due to the theorem that conditioning reduces entropy. In the next, we \yingicml{prove that the algorithm of meta-augmentation fails to generate tasks with additional information.}
\begin{align}
& I(\theta_0;\mathbf{X}^q, \mathbf{Y}^q+\epsilon|\mathbf{X}^q, \mathbf{Y}^q) \nonumber \\
& = H(\theta_0|\mathbf{X}^q, \mathbf{Y}^q)-H(\theta_0|\mathbf{X}^q, \mathbf{Y}^q,\mathbf{X}^q, \mathbf{Y}^q+\epsilon) \nonumber \\
= & \mathbb{E}[-\log(p(\theta_0|\mathbf{X}^q, \mathbf{Y}^q))] - \mathbb{E}[-\log(p(\theta_0|\mathbf{X}^q, \mathbf{Y}^q,\mathbf{X}^q, \mathbf{Y}^q+\epsilon))] \nonumber \\
= & \mathbb{E}[-\log(p(\theta_0|\mathbf{X}^q, \mathbf{Y}^q))] - \mathbb{E}[-\log(\frac{p(\mathbf{X}^q,\mathbf{Y}^q,\mathbf{Y}^q+\epsilon|\theta_0)p(\theta_0)}{p(\mathbf{X}^q,\mathbf{Y}^q,\mathbf{Y}^q+\epsilon)})] \nonumber \\
= & \mathbb{E}[-\log(p(\theta_0|\mathbf{X}^q, \mathbf{Y}^q))] - \mathbb{E}[-\log(\frac{p(\mathbf{X}^q,\mathbf{Y}^q,\mathbf{Y}^q|\theta_0)p(\epsilon|\theta_0)p(\theta_0)}{p(\mathbf{X}^q,\mathbf{Y}^q)p(\epsilon)})] \nonumber \\
= & \mathbb{E}[-\log(p(\theta_0|\mathbf{X}^q, \mathbf{Y}^q))] - \mathbb{E}[-\log(p(\theta_0|\mathbf{X}^q, \mathbf{Y}^q))] = 0
\end{align}

%% file: app-method.tex
\section{Additional Information for \ours\ and MMCF}
\subsection{Figure for the Beta Distribution}
\label{sec:app_beta}
\begin{figure}[h]
    \centering
    \includegraphics[height=0.35\textwidth]{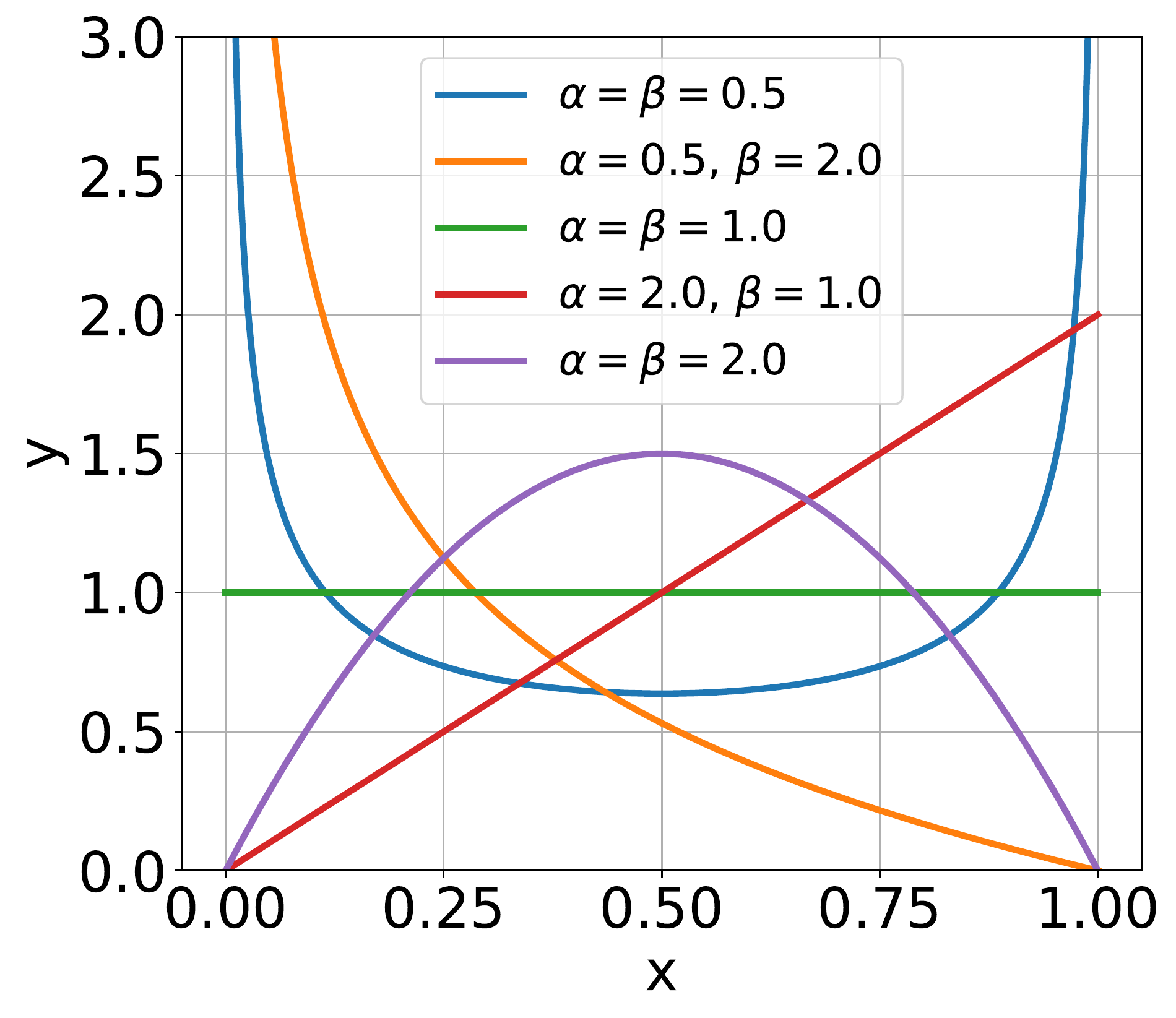}
    \caption{Illustration of the Beta Distribution. Here $\alpha=\beta$ and $\alpha\neq\beta$ represent the symmetric and skewed distributions, respectively.}
    \label{fig:beta}
\end{figure}

For the \begin{small}$\mathrm{Beta}(\alpha, \beta)$\end{small} distribution, we illustrate both symmetric (\begin{small}$\alpha=\beta$\end{small}) and skewed (i.e., \begin{small}$\alpha\neq\beta$\end{small}) scenarios in Figure~\ref{fig:beta}.
\subsection{Pseudo-codes}
\label{sec:app_pseducode}
Take MAML-MMCF as examples, we show the pseudo-codes for meta-training in Alg.~\ref{alg:mmcf_training}, respectively. The meta-testing process of MetaMix and MMCF are same, which is described in Alg.~\ref{alg:metamix_testing}.

\begin{algorithm}[h]
    \caption{Meta-training Process of MAML-MMCF}
    \label{alg:mmcf_training}
    \begin{algorithmic}[1]
    \REQUIRE Task distribution $p(\mathcal{T})$; Learning rate $\mu$, $\eta$; Beta distribution parameters $\alpha$, $\beta$; MetaMix candidate layer set $\mathcal{C}$ 
    \STATE Randomly initialize parameter $\theta_0$
    \WHILE{not converge}
    \STATE Sample a batch of tasks \begin{small}$\{\mathcal{T}_i\}_{i=1}^{n}$\end{small}
    \FORALL{\begin{small}$\mathcal{T}_i$\end{small}}
    \STATE Sample support set \begin{small}$\mathcal{D}^{s}_i=\{(\mathbf{x}^{s}_{i,j},\mathbf{y}^{s}_{i,j})\}_{j=1}^{K^s}$\end{small} and query set  \begin{small}$\mathcal{D}^{q}_i=\{(\mathbf{x}^{q}_{i,j},\mathbf{y}^{q}_{i,j})\}_{j=1}^{K^q}$\end{small} from \begin{small}$\mathcal{T}_i$\end{small}
    \STATE Sample a mixed layer $l$ from $\mathcal{C}$
    \STATE Sample Channel Shuffle parameter \begin{small}$\mathbf{R}_{c,c'}$\end{small} for each pair of classes \begin{small}$c$\end{small} and \begin{small}$c'$\end{small}
    \STATE Perform Channel Shuffle on the support set as (use a pair of classes as an example) via Eqn. (6) in the original paper: \begin{small}$\mathbf{X}^{s,cf}_{i;c}=\pmb{\mathrm{R}}_{c,c'} f_{\phi^l_i}(\mathbf{X}^{s}_{i;c})+(\mathbf{I}-\pmb{\mathrm{R}}_{c,c'})f_{\phi^l_i}(\mathbf{X}^{s}_{i;c'}),\; \mathbf{Y}^{s,cf}_{i;c}=\mathbf{Y}_{i;c}^{s}.$\end{small}
    \STATE Compute the task-specific parameter $\phi_i$ via the inner-loop gradient descent, i.e., \begin{small}$\phi_i=\theta_0-\mu\nabla_{\theta_0}\mathcal{L}(f_{\theta_0}(\mathbf{X}_i^{s,cf}), \mathbf{Y}_i^{s,cf})$\end{small}
    \STATE Perform Channel Shuffle on the query set via Eqn. (6) in the original paper: \begin{small}$\mathbf{X}^{q,cf}_{i;c}=\pmb{\mathrm{R}}_{c,c'} f_{\phi_i^l}(\mathbf{X}^{q}_{i;c})+(\mathbf{I}-\pmb{\mathrm{R}}_{c,c'})f_{\phi_i^l}(\mathbf{X}^{q}_{i;c'}),\; \mathbf{Y}^{q,cf}_{i;c}=\mathbf{Y}_{i;c}^{q}.$\end{small}
    \STATE Sample MetaMix parameter \begin{small}$\pmb{\lambda}\sim \mathrm{Beta}(\alpha, \beta)$\end{small}
    \STATE Forward both support and query sets and mixed them at layer \begin{small}$l$\end{small} as: \begin{small}$\mathbf{X}^{mmcf}_{i,l}=\pmb{\lambda} f_{\phi_i^l}(\mathbf{X}^{s,cf}_i)+(\mathbf{I}-\pmb{\lambda})f_{\phi_i^l}(\mathbf{X}^{q,cf}_i)$\end{small}, \begin{small}$\mathbf{Y}^{mmcf}_i=\pmb{\lambda}\mathbf{Y}_i^{s,cf}+(\mathbf{I}-\pmb{\lambda})\mathbf{Y}_i^{q,cf}$\end{small}
    \STATE Continual forward \begin{small}$\mathbf{X}^{mix}_{i,l}$\end{small} to the rest of layers and compute the loss as \begin{small}$\mathcal{L}(f_{\phi_i^{L-l}}(\mathbf{X}_{i,l}^{mmcf}), \mathbf{Y}_{i}^{mmcf})$\end{small}
    \ENDFOR
    \STATE Update \begin{small}$\theta_0\leftarrow \theta_0 - \frac{1}{n}\sum_{i=1}^{n}
    \mathbb{E}_{\pmb{\lambda}\sim \mathrm{Beta}(\alpha, \beta)}\mathbb{E}_{l\sim \mathcal{C}}[\mathcal{L}(f_{\phi_i^{L-l}}(\mathbf{X}_{i,l}^{mmcf}), \mathbf{Y}_i^{mmcf})]$\end{small}
    \ENDWHILE
    \end{algorithmic}
\end{algorithm}

\begin{algorithm}[h]
    \caption{Meta-testing Process of both MAML-MetaMix or MAML-MMCF}
    \label{alg:metamix_testing}
    \begin{algorithmic}[1]
    \REQUIRE Learning rate $\mu$; Optimized parameter $\theta_0^{*}$ via MMCF or MetaMix
    \STATE Compute the task-specific parameter \begin{small}$\phi_t$\end{small} as \begin{small}$\phi_t=\theta_0^{*}-\mu\nabla_{\theta_0}\mathcal{L}(f_{\theta_0}(\mathbf{X}_t^s), \mathbf{Y}_t^s)$\end{small}
    \STATE Predict \begin{small}$\mathbf{\hat{Y}}_t^q$\end{small} on the query set \begin{small}$\mathcal{D}_t^q$\end{small}
    \STATE Evaluate the performance via predicted value \begin{small}$\mathbf{\hat{Y}}_t^q$\end{small} and actual value \begin{small}$\mathbf{Y}_t^q$\end{small} 
    \end{algorithmic}
\end{algorithm}

%% file: app-generalization.tex
\section{Detailed Proof of Generalization Analysis}

\begin{Proof}[Proof of Theorem 1]
We first state a standard uniform deviation bound based on Rademacher complexity (c.f. \cite{bartlett2002rademacher})
\begin{lemma}
 Let the sample $\{z_1, . . . , z_N\}$ be drawn i.i.d. from a distribution $P$ over $\mathcal Z$ and let $\mathcal G$ denote a class of functions on $\mathcal Z$ with members mapping from $\mathcal Z$ to $[a, b]$. Then for $\delta>0$, we have that with probability at least $1-\delta$ over the draw of the sample,
\begin{equation}
\sup_{g\sim G}\|\mathbb{E}_{\hat P} g(z)-\mathbb{E}_{P}g(z) \|\le 2R(G; z_1,...,z_n)+\sqrt\frac{\log(1/\delta)}{n},
\end{equation}
where $R(G; z_1,...,z_n)$ denotes the Rademacher complexity of the function class $\mathcal G$.
\end{lemma}

We now write $R(\{\mathbf{Z}_i\}_{i=1}^{n_T})-R$ as
\begin{align*}
R(\{\mathbf{Z}_i\}_{i=1}^{n_T})-R=&\mathbb{E}_{\mathcal{T}_i\sim\hat p(\mathcal{T})}\mathbb{E}_{(\mathbf{X}_i,\mathbf{Y}_i)\sim\hat p(\mathcal{T}_i)}\mathcal{L}(f_{\phi_i}(\mathbf{X}_i^q), \mathbf{Y}_i^q)-\mathbb{E}_{\mathcal{T}_i\sim p(\mathcal T)}\mathbb{E}_{(\mathbf{X}_i,\mathbf{Y}_i)\sim T_i}[\mathcal{L}(f_{\phi_i}(\mathbf{X}_i^q), \mathbf{Y}_i^q)]\\
=&\underbrace{\mathbb{E}_{\mathcal{T}_i\sim\hat p(\mathcal{T})}\mathbb{E}_{(\mathbf{X}_i,\mathbf{Y}_i)\sim\hat p(\mathcal{T}_i)}\mathcal{L}(f_{\phi_i}(\mathbf{X}_i^q), \mathbf{Y}_i^q)-\mathbb{E}_{\mathcal{T}_i\sim\hat p(\mathcal{T})}\mathbb{E}_{(\mathbf{X}_i,\mathbf{Y}_i)\sim \mathcal{T}_i}[\mathcal{L}(f_{\phi_i}(\mathbf{X}_i^q), \mathbf{Y}_i^q)]}_{(i)}\\
&+\underbrace{\mathbb{E}_{\mathcal{T}_i\sim\hat p(\mathcal{T})}\mathbb{E}_{(\mathbf{X}_i,\mathbf{Y}_i)\sim \mathcal{T}_i}\mathcal{L}(f_{\phi_i}(\mathbf{X}_i^q), \mathbf{Y}_i^q)-\mathbb{E}_{\mathcal{T}_i\sim p(\mathcal{T})}\mathbb{E}_{(\mathbf{X}_i,\mathbf{Y}_i)\sim \mathcal{T}_i}[\mathcal{L}(f_{\phi_i}(\mathbf{X}_i^q),\mathbf{Y}_i^q)]}_{(ii)}
\end{align*}

Recall that we consider the function  $f_{\phi_i}(\mathbf X_i)=\phi_i^\top\sigma(\mathbf{W}\mathbf X_i)$ and the function class 
\begin{equation}
\mathcal{F}_\mathcal{T}=\{\phi^\top\sigma(\mathbf{W}\mathbf X): \phi^\top \Sigma_{\sigma,\mathcal{T}}\phi\le\gamma \}.
\end{equation}

For each $\mathcal{T}_i$, let us consider $f_{\phi_i}(\cdot)\in\mathcal F_{\mathcal{T}}$. By Theorem A.1 in \cite{zhang2020does}, we have the following result for the Rademacher complexity:
\begin{equation}
\begin{aligned}
R(\mathcal F_{\mathcal{T}}; z_1,...,z_n)\le&2\sqrt{\frac{\gamma\cdot(rank( \Sigma_{\sigma,\mathcal{T}})+\|\Sigma_{\sigma,\mathcal{T}}^{\mathbf{W}\dagger/2}\mu_{\sigma,\mathcal{T}}\|)}{K^m}}\\
\le&2\sqrt{\frac{\gamma\cdot(r+B)}{K^m}}.
\end{aligned}
\end{equation}

Then the first term (i) can be bounded as below.
\begin{equation}
\begin{aligned}
&\mathbb{E}_{\mathcal{T}_i\sim\hat p(\mathcal{T})}\mathbb{E}_{(\mathbf X_i,\mathbf Y_i)\sim\hat p(T_i)}\mathcal{L}(f_{\phi_i}(\mathbf X_i^q), \mathbf Y_i^q)-\mathbb{E}_{\mathcal{T}_i\sim\hat p(\mathcal T)}\mathbb{E}_{(\mathbf X_i,\mathbf Y_i)\sim \mathcal{T}_i}[\mathcal{L}(f_{\phi_i}(\mathbf X_i^q), \mathbf Y_i^q)]\\
\le &\mathbb{E}_{T_i\sim\hat p(\mathcal{T})}|\mathbb{E}_{(\mathbf X_i,\mathbf Y_i)\sim\hat p(\mathcal{T}_i)}\mathcal{L}(f_{\phi_i}(\mathbf X_i^q), \mathbf Y_i^q)-\mathbb{E}_{(\mathbf X_i,\mathbf Y_i)\sim \mathcal{T}_i}[\mathcal{L}(f_{\phi_i}(\mathbf X_i^q), \mathbf Y_i^q)]\\
\le &C_1\sqrt{\frac{\gamma\cdot(r+B)}{K^m}}+C_2\sqrt\frac{\log(n_T/\delta)}{K^m},
\end{aligned}
\end{equation}
where the additional $\log(n_T)$ term in the last inequality above is due to we take union bound on $n_T$ tasks.

Denote function $g: \mathcal{T}\to \mathbb{R}$ such that $g_f(\mathcal{T})=\mathbb{E}_{(\mathbf{X},\mathbf{Y})\sim \mathcal{T}}(\mathcal{L}(f_{\phi}(\mathbf{X}),\mathbf{Y}))$. Denote 
\begin{equation}
\mathcal G=\{g_f(\mathcal{T}): g_f(\mathcal{T})=\mathbb{E}_{(\mathbf{X},\mathbf{Y})\sim \mathcal{T}}(\mathcal{L}(f_{\phi}(\mathbf{X}),\mathbf{Y})), f_{\phi}\in \mathcal F_{\mathcal T}\}.
\end{equation}
 The second term (ii) requires computing the Rademacher complexity for the function class over distributions \begin{equation}
 \begin{aligned}
R(\mathcal G; \mathcal{T}_1,...,\mathcal{T}_{n_T})=&\mathbb{E} \sup_{g\in \mathcal{G}}\frac{1}{n_T}|\sum_{i=1}^{n_T}\sigma_i g(\mathcal{T}_i)|=\mathbb{E} \sup_{g\in \mathcal{G}}\frac{1}{n_T}|\sum_{i=1}^{n_T}\sigma_i \mathbb{E}_{(\mathbf{X},\mathbf{Y})\sim \mathcal{T}_i}(f_{\phi_i}(\mathbf{X})-\mathbf{Y})^2|\\
\le&\mathbb{E} \sup_{g\in \mathcal{G}}\frac{1}{n_T}|\sum_{i=1}^{n_T}\sigma_i \mathbb{E}_{(\mathbf{X},\mathbf{Y})\sim \mathcal{T}_i}f_{\phi_i}(\mathbf{X})|+\mathbb{E} \sup_{g\in \mathcal{G}}\frac{1}{n_T}|\sum_{i=1}^{n_T}\sigma_i \mathbb{E}_{(\mathbf{X},\mathbf{Y})\sim \mathcal{T}_i}\mathbf{Y}|\\
\le&\mathbb{E} \sup_{g\in G}\frac{1}{n_T}|\sum_{i=1}^{n_T}\sigma_i(\Sigma_{\sigma,\mathcal{T}}^{2}\phi_i)^\top \Sigma_{\sigma,\mathcal{T}}^{\dagger/2}\mu_{\sigma,\mathcal{T}} |+\sqrt\frac{1}{n_T}\\
\le& \sqrt{\frac{\gamma\cdot B+1}{n_T}}
\end{aligned}
\end{equation}

Then we have the following bound on (ii): 
\begin{equation}
\begin{aligned}
&\mathbb{E}_{\mathcal{T}_i\sim\hat p(\mathcal{T})}\mathbb{E}_{(\mathbf X_i,\mathbf Y_i)\sim \mathcal{T}_i}\mathcal{L}(f_{\phi_i}(\mathbf X_i^q), \mathbf Y_i^q)-\mathbb{E}_{\mathcal{T}_i\sim p(\mathcal T)}\mathbb{E}_{(\mathbf X_i,\mathbf Y_i)\sim \mathcal{T}_i}[\mathcal{L}(f_{\phi_i}(\mathbf X_i^q), \mathbf Y_i^q)]\\
\le& C_3\sqrt{\frac{\gamma\cdot(B+1)}{n_T}}+C_4\sqrt\frac{\log(1/\delta)}{n_T}.
\end{aligned}
\end{equation}

Combining the pieces, we obtain the desired result. With probability at least $1 -\delta$, 
 \begin{equation}
 |R(\{\mathbf{Z}_i\}_{i=1}^{n_T})-R|\le C_1\sqrt{\frac{\gamma\cdot(r+B)}{K^m}}+C_2\sqrt\frac{\log(n_T/\delta)}{K^m}+C_3\sqrt{\frac{\gamma\cdot B+1}{n_T}}+C_4\sqrt\frac{\log(1/\delta)}{n_T}.
 \end{equation}
\end{Proof}

Besides the detailed proof, we also provide the empirical results to show the equivalence between the symmetric version of MAML-MetaMix for generalization analysis (i.e., Mixup$(\mathcal{D}^s\oplus \mathcal{D}^q, \mathcal{D}^s\oplus \mathcal{D}^q)$) and the proposed MAML-MetaMix (i.e., Mixup$(\mathcal{D}^s, \mathcal{D}^q)$). The experiments are conducted on both omniglot and miniImagenet under the non-exclusive setting. In Table~\ref{tab:generalization_sq_sq}, we report the comparison results:

\begin{table}[h]
\small
\caption{Performance comparison between Mixup$(\mathcal{D}^s\oplus \mathcal{D}^q, \mathcal{D}^s\oplus \mathcal{D}^q)$ and Mixup$(\mathcal{D}^s, \mathcal{D}^q)$.}
\label{tab:generalization_sq_sq}
\begin{center}
\begin{tabular}{l|cc|cc}
\toprule
\multirow{2}{*}{Model} & \multicolumn{2}{c|}{Omniglot} & \multicolumn{2}{c}{MiniImagenet}\\
& 20-way 1-shot & 20-way 5-shot & 5-way 1-shot & 5-way 5-shot \\\midrule
MAML-MetaMix (Mixup$(\mathcal{D}^s, \mathcal{D}^q))$ & $91.53\pm0.53\%$ & $97.63\pm0.15\%$ & $38.53\pm1.79\%$ & $57.55\pm1.01\%$\\
MAML-MetaMix (Mixup$(\mathcal{D}^s\oplus \mathcal{D}^q, \mathcal{D}^s\oplus \mathcal{D}^q))$ & $91.93\pm0.52\%$ & $97.17\pm0.16\%$ & $38.27\pm1.72\%$ & $57.38\pm0.96\%$ \\\bottomrule

\end{tabular}
\end{center}
\vspace{-1em}
\end{table}

\begin{Proof}[Proof of Theorem 2]
To prove Theorem 2, first, we would like to note that since $\frac{1}{K^{m_0}}\sum_{j=1}^{K^{m_0}}\sigma(\mathbf{W}\mathbf{x}_{i,j;0})=\frac{1}{K^{m_1}}\sum_{j=1}^{K^{m_1}}\sigma(\mathbf{W}\mathbf{x}_{i,j;1})=0$, we have
\begin{equation}
    \mathbb{E}[\mathbf{x}^{cf}_{i,j;k}\mid \mathbf{x}_{i,j;k}]= \sigma(\mathbf{W}\mathbf{x}_{i,j;k}).
\end{equation}

Recall that $\mathcal{L}(f_{\phi_i}(\mathbf{X}_{i}),\mathbf{Y}_{i})=\frac{1}{2}(f_{\phi_i}(\mathbf{X}_i)-\mathbf{Y}_i)^2$. Then let us compute the second-order Taylor expansion on $\mathcal{L}(\mathbf{Z}_i^{cf})=\frac{1}{K^m}\sum_{j=1}^{K^m}\mathcal{L}({\phi_i}^\top(\mathbf{x}^{cf}_{i,j}), \mathbf \mathbf{y}^{}_{i,j})$ with respect to $(\mathbf{x}^{cf}_{i,j})$ around $\mathbb{E}[\mathbf{x}^{cf}_{i,j;k}\mid \mathbf{x}_{i,j;k}]= \sigma(\mathbf{W}\mathbf{x}_{i,j;k})$, we have that the  Taylor expansion of $\mathbb{E}_{\xi}\mathcal{L}(\mathbf{Z}_i^{cf})$ up to the second-order equals to
\begin{align}\label{eq:second}
\mathcal{L}(\mathbf{Z}_i)+\frac{1}{K^m}\sum_{j=1}^{K^{m_0}}+\frac{1}{K^m}\sum_{j=1}^{K^m} \phi_i^\top Cov(\mathbf{x}^{cf}_{i,j}\mid \mathbf{Z}_i)\phi_i
\end{align}
Let us denote $ \sigma(\mathbf{W}\mathbf{x}_{i,j;k})$ by $\mathbf{x}_{i,j;k}^\sigma$.

For the quadratic term, we have that given $\mathbf{Z}_i$ 
\begin{equation}
\begin{aligned}
Cov(\mathbf{x}^{cf}_{i,j})=&\frac{1}{\delta^2}Cov(\left(\mathbf{R} \sigma(\mathbf{W}\mathbf{x}_{i,j;k})+(\mathbf{I}-\mathbf{R})\sigma(\mathbf{W}\mathbf{x}_{i,j';1-k})\right))\\
=&\frac{1}{\delta^2}(Cov\left(\mathbf{R} \mathbf{x}_{i,j;k}^\sigma\right)+Cov\left(\mathbf{R} \mathbf{x}_{i,j;k}^\sigma,(\mathbf{I}-\mathbf{R}) \mathbf{x}_{i,j';1-k}^\sigma\right)+Cov\left(\mathbf{I}-\mathbf{R}) \mathbf{x}_{i,j';1-k}^\sigma\right))\\
=&\frac{1}{\delta^2}(\delta(1-\delta)diag(\mathbf{x}_{i,1;k}^{\sigma\circ 2})+0+\delta(1-\delta)\frac{1}{K^{m_{1-k}}}\sum_{j=1}^{K^{m_{1-k}}}\mathbf{x}_{i,j;1-k}^\sigma\mathbf{x}_{i,j;1-k}^{\sigma\top})
\end{aligned}
\end{equation}

Plugging into Eq~\eqref{eq:second}, we obtain 
\begin{equation}
\begin{aligned}
\mathcal{L}(\mathbf{Z}_i)&+\frac{1-\delta}{\delta}\phi_i^\top(\frac{1}{K^m}\sum_{j=1}^{K^m} diag( \sigma(\mathbf{W}\mathbf{x}_{i,j})^{\circ 2})\phi_i+\\
&+\phi_i^\top(\frac{1}{K^{m_0}}\sum_{j=1}^{K^{m_0}}\sigma(\mathbf{W}\mathbf{x}_{i,j;0} )\sigma(\mathbf{W}\mathbf{x}_{i,j;0} )^\top+\frac{1}{K^{m_1}}\sum_{j=1}^{K^{m_1}}\sigma(\mathbf{W}\mathbf{x}_{i,j;1} )\sigma(\mathbf{W}\mathbf{x}_{i,j;1} )^\top) \phi_i.
\end{aligned}
\end{equation}

\end{Proof}

%% file: app-setting.tex
\section{Detailed Experimental Setup}
In this section, we provide more details 
\kai{of} the experimental setup\kai{s} of our paper. All experiments are run on a GPU cluster and implemented by Tensorflow~\citep{abadi2016tensorflow}. 
In the next, we discuss the setup\kai{s} for all the problems, including drug activity prediction, pose prediction, and image classification.
\subsection{Drug Activity Prediction}
\label{sec:app_drugsetting}
For drug activity prediction, we use the publicly available dose-response activity assays from ChEMBL\footnote{\href{https://www.ebi.ac.uk/chembl}{https://www.ebi.ac.uk/chembl}} and preprocessed in~\citep{martin2019all}.
All 4,276 assays, as 4,276 tasks, are accessible and downloadable from this site\footnote{\href{https://pubs.acs.org/doi/10.1021/acs.jcim.9b00375\#i21}{https://pubs.acs.org/doi/10.1021/acs.jcim.9b00375\#i21}}.
In each assay, there are a few training drug compounds with biologically tested activities against the target protein in this assay, as well as several testing compounds. 
The split of training and testing compounds follows the realistic split in~\citep{martin2019all}.
The number of drug compounds varies from assay to assay, with an median of only 70 drug compounds per assay.
To describe each compound, we follow~\citep{martin2019all} to use 1,024 dimensional Morgan fingerprint implemented in RDkit\footnote{\href{http://rdkit.org/}{http://rdkit.org/}}.
As mentioned in the experimental section in the main text, we randomly take 100 assays as meta-testing assays, and 76 assays for meta-validation and the rest of 4100 assays for meta-training. 
Here we report the assay IDs that belong to meta-validation and meta-testing, respectively, for all the four groups.
Note that due to space limit we do not report the assay IDs for meta-training, which can be easily obtained by deducting the meta-validation and meta-testing assays from all 4276 assays.
\begin{itemize}[leftmargin=*]
\item \textbf{Group 1}
\begin{itemize}
    \item \emph{Meta-validation:} \begin{small}972800, 688641, 610565, 1536390, 211079, 1625735, 1641357, 688654, 1641103, 457234, 450707, 195220, 1366808, 49308, 924, 49312, 828065, 737313, 1528100, 596645, 1641767, 1535401, 688427, 969260, 453677, 978479, 1641008, 574385, 911154, 446257, 878513, 1640955, 902584, 1276473, 752567, 306492, 736957, 1640384, 1454018, 2755, 579907, 1527622, 761927, 89542, 809158, 978889, 556876, 478840, 688464, 1330005, 144341, 1528791, 1301597, 1641310, 209245, 608993, 1528801, 89064, 1527913, 4202, 688616, 1513, 510189, 1641197, 1527791, 688495, 89839, 1641201, 1528688, 752371, 688379, 938230, 596087, 835704, 566779, 688767.\end{small}
    \item \emph{Meta-testing:} \begin{small}752640, 972801, 737284, 954885, 1528837, 1587725, 1527823, 1640977, 157713, 1285138, 1437208, 1349151, 1592870, 93228, 465460, 954934, 84556, 1567308, 1577550, 1285709, 654928, 620647, 864364, 575603, 1280627, 688257, 1443970, 1527947, 737424, 201877, 1457820, 603293, 809120, 883875, 1641128, 1534634, 1641655, 955073, 954571, 736971, 577227, 45264, 455393, 728290, 688357, 1301747, 105205, 865015, 665348, 820998, 759559, 1301769, 609034, 80649, 1641240, 965916, 34078, 1470241, 1348900, 333106, 1527607, 954703, 1641298, 1641300, 727385, 304989, 981861, 212325, 756584, 331630, 473976, 63356, 51590, 1640328, 954762, 1642379, 1527698, 1527704, 543133, 954781, 1301405, 619939, 605612, 585134, 1433006, 934321, 1642435, 1637320, 936907, 54735, 70610, 1508820, 1292758, 104407, 992729, 199642, 160234, 1528304, 629753, 931327. \end{small}
\end{itemize}
\item \textbf{Group 2}
\begin{itemize}
    \item \emph{Meta-validation:} \begin{small}7296, 1276546, 87173, 688645, 1350406, 955016, 697223, 1163, 201739, 809231, 1528850, 1528212, 752533, 971798, 954388, 1626011, 1528480, 501795, 1527972, 470053, 1640867, 809128, 737064, 1642538, 954282, 978478, 786095, 29233, 1642418, 737075, 1536179, 1641399, 1527735, 609465, 1640506, 1641659, 307259, 1537597, 769089, 140229, 789189, 860488, 766795, 48587, 1528909, 1451727, 219472, 737105, 955090, 311637, 1528022, 1632983, 727385, 1456602, 1641179, 688347, 67039, 434528, 1564001, 727521, 688483, 595939, 1436004, 736997, 1528160, 1640426, 4202, 102381, 45422, 1641073, 47858, 37363, 1641720, 688889, 1301756, 556797.
    \end{small}
    \item \emph{Meta-testing:} \begin{small} 835072, 539657, 1641997, 1639955, 1638422, 1639959, 1622038, 637980, 28188, 91168, 954915, 425511, 688685, 155185, 39493, 155208, 1641035, 1288277, 755797, 954462, 812132, 87656, 1536113, 48248, 744057, 210045, 1642144, 50337, 325795, 1527974, 1642150, 814256, 1641143, 438974, 217297, 1641170, 688340, 1641688, 688357, 649964, 930033, 447747, 566532, 1641737, 49425, 562451, 817939, 688403, 817944, 52506, 452895, 984872, 311595, 899888, 646978, 1642307, 664904, 1641802, 1466703, 1466704, 809297, 147797, 1640791, 305497, 209245, 603488, 701282, 752485, 302952, 122731, 563052, 1561972, 1528692, 1642361, 1528698, 737150, 1301374, 51590, 364426, 1642378, 899993, 752538, 1640355, 1446827, 62394, 842684, 1640893, 44489, 688589, 208335, 1642449, 858065, 1640919, 1528791, 1528294, 1520, 1640952, 92156, 63997, 69119.\end{small}
\end{itemize}
\item \textbf{Group 3}
\begin{itemize}
    \item \emph{Meta-validation:} \begin{small}856700, 688641, 448646, 1613063, 1301767, 624014, 559247, 1527953, 1640339, 49558, 737046, 809242, 1642522, 1641371, 1527965, 592925, 954655, 688416, 305569, 538786, 1535011, 208672, 1592863, 688550, 1527974, 1527976, 1642272, 305065, 809259, 1640189, 96941, 688685, 954799, 978480, 934321, 1637168, 29233, 45236, 306221, 1535033, 1640506, 1290683, 158524, 936637, 647615, 422463, 1459648, 1640904, 954953, 1361352, 654923, 1641164, 954959, 1301583, 688210, 1508820, 45272, 688346, 737371, 162397, 775393, 535396, 1301477, 4197, 651627, 75756, 3819, 737391, 1641201, 1528692, 1294964, 456311, 737273, 1301756, 1640573, 42878.\end{small}
    \item \emph{Meta-testing:} \begin{small}688643, 688645, 159749, 1527820, 539663, 303638, 1638422, 954399, 330271, 70695, 1295917, 1642542, 1527862, 1528890, 200254, 540741, 1365575, 761928, 688201, 37966, 1537108, 336476, 511069, 1301599, 1528416, 206959, 478840, 1503357, 1642117, 1536654, 688274, 1527963, 688288, 688293, 688816, 745138, 438974, 88771, 459971, 714443, 1289425, 1451729, 1284820, 954602, 954604, 208118, 198910, 809216, 610565, 1528071, 453897, 770827, 216843, 828171, 306447, 562451, 1642271, 1528097, 28965, 367910, 1642296, 1528125, 1528145, 1555281, 49489, 493905, 876885, 1290079, 468834, 1528677, 756582, 1338728, 1528170, 845165, 1528696, 617338, 1301888, 102785, 1527682, 940424, 1528724, 809380, 1446827, 864186, 1291714, 642499, 688586, 1513931, 32721, 954834, 1536468, 688598, 1556442, 901084, 954845, 1527780, 1640932, 477677, 829947, 1528829.\end{small}
\end{itemize}
\item \textbf{Group 4}
\begin{itemize}
    \item \emph{Meta-validation:} \begin{small}688391, 1641992, 1641737, 306314, 311816, 813068, 68748, 1536775, 823822, 1639955, 696215, 1469079, 971801, 41884, 637980, 1298461, 76063, 688416, 1637151, 619938, 1642274, 885155, 572966, 510887, 1641000, 1528488, 954411, 1642415, 138287, 1527985, 56498, 27571, 458930, 311855, 809146, 307259, 950588, 736957, 1527742, 809152, 592450, 809156, 1528648, 954957, 209231, 490576, 1528273, 1641170, 45265, 1528917, 1528149, 1292759, 809175, 53367, 737370, 822749, 154333, 67039, 737273, 1640162, 737379, 763492, 809193, 954987, 104172, 510189, 1528695, 208754, 688243, 45044, 954482, 1640307, 1528183, 1642489, 1527674, 688254.\end{small}
    \item \emph{Meta-testing:} \begin{small}1556484, 688644, 737290, 1301520, 1642001, 1528850, 688661, 971799, 1642520, 46624, 1537067, 1641005, 1641010, 688185, 954938, 443966, 599616, 439367, 1537607, 954956, 688719, 615506, 654934, 1589851, 104542, 457824, 1641573, 1527916, 737391, 1301619, 211078, 52874, 955024, 32404, 158358, 566940, 50848, 1527970, 1349288, 1586856, 860330, 688818, 1536181, 48316, 491718, 1296583, 954567, 1566412, 66255, 66267, 809183, 425699, 467683, 464617, 752377, 508163, 872708, 1640197, 1301765, 809221, 809239, 1528102, 809255, 1536298, 1640747, 688431, 1636657, 213817, 1466703, 1301330, 1545042, 737622, 1452895, 950625, 856937, 493931, 954743, 809346, 558984, 1642378, 591251, 1640852, 305050, 934299, 1640862, 1640351, 1642418, 558515, 1511354, 1330619, 1528770, 1291715, 901575, 1640904, 1439182, 1537998, 737235, 1301469, 37371, 797692. \end{small}
\end{itemize}
\end{itemize}

The base model of drug activity prediction is a two-layer Multilayer Perceptron(MLP) neural network with 500 neurons in each layer. Each fully connected layer is followed by a batch normalization layer and leaky ReLU activation (negative slope is 0.01). In $\mathrm{Beta}(\alpha, \alpha)$, we set $\alpha=0.5$. We set the candidate layer $\mathcal{C}$ as layer 1 and layer 2. 
During meta-training, the task batch size, the outer-loop learning rate,  the inner-loop learning rate are set to 8, 0.001, and 0.01, respectively.
The meta-training process altogether runs for 50 epochs, each of which includes 500 iterations.
In either meta-training or meta-testing, the number of inner-loop adaptation steps equals to 5.

\subsection{Pose Prediction}
\label{sec:app_posesetting}
In the pose prediction problem, we follow~\citep{yin2020meta} to preprocess the pose tasks\footnote{code link: \href{https://github.com/google-research/google-research/tree/master/meta\_learning\_without\_memorization/pose\_data}{https://github.com/google-research/google-research/tree/master/meta\_learning\_without\_memorization/pose\_data}}. The meta-training and meta-testing include 50 and 15 categories, respectively, where each category contains 100 gray images in the size of $128\times 128$.

In pose prediction, following~\cite{yin2020meta}, the base model is comprised of a fixed encoder with three convolutional blocks and an adapted decoder with four convolutional blocks. Each convolutional block is composed of a convolutional layer, a batch normalization layer and a ReLU activation layer. During the inner-loop optimization, we fix the encoder and only update the parameters in the decoder (i.e., the encoder layers are only meta-updated in the outer-loop optimization). For the hyperparameters in pose prediction, both inner-loop and outer-loop learning rates are set as 0.01. In addition, we set the hyperparameter $\alpha$ in the Beta distribution as 0.5 and the number of adaptation steps in the inner-loop optimization as 5. The candidate set $\mathcal{C}$ for mixup is set to include the input layer (layer 0) as well as all hidden layers (i.e., layer 1, layer 2, and layer 3). All hyperparameters are 
selected according to the performance on the meta-validation set (10 categories), which are randomly selected from the meta-training categories. 
\subsection{Image Classification}
\label{sec:app_imagesetting}
In image classification, the image sizes of Omniglot, MiniImagenet, Multi-dataset are set to be $28\times28\times1$, $84\times84\times3$ and $84\times84\times3$, respectively. Under the non-mutually exclusive setting, taking 5-way miniImagenet as an example, 64 meta-training classes are split to 5 sets, where 4 sets have 13 classes and the rest one has 12 classes. For each set, a fixed class label is assigned to 
each class within this set, which remains unchanged across different tasks. During meta-training, we randomly select one class from each set and take all the five selected classes to construct a task, which ensures that each class consistently has one label across tasks. In our experiments, we list the classes within each set as follows.
\begin{itemize}[leftmargin=*]
    \item \textbf{Set 1}: n07584110, n04243546, n03888605, n03017168, n04251144, n02108551, n02795169, n03400231, n03476684, n04435653, n02120079, n01910747, n03062245
    \item \textbf{Set 2}: n03347037, n04509417, n03854065, n02108089, n04067472, n04596742, n01558993, n04612504, n02966193, n07697537, n01843383, n03838899, n02113712
    \item \textbf{Set 3}: n04604644, n02105505, n02108915, n03924679, n01704323, n09246464, n04389033, n03337140, n06794110, n04258138, n02747177, n13054560, n04443257
    \item \textbf{Set 4}: n13133613, n01770081, n02606052, n02687172, n02101006, n03676483, n04296562, n02165456, n04515003, n01749939, n02111277, n02823428, n01532829
    \item \textbf{Set 5}: n02091831, n07747607, n03998194, n02089867, n02074367, n02457408, n04275548, n03220513, n03527444, n03908618, n03207743, n03047690
\end{itemize}
A similar process is applied to Omniglot, where 1200 meta-training classes are randomly split into 20 sets with 60 classes in each set. In Multi-dataset, each subdataset is split into 5 sets. In the subdatasets Bird, Aircraft and Fungi, we have 4 sets each of which includes 13 classes while the rest one includes 12. In the subdataset Texture, however, each set contains 6 classes. 

For all datasets, we utilize the classical convolutional neural network with 4 convolutional blocks as the base model~\citep{finn2017model,snell2017prototypical}. It is worth to mention that~\citep{yin2020meta} adopts a deeper network as the base model under the non-mutually exclusive setting. The deeper network includes 3 convolutional layers with a fully connected layer as the encoder and 3 convolutional decoder layers, where the encoder is fixed during inner-loop optimization. In our practice, the shallower network achieves better performance in all meta-learning algorithms, as a result of 
more serious overfitting issues caused by the deeper network. In Table~\ref{tab:deep_shallow_compare}, we illustrate the comparison of pre inner-update accuracy and meta-testing post inner-update accuracy during meta-training under the Omniglot 20-way, 1-shot setting, where MAML, MR-MAML are included as baselines. The results indicate that the deeper structure is easier to memorize all data samples via the learned initialization; therefore, we adopt the shallow network (i.e., standard 4-block convolutional layers) in this experiment.
\begin{table}[h]
\caption{Comparison between the shallow and deeper base model under the Omniglot 20-way 1-shot setting.}
\small
\label{tab:deep_shallow_compare}
\begin{center}
\begin{tabular}{l|ccccc}
\toprule
\multirow{2}{*}{Methods} & \multicolumn{2}{c}{Meta-training Pre-update} & \multicolumn{2}{c}{Meta-testing Post-update}  \\
& Shallow & Deep & Shallow & Deep\\\midrule
MAML & $14.38\pm0.40\%$ & $98.59\pm0.05\%$ & $87.40\pm0.59\%$ & $8.82\pm0.42\%$ \\
MR-MAML & $5.63\pm0.36\%$ & $5.12\pm0.34\%$ & $89.28\pm0.59\%$ & $83.75\pm0.67\%$\\
\bottomrule
\end{tabular}
\end{center}
\end{table}

For hyperparameter settings, in both MiniImagenet and Multi-dataset, the inner-loop learning rate $\mu$ and the outer-loop learning rate $\eta$ are set as 0.01, 0.001, respectively. In Omniglot, $\mu$ and $\eta$ are set as 0.1, 0.005, respectively. The hyperparameter $\alpha$ of the Beta distribution $\mathrm{Beta}(\alpha, \alpha)$ is set as 2.0 for all datasets. Besides, the candidate layer set $\mathcal{C}$ for both MiniImagenet and Multi-dataset is set as layer (0, 1, 2, 3). In Omniglot, the candidate set $\mathcal{C}$ is set as layer (1, 2, 3). All hyperparameters are determined by the performance on the meta-validation set.

%% file: app-drug.tex
\section{Additional Results for Drug Activity Prediction}
\subsection{Hyperparameter Analysis on Drug Activity Prediction}
\subsubsection{Analysis of the Candidate Layer Set $\mathcal{C}$}
We further analyze the effect of different candidate layer sets $\mathcal{C}$ on drug activity prediction. The results are reported in Table~\ref{tab:drug_different_layer}. Compared with 
$\mathcal{C}={2}$, $\mathcal{C}={1}$ leads to higher performances, suggesting that mixing low-level representations with the resulting compactness contributes more to the overall improvement. Furthermore, mixing all layers (i.e., $\mathcal{C}=(1,2)$) achieves the best performance, indicating the necessity of jointly mixing the representations in all levels.
\begin{table}[h]
\small
\caption{Effect of the candidate layer set $\mathcal{C}$ in \ours.}
\label{tab:drug_different_layer}
\begin{center}
\setlength{\tabcolsep}{1mm}{
\begin{tabular}{l|ccc|ccc|ccc|ccc}
\toprule
\multirow{2}{*}{Mixed layers $\mathcal{C}$}  & \multicolumn{3}{c|}{Group 1} & \multicolumn{3}{c|}{Group 2} & \multicolumn{3}{c|}{Group 3} & \multicolumn{3}{c}{Group 4}\\
 & Mean & Med. & \begin{small}$>$\end{small}0.3 & Mean & Med. & \begin{small}$>$\end{small}0.3 & Mean & Med. & \begin{small}$>$\end{small}0.3 & Mean & Med. & \begin{small}$>$\end{small}0.3 \\\midrule
(1) &  $0.412$ & $0.362$ & $59$ & $0.326$ & $0.302$ & $50$ & $0.355$ & $0.318$ & $51$ & $0.390$ & $0.349$ & $56$ \\
(2) & $0.405$  & $0.324$ & $51$ & $0.324$ & $0.256$ & $44$ & $0.354$ & $0.304$ & $50$ & $0.387$ & $0.353$ & $57$  \\\midrule
(1,2)   & $\mathbf{0.413}$ & $\mathbf{0.393}$ & 59 & $\mathbf{0.337}$ & $\mathbf{0.301}$ & $\mathbf{51}$ & $\mathbf{0.381}$ & $\mathbf{0.362}$ & $\mathbf{55}$ & $\mathbf{0.380}$ & $\mathbf{0.348}$ & $\mathbf{55}$\\\bottomrule
\end{tabular}}
\end{center}
\end{table}

\subsubsection{Analysis of the Mixup Ratio}
\begin{figure}[h]
\centering
	\begin{subfigure}[c]{0.32\textwidth}
		\centering
		\includegraphics[height=35mm]{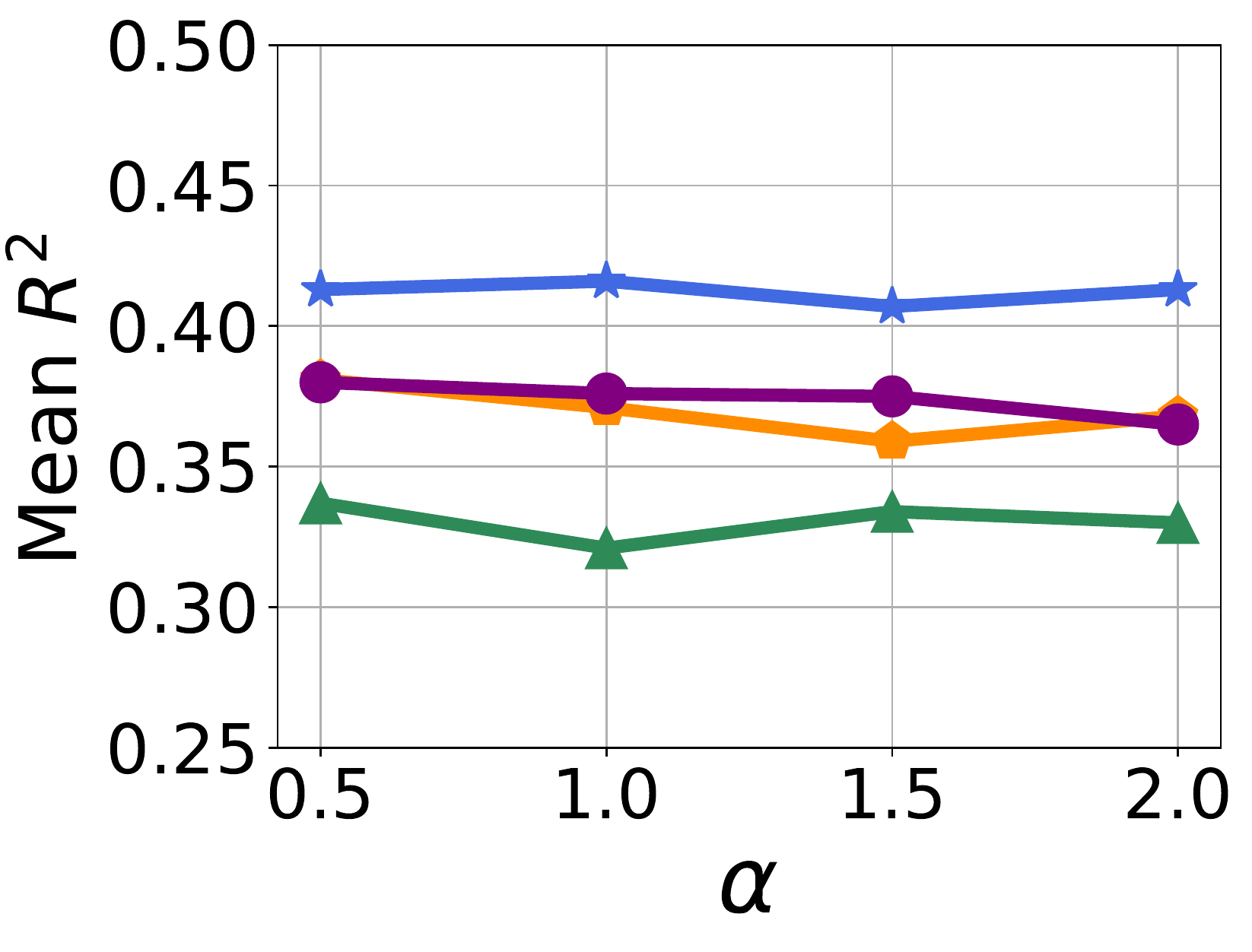}
		\caption{\label{fig:drug_alpha_mean}}
	\end{subfigure}
	\begin{subfigure}[c]{0.32\textwidth}
		\centering
		\includegraphics[height=35mm]{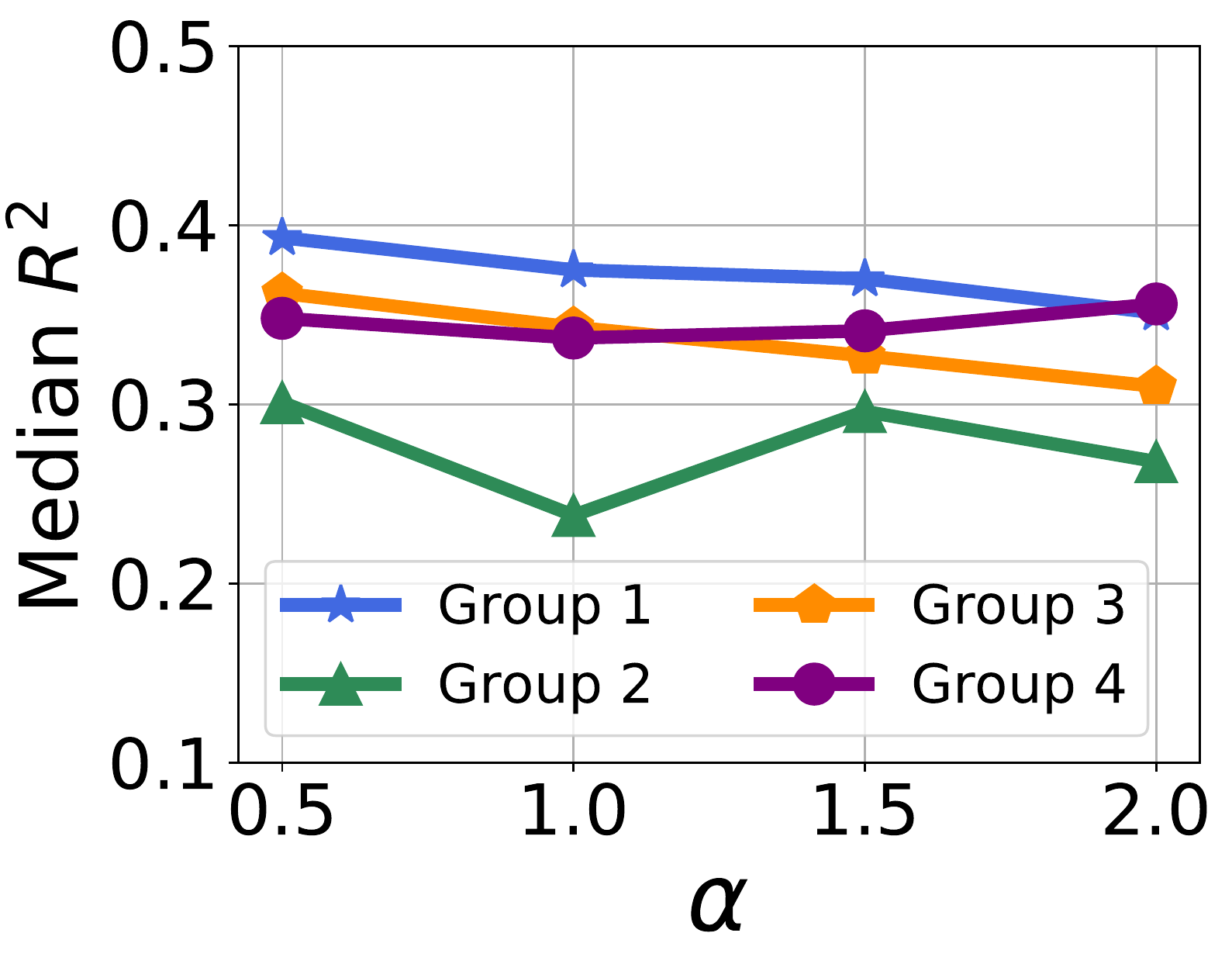}
		\caption{\label{fig:drug_alpha_medium}}
	\end{subfigure}
	\begin{subfigure}[c]{0.32\textwidth}
		\centering
		\includegraphics[height=35mm]{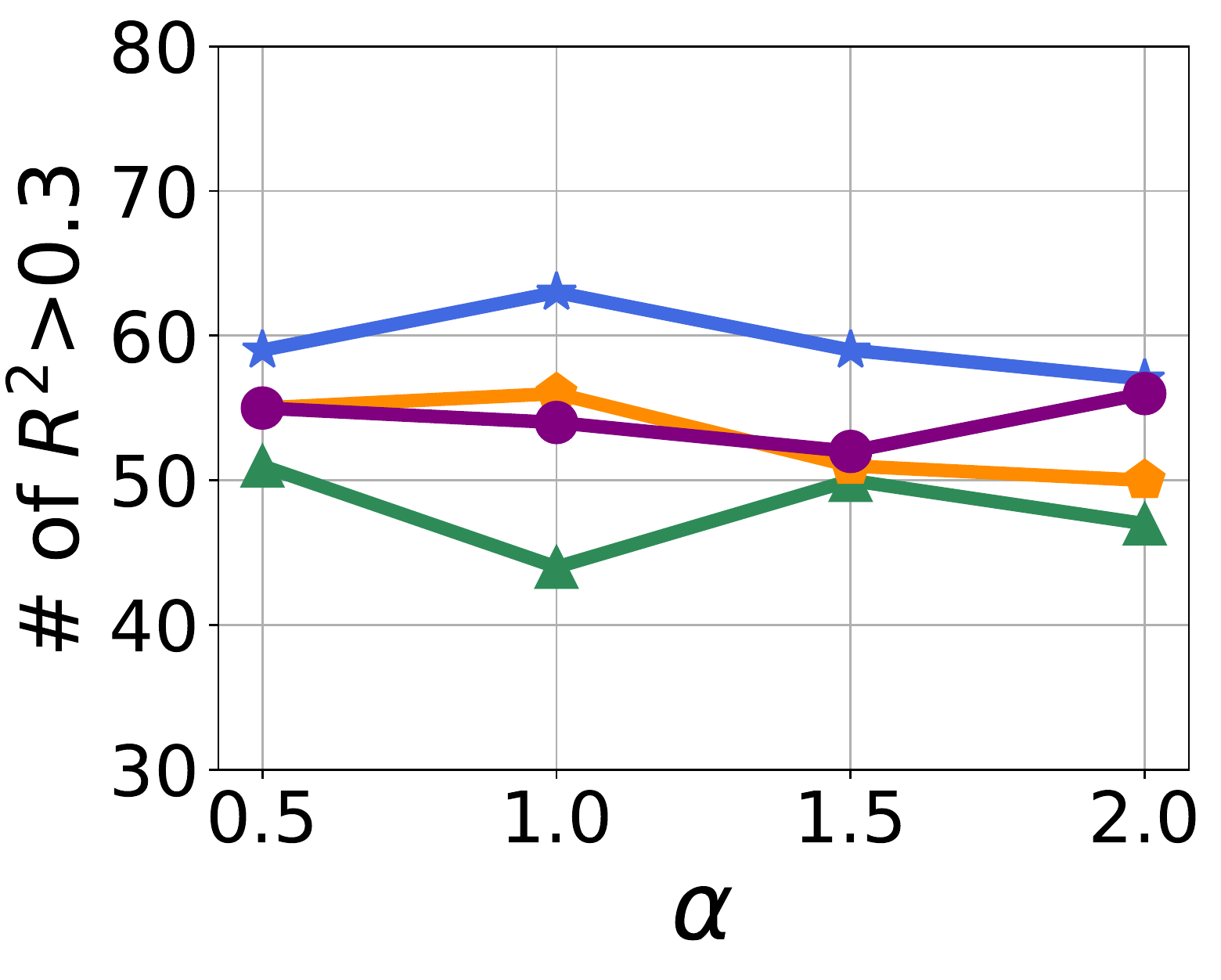}
		\caption{\label{fig:drug_0.3}}
	\end{subfigure}
	\caption{Performance on drug activity prediction w.r.t. the change of $\alpha$ in $\mathrm{Beta}(\alpha, \alpha)$. The three subfigures (a), (b), (c) represent the results under different evaluation metrics.}
	\label{fig:analysis_alpha_drug}
\end{figure}

\begin{figure}[h]
\centering
	\begin{subfigure}[c]{0.32\textwidth}
		\centering
		\includegraphics[height=35mm]{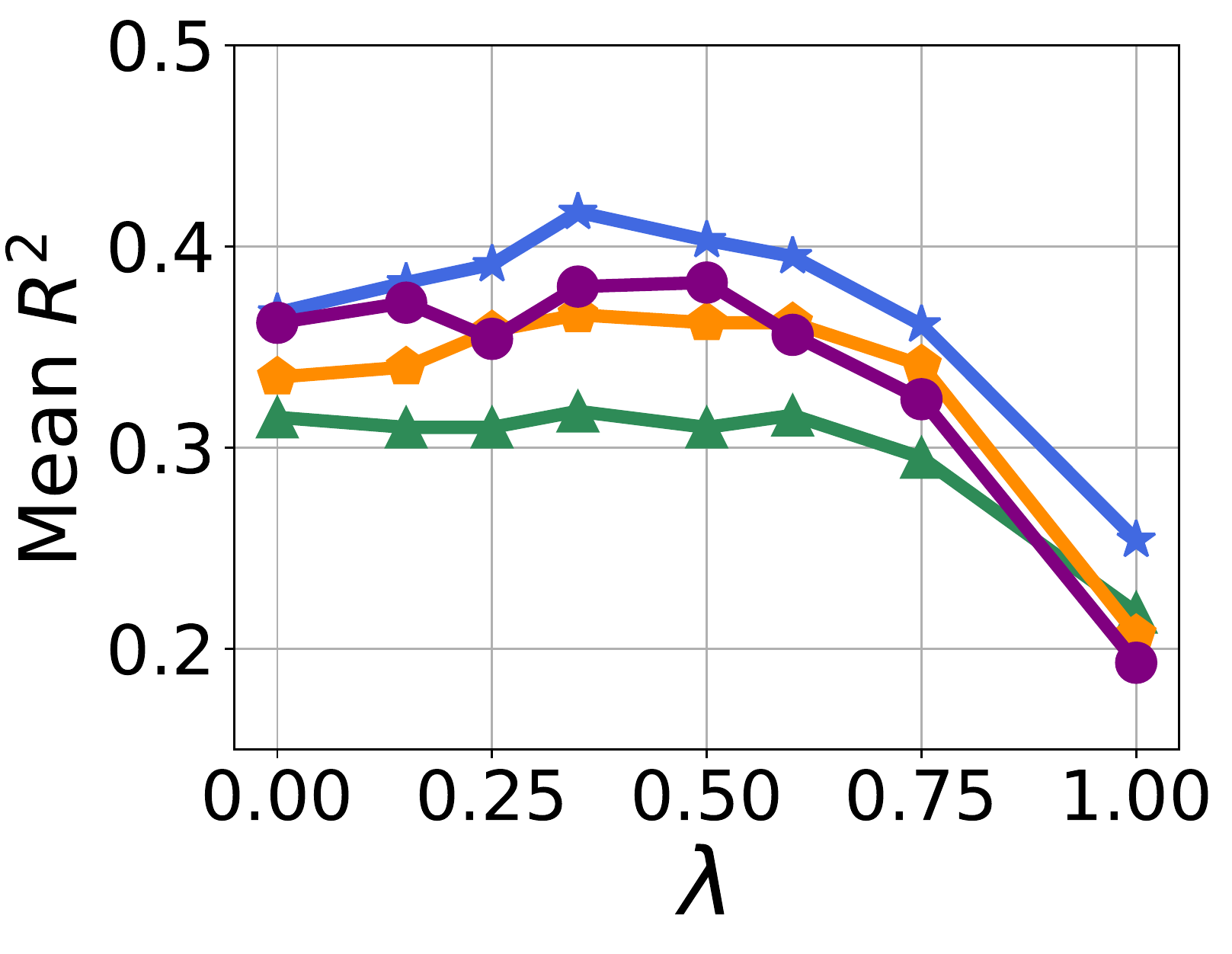}
		\caption{\label{fig:drug_lambda_mean}}
	\end{subfigure}
	\begin{subfigure}[c]{0.32\textwidth}
		\centering
		\includegraphics[height=35mm]{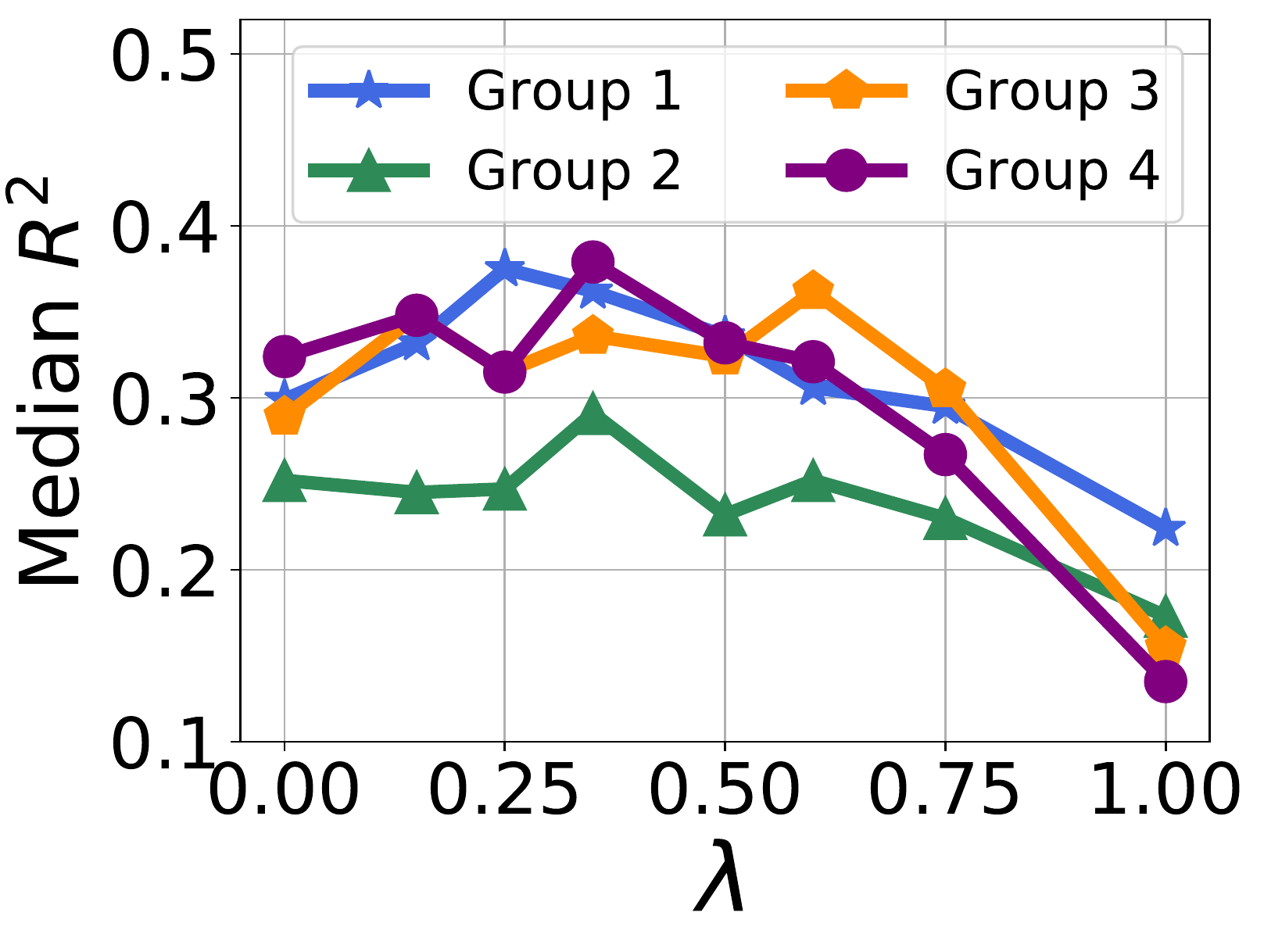}
		\caption{\label{fig:drug_lambda_medium}}
	\end{subfigure}
	\begin{subfigure}[c]{0.32\textwidth}
		\centering
		\includegraphics[height=35mm]{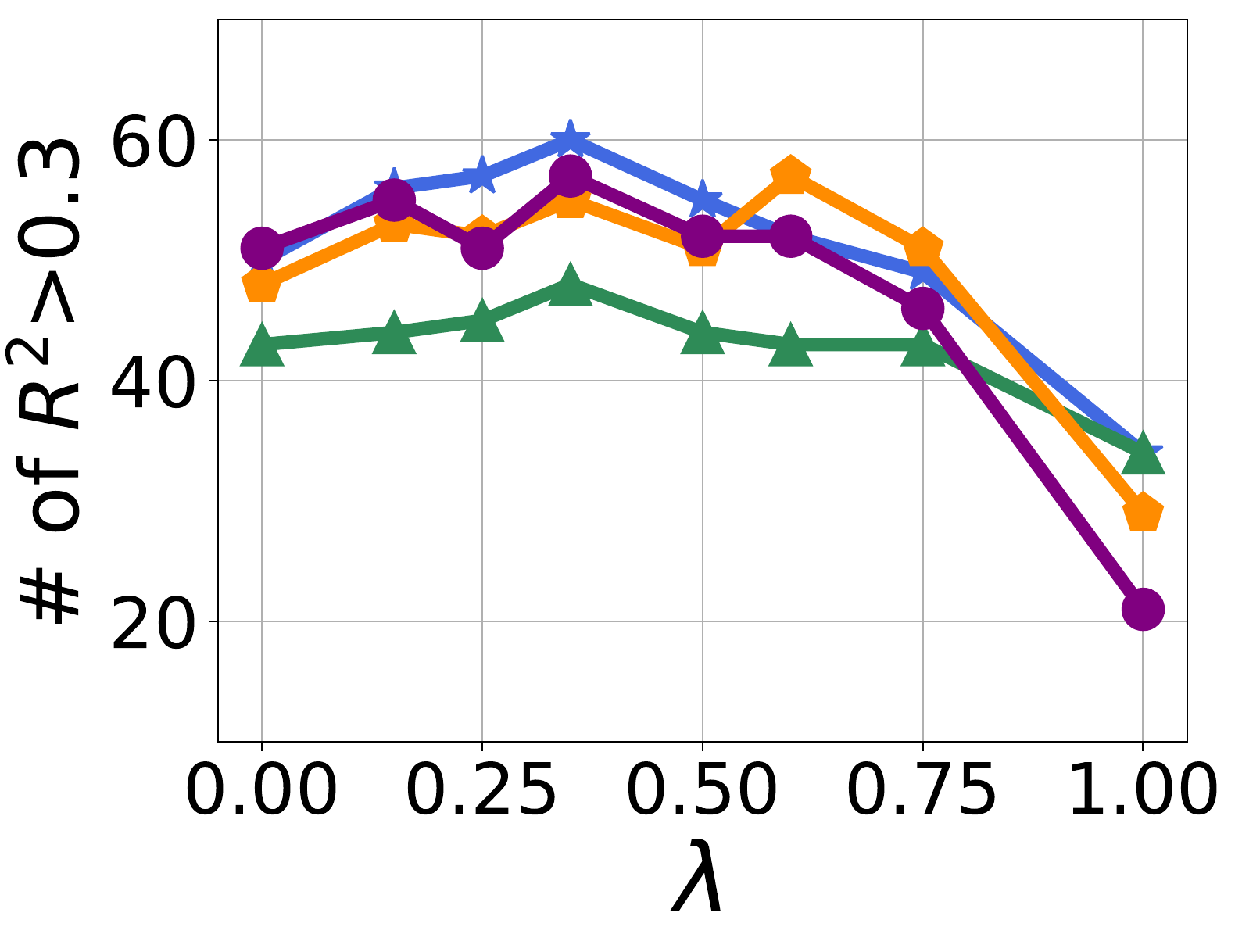}
		\caption{\label{fig:drug_0.3}}
	\end{subfigure}
	\caption{Performance w.r.t. the fixed $\lambda$ in \ours\ (i.e., $\pmb{\lambda} f_{\phi_i^l}(\mathbf{X}^s_i)+(\mathbf{I}-\pmb{\lambda})f_{\phi_i^l}(\mathbf{X}^q_i)$). The three subfigures (a), (b), (c) show the performance under different evaluation metrics (i.e, mean $R^2$, median $R^2$, the number of assays with $R^2>0.3$).}
	\label{fig:analysis_lambda_drug}
\end{figure}

In \ours, the mixup ratio for the support and the query sets are controlled by the parameter $\lambda$, which is sampled from the Beta distribution $\mathrm{Beta}(\alpha, \alpha)$. Here, we analyze the performance w.r.t. the change of mixup ratio. Specifically, we conduct two experiments: (1) we analyze the performance concerning the change of hyperparameter $\alpha$; (2) we fix the mixup ratio $\lambda$ without being sampled from the Beta distribution. The results for the experiments (1) and (2) are shown in Figure~\ref{fig:analysis_alpha_drug} and Figure~\ref{fig:analysis_lambda_drug}, respectively. In the analysis of $\alpha$, though the overall performance is slightly better when $\alpha=0.5$, our \ours\ strategy is still robust and not very sensitive to the shape of $\mathrm{Beta}$ distribution (i.e., different $\alpha$). The conclusion is further strengthened by the analysis of fixed \begin{small}
$\lambda$\end{small} in Figure~\ref{fig:analysis_lambda_drug}, where the performance remains relative stable between \begin{small}$\lambda\in[0.4, 0.75]$\end{small}.

%% file: app-pose.tex
\section{Additional Results for Pose Prediction}
\subsection{Effect of Mixup Strategies on Pose Prediction}
\label{sec:app_poseablation}
In pose prediction, the performance w.r.t. different mixup strategies are reported in Table~\ref{tab:pose_combine_strategy}. The superiority of \ours\ over \kai{the} other strategies further corroborates our analysis that \ours\ is capable of improving the meta-generalization by enhancing the dependence on the support set in the outer-loop optimization.
\begin{table}[h]
\small
\caption{Performance (MSE) of pose prediction w.r.t. different mixup strategies. All data augmentation strategies are applied on MAML.}
\label{tab:pose_combine_strategy}
\begin{center}
    \begin{tabular}{l|ccccc}
    \toprule
    Setting & \begin{small}$\mathcal{D}^{q}$\end{small} & Mixup(\begin{small}$\mathcal{D}^s$\end{small}, \begin{small}$\mathcal{D}^s$\end{small}) & Mixup(\begin{small}$\mathcal{D}^q$\end{small}, \begin{small}$\mathcal{D}^q$\end{small}) & \begin{small}$\mathcal{D}^{cob}$\end{small} & \textbf{\ours} \\\midrule
    10-shot & \yao{$3.098\pm0.242$} & \yao{$4.937\pm0.210$} & \yao{$2.881\pm0.194$} & \yao{$3.112\pm0.165$} & \yao{$\mathbf{2.438\pm0.196}$}\\
    15-shot & \yao{$2.413\pm0.177$} & \yao{$2.701\pm0.168$} & \yao{$2.175\pm0.153$} & \yao{$2.397\pm0.173$} & \yao{$\mathbf{2.003\pm0.147}$}\\\bottomrule
    \end{tabular}
\end{center}
\end{table}

\subsection{Hyperparameter Analysis on Pose Prediction}
\label{sec:app_posehyper}
\subsubsection{Analysis of the Candidate Layer Set $\mathcal{C}$}
Table~\ref{tab:pose_layer} reports the performance w.r.t. the change of the mixup layer set $\mathcal{C}$. Though including the input layer, i.e., layer 0, slightly hurts the performance in some cases, \ours\ 
\kai{still achieves} relative\kai{ly} stable improvements with different mixup layer sets $\mathcal{C}$
\kai{. Besides,} mixing more layers in general enjoys better performance. 
\begin{table}[h]
\small
\caption{Performance (Accuracy) w.r.t. MAML-\ours\ candidate layer set $\mathcal{C}$ under Pose 15-shot setting.}
\label{tab:pose_layer}
\begin{center}
\begin{tabular}{c|c|c|c|c|c|c|c}
\toprule
$|\mathcal{C}|=1$ & Performance & $|\mathcal{C}|=2$ & Performance & $|\mathcal{C}|=3$ & Performance & $|\mathcal{C}|=4$ & Performance \\\midrule
(0)    & \yao{$2.117\pm0.152$} & (0,1) & $\yao{2.136\pm0.165}$ & (0,1,2) & $\yao{2.032\pm0.135}$ & (0,1,2,3) & \yao{$2.003\pm0.147$} \\
(1)    & $\yao{2.120\pm0.150}$ & (0,2) & $\yao{2.080\pm0.156}$ & (0,1,3) & $\yao{2.090\pm0.164}$ & - & - \\
(2)    & $\yao{2.127\pm0.165}$ & (0,3) & $\yao{2.053\pm0.171}$ & (0,2,3) & $\yao{2.047\pm0.149}$ & - & - \\
(3)    & $\yao{2.093\pm0.154}$ & (1,2) & $\yao{2.112\pm0.173}$ & (1,2,3) & $\yao{2.109\pm0.162}$ & - & -\\
-    & - & (1,3) & $\yao{2.094\pm0.153}$ & - & - & - & -\\
-    & - & (2,3) & $\yao{2.134\pm0.167}$ & - & - & - & -\\\bottomrule
\end{tabular}
\end{center}
\end{table}

\subsubsection{Analysis of the Mixup Ratio}
\begin{figure}[h]
\centering
	\begin{subfigure}[c]{0.48\textwidth}
		\centering
		\includegraphics[height=35mm]{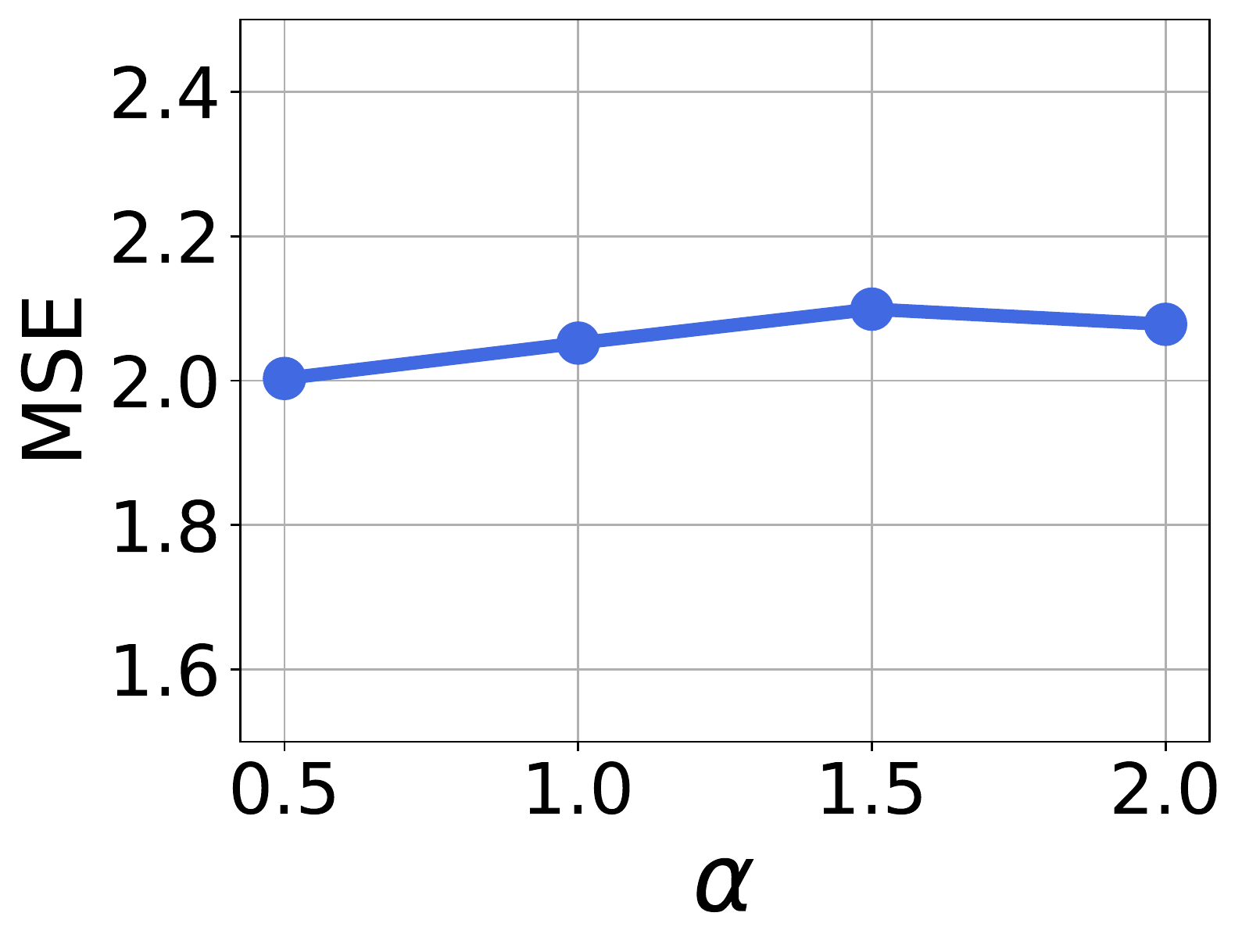}
		\caption{\label{fig:pose_alpha}}
	\end{subfigure}
	\begin{subfigure}[c]{0.48\textwidth}
		\centering
		\includegraphics[height=35mm]{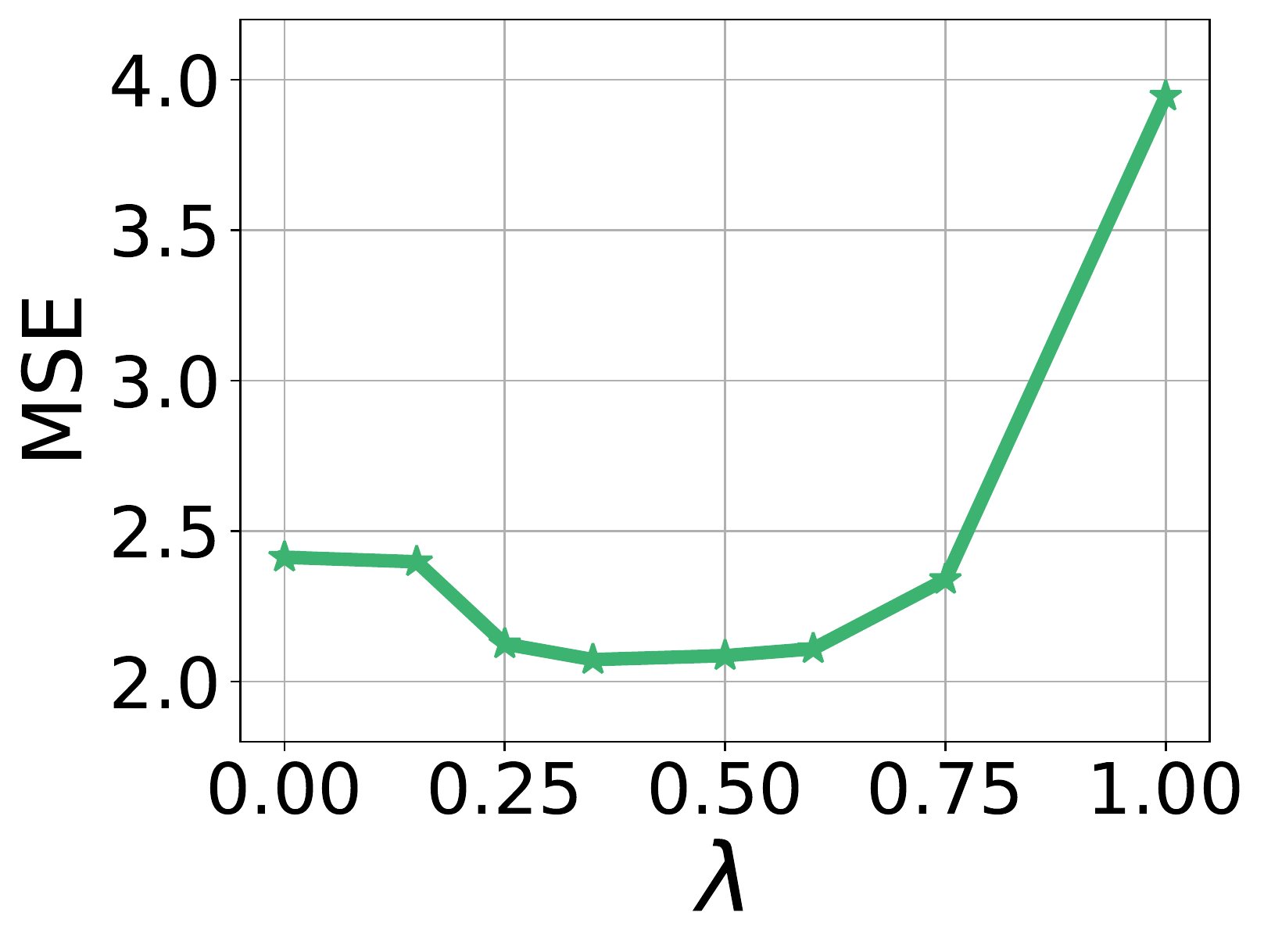}
		\caption{\label{fig:pose_lambda}}
	\end{subfigure}
	\caption{Performance w.r.t. (a) $\alpha$ in $\mathrm{Beta}(\alpha, \alpha)$, and (b): the fixed mixup ratio $\lambda$ under 15-shot pose prediction.}
	\label{fig:analysis_pose}
\end{figure}

In pose prediction, we analyze \kai{the effect of} the mixup ratio by investigating the performance w.r.t. the changes of two key parameters: (1) $\alpha$ in $\mathrm{Beta}(\alpha, \alpha)$; (2) the mixup ratio $\lambda$ in $\mathbf{X}^{mix}_{i,l}=\pmb{\lambda} f_{\phi_i^l}(\mathbf{X}^s_i)+(\mathbf{I}-\pmb{\lambda})f_{\phi_i^l}(\mathbf{X}^q_i)$. We show the results of $\alpha$ and $\lambda$ in Figure~\ref{fig:pose_alpha} and Figure~\ref{fig:pose_lambda}, respectively. The stability of performance w.r.t. $\alpha$ and the stable region $[0.4, 0.75]$ in the analysis of $\lambda$ indicate\kai{s} the robustness of \ours\ under different Beta distribution shapes. \yao{In addition, the preference of $\alpha\!=\!0.5$ may result from that the regression problem is not linear and the value after mix-up should not deviate too much. }

%% file: app-image.tex
\section{Additional Results for Image Classification}
\subsection{Additional Results on Multi-dataset}
\label{sec:app_multifull}
In Table~\ref{tab:heter_image_classification_full}, we report the results (accuracy with 95\% confidence interval) on Muli-datasets.
\begin{table}[h]
\small
\caption{Accuracy with 95\% confidence interval on Multi-dataset.}
\label{tab:heter_image_classification_full}
\begin{center}
\begin{tabular}{l|l|cccc}
\toprule
Setting & Model  & Bird & Texture & Aircraft & Fungi \\\midrule
\multirow{6}{*}{\shortstack{5-way\\1-shot}} 
& MMAML    & $40.03\pm1.87\%$ & $25.43\pm1.61\%$ & $29.33\pm1.69\%$ & $31.13\pm1.63\%$\\
& HSML & $40.49\pm1.78\%$ & $26.40\pm1.66\%$ & $31.67\pm1.68\%$ & $30.43\pm1.66\%$ \\
& ARML & $40.83\pm1.81\%$ & $27.03\pm1.63\%$ & $30.17\pm1.67\%$ & $30.66\pm1.61\%$ \\\cmidrule{2-6}
& \textbf{MMAML-MMCF}  & $51.31 \pm 1.84\%$ & $29.62 \pm 1.76\%$ & $\mathbf{35.41 \pm 1.75\%}$ & $37.67 \pm 1.80\%$\\
& \textbf{HSML-MMCF}  & $51.78 \pm 1.89\%$ & $29.51 \pm 1.80\%$ & $34.97 \pm 1.74\%$ & $38.20 \pm 1.84\%$\\
& \textbf{ARML-MMCF} & $\mathbf{53.17 \pm 1.86\%}$ & $\mathbf{30.08\pm 1.76\%}$ & $35.04 \pm 1.78\%$ & $\mathbf{38.70\pm 1.83\%}$ \\\midrule
\multirow{6}{*}{\shortstack{5-way\\5-shot}}
& MMAML    & $61.64\pm0.96\%$ & $34.76\pm0.80\%$ & $51.89\pm0.93\%$ & $44.48\pm0.96\%$\\
& HSML & $61.07\pm1.04\%$ & $35.48\pm0.83\%$ & $48.07\pm0.91\%$ & $43.42\pm0.94\%$ \\
& ARML & $64.31\pm0.99\%$ & $36.11\pm0.83\%$ & $50.76\pm0.97\%$ & $46.11\pm0.95\%$ \\\cmidrule{2-6}
& \textbf{MMAML-MMCF}  & $72.04 \pm 0.93\%$ & $40.14 \pm 0.85\%$ & $64.59 \pm 0.90\%$ & $51.11 \pm 1.00\%$\\
& \textbf{HSML-MMCF}  & $72.53 \pm 0.92\%$ & $40.39 \pm 0.83\%$ & $64.31\pm 0.92\%$ & $51.04 \pm 1.04\%$\\
& \textbf{ARML-MMCF}  & $\mathbf{73.30 \pm 0.90\%}$ & $\mathbf{40.88 \pm 0.83\%}$ & $\mathbf{65.18\pm0.89\%}$ & $\mathbf{51.56 \pm 1.03\%}$ \\\bottomrule
\end{tabular}
\end{center}
\vspace{-2em}
\end{table}

\subsection{Results under Mutually-exclusive Setting}
\label{sec:app_imagenormal}
In Table~\ref{tab:mutually_exclusive_image_classification}, we report the results under the standard mutually-exclusive setting on MiniImagenet. Under the mutually-exclusive setting, the mechanism of label shuffling is introduced to construct meta-training tasks, which significantly alleviates the meta-overfitting issue. However, applying the proposed MetaMix and Channel Shuffle on this setting still achieves comparable and even better performance than original MAML, which further demonstrates the effectiveness of our data augmentation strategies to improve meta-generalization.
\begin{table}[h]
\small
\caption{Performance (Accuracy) of MiniImagenet under the mutually-exclusive setting.}
\label{tab:mutually_exclusive_image_classification}
\begin{center}
\begin{tabular}{l|cc}
\toprule
\multirow{2}{*}{Model}  & \multicolumn{2}{c}{MiniImagenet}\\
 & 5-way 1-shot & 5-way 5-shot \\\midrule
MAML & $48.70\pm1.84\%$ & $63.11\pm0.92\%$\\
MAML-Channel Shuffle & $50.08 \pm 1.86\%$ & $64.70 \pm 0.95\%$ \\
MAML-\ours\ & $50.02\pm1.83\%$ & $64.13\pm0.95\%$ \\
MAML-MMCF & $\mathbf{50.35 \pm 1.82\%}$ & $\mathbf{64.91 \pm 0.96\%}$\\\bottomrule
\end{tabular}
\end{center}
\end{table}

\subsection{Hyperparameter Analysis}
\label{sec:app_imagehyper}
\subsubsection{Analysis of the Candidate Layer Set $\mathcal{C}$}
In Table~\ref{tab:layer_set_miniimagenet}, we analyze the effect of the candidate layer set $\mathcal{C}$ and report the performance of MAML-MMCF under the 5-shot MiniImagenet scenario. Similar to the findings in drug activity \kai{prediction}, in all scenarios, incorporating MMCF into MAML improves the performance, indicating the robustness of MMCF 
\kai{with} different candidate layer sets. Besides, we 
\kai{observe} that involving all layers achieves the best performance.
\begin{table}[h]
\caption{Performance (Accuracy) w.r.t. the selected layer set $\mathcal{C}$ under the MiniImagenet 5-shot scenario.}
\small
\label{tab:layer_set_miniimagenet}
\begin{center}
\begin{tabular}{c|c|c|c|c|c}
\toprule
$|\mathcal{C}|=1$ & Performance & $|\mathcal{C}|=2$ & Performance & $|\mathcal{C}|=3$  & Performance \\\midrule
(1)    & \yao{$57.74 \pm 0.95\%$} & (1,2) & \yao{$57.88 \pm 0.94\%$} & (1,2,3) & \yao{$58.96 \pm 0.95\%$} \\
(2)    & \yao{$57.31 \pm 0.96\%$} & (1,3) & \yao{$56.91 \pm 0.97\%$} & - & -  \\
(3)    & \yao{$57.19 \pm 0.98\%$} & (2,3) & \yao{$58.19 \pm 0.93\%$} & - & - \\\bottomrule
\end{tabular}
\end{center}
\vspace{-2em}
\end{table}

\subsubsection{Analysis of the Mixup Ratio and the Skewed Beta Distribution}
\begin{figure}[h]
	\centering
	\begin{subfigure}[c]{0.24\textwidth}
		\centering
		\includegraphics[height=24mm]{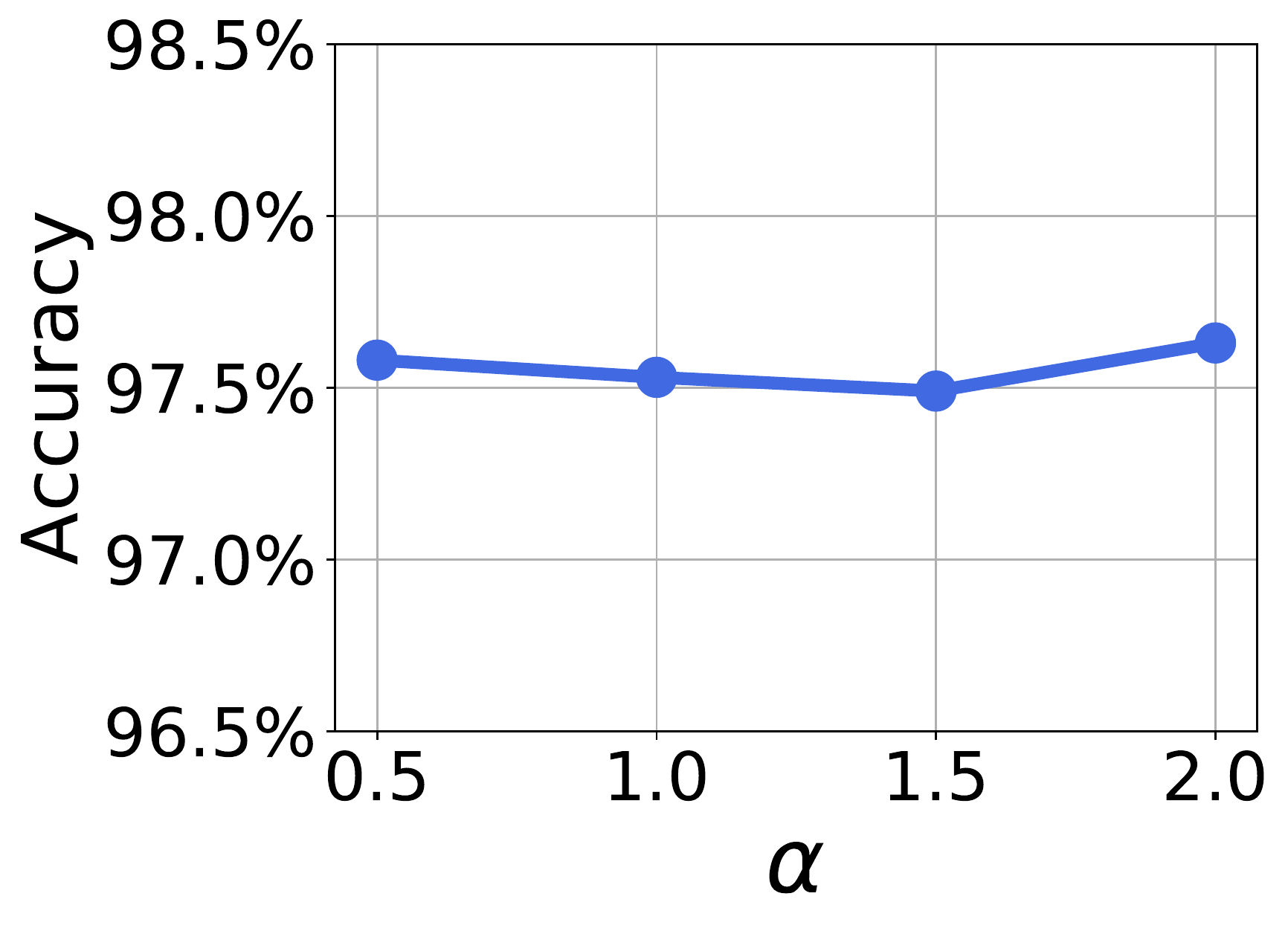}
		\caption{\label{fig:omniglot_alpha_app}}
	\end{subfigure}
	\begin{subfigure}[c]{0.24\textwidth}
		\centering
		\includegraphics[height=24mm]{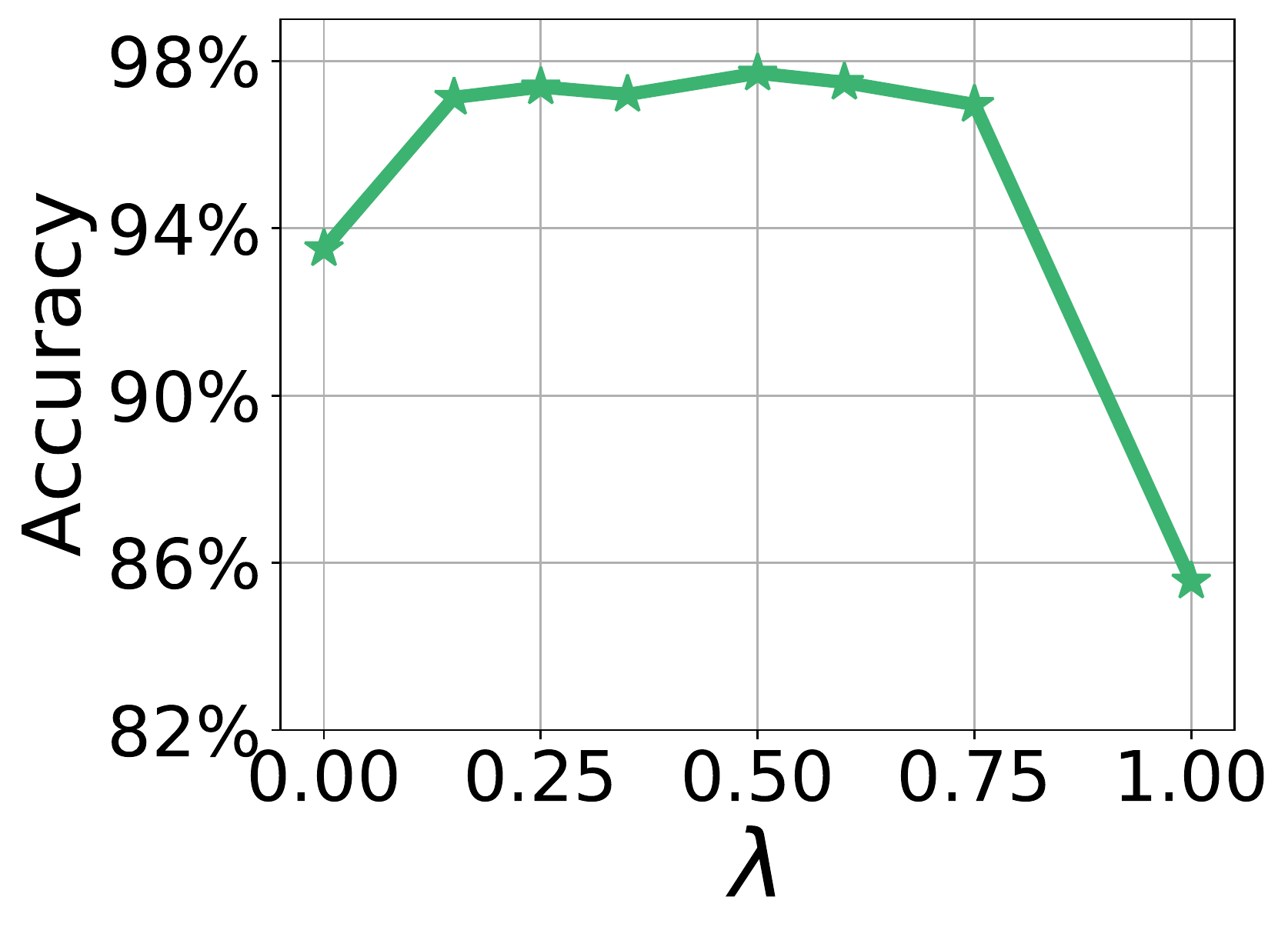}
		\caption{\label{fig:omniglot_lambda_app}}
	\end{subfigure}
	\begin{subfigure}[c]{0.24\textwidth}
		\centering
		\includegraphics[height=24mm]{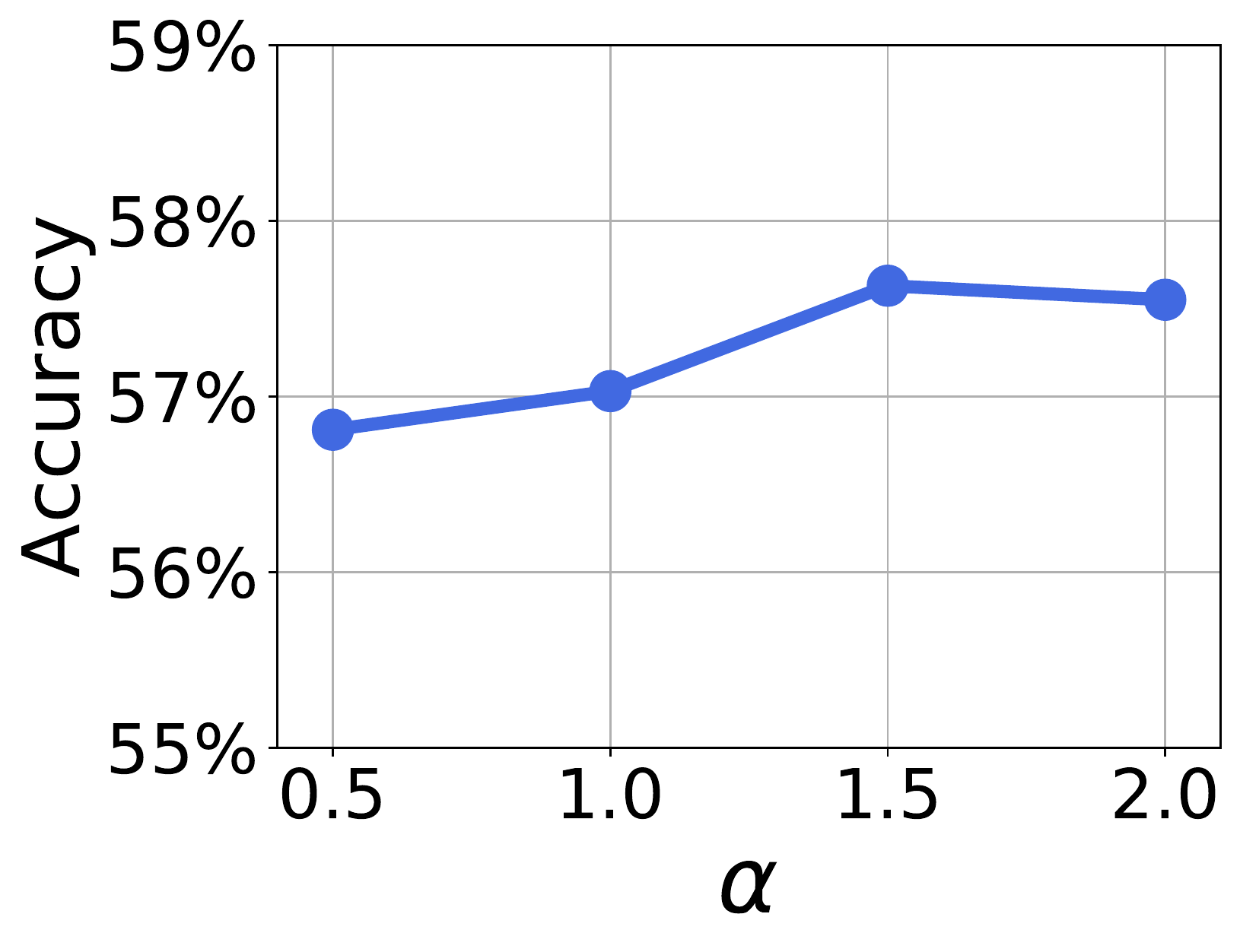}
		\caption{\label{fig:miniimagenet_alpha_app}}
	\end{subfigure}
	\begin{subfigure}[c]{0.24\textwidth}
		\centering
		\includegraphics[height=24mm]{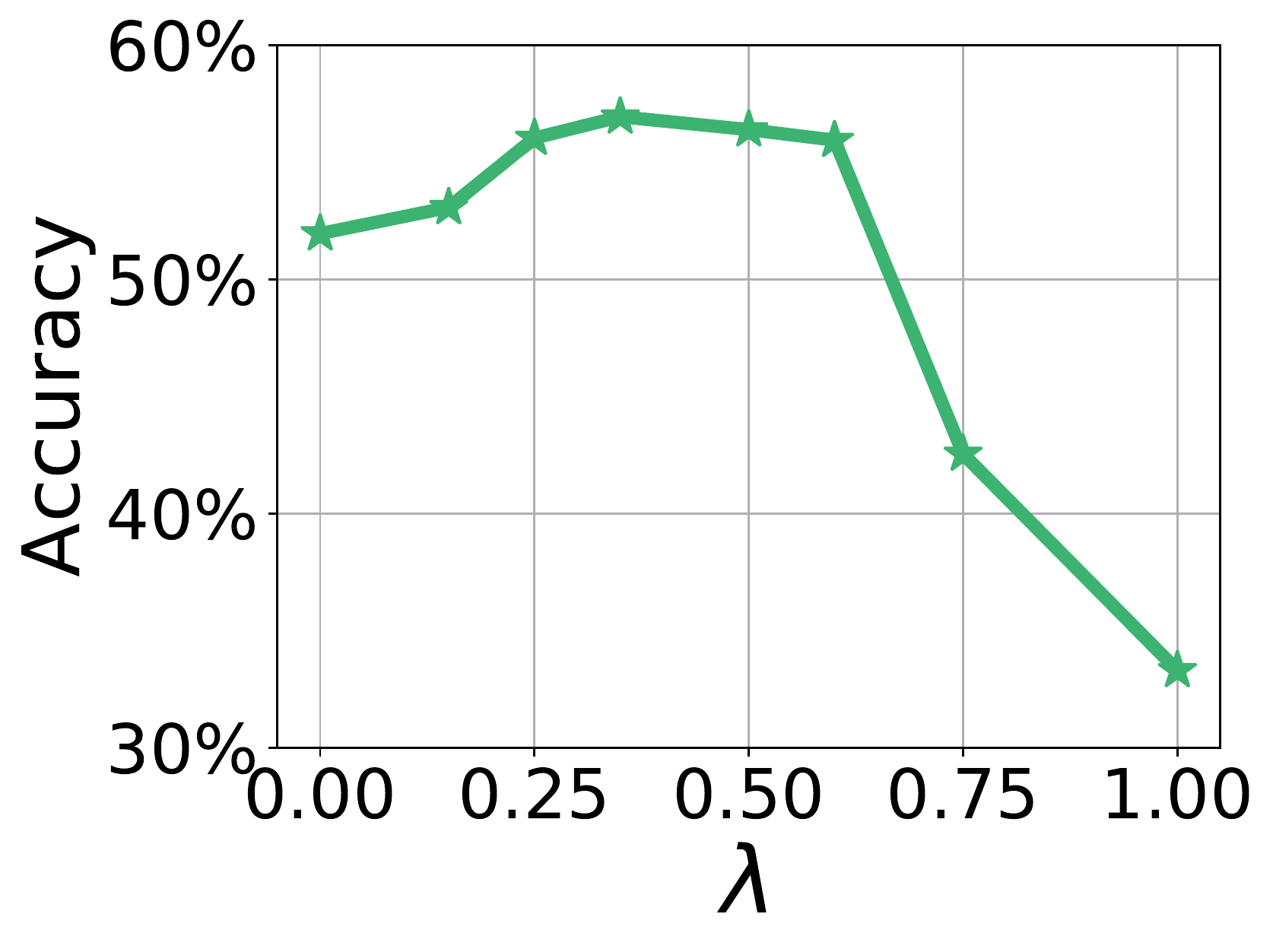}
		\caption{\label{fig:miniimagenet_lambda_app}}
	\end{subfigure}
	\caption{Performance w.r.t. (a)(c) $\alpha$ in $\mathrm{Beta}(\alpha, \alpha)$ distribution; (b)(d) mixup ratio $\lambda$. (a)(b) show the results under the Omniglot 20-way, 5-shot setting; (c)(d) illustrate the performance under the MiniImagenet 5-way, 5-shot scenario.}
	\label{fig:performance_sensitivity_image}
\end{figure}

Under the Omniglot and MiniImagenet 5-shot setting, we further investigate the effect of key hyperparameters for MetaMix (i.e., $\alpha$ and $\lambda$). The results of MAML-MetaMix on MiniImagenet and Omniglot are shown in Figure~\ref{fig:performance_sensitivity_image}. Similar to the previous analys\yingicml{es} 
on drug activity prediction and pose prediction, the stability of performance w.r.t. $\alpha$ and the stable region in $\lambda$ analysis demonstrate\kai{s} the robustness of \ours. The conclusion is further supported by the analysis of skewed Beta distribution (i.e., \begin{small}$\alpha\neq \beta$\end{small}), whose results under the MiniImagenet 5-shot setting are reported in Table~\ref{tab:skew_beta}.

\begin{table}[!h]
\caption{Effect of skewed Beta distribution (i.e., $\lambda\sim \mathrm{Beta}(\alpha, \beta)$ and $\alpha\neq \beta$) under the MiniImagenet 5-shot setting.}
\small
\label{tab:skew_beta}
\begin{center}
\begin{tabular}{c|cccc}
\toprule
Settings & $\alpha=0.5$ & $\alpha=1.0$ & $\alpha=2.0$  & no \ours\\\midrule
$\beta=0.5$ & $55.35\pm0.96\%$ & $53.82\pm0.99\%$ & $53.05\pm0.93\%$ & \multirow{3}{*}{$51.95\pm0.97\%$}\\
$\beta=1.0$ & $53.38\pm0.94\%$ & $56.12\pm1.02\%$ & $54.91\pm1.01\%$ & \\
$\beta=2.0$ & $50.01\pm0.96\%$ & $53.69\pm0.96\%$ & $57.55\pm0.97\%$ &\\
\bottomrule
\end{tabular}
\end{center}
\end{table}